%% file: main.tex
\documentclass[10pt,twocolumn]{article}

\usepackage{arxiv_uai_style}
\input{preamble.tex}

\title{Data-driven discovery of intrinsic Lie point symmetries of differential equation systems}
\author{%
\begin{tabular*}{0.92\textwidth}{@{\extracolsep{\fill}}ccc}
\textbf{Ilker Gurcan}\textsuperscript{1} &
\textbf{Kadir Ustun}\textsuperscript{2} &
\textbf{Ramazan Gokberk Cinbis}\textsuperscript{1,3}
\end{tabular*}\\[1.2ex]
\textsuperscript{1}Department of Computer Engineering, Middle East Technical University, Ankara, T\"{u}rkiye\\
\textsuperscript{2}ASELSAN Inc., Ankara, T\"{u}rkiye\\
\textsuperscript{3}ROMER Robotics and AI Center, Middle East Technical University, Ankara, T\"{u}rkiye
}
\date{}

\makeatletter
\global\let\arxivtitlecopy\@title
\global\let\arxivauthorcopy\@author
\makeatother

\begin{document}
\maketitle

\begin{abstract}
Discovering continuous symmetries directly from differential-equation data can reveal geometric structure for model reduction and scientific machine learning without symbolic determining equations or a closed-form representation of the governing system. We formulate intrinsic Lie point-symmetry discovery as an operator-theoretic problem on sampled jet-space data. Admissible generators are identified from a sampled determining operator and separated from redundant, solution-dependent, and geometrically trivial directions. The framework combines prolongation, sequential subspace restrictions, and a fixed nonlinear dictionary linear in trainable coefficients. A rank-stabilization preflight test diagnoses systems whose represented intrinsic symmetry algebra does not stabilize under dictionary refinement. On a controlled differential-equation benchmark with an analytically known 11-dimensional represented intrinsic symmetry algebra, the method recovers the complete represented generator set, reproduces its Lie-bracket structure to numerical precision, and approximates a known generator beyond the training domain. Here we show that explicit intrinsic continuous-symmetry generators can be recovered from data obtained from exact or approximate differential systems.
\end{abstract}

\section{Introduction}
\label{sec:introduction}
Symmetry has long played a central role in both mathematics and the physical sciences. A landmark development was Galois's characterization of the solvability of polynomial equations by radicals in terms of the structure of their associated permutation groups, now known as \textit{Galois groups}; this connection became foundational to Galois theory and modern field theory~\cite[Part~IV]{dummit2003abstract}. Meanwhile, a wide range of physical phenomena can be described mathematically by systems of differential equations (DEs). Motivated in part by Galois's ideas, Sophus Lie sought an analogous theory for DEs and developed continuous transformation groups as a systematic means of studying, simplifying, and integrating them~\cite{helgason1990sophus}. However, the advanced machinery for both Lie groups and differential systems was developed by \'{E}lie Cartan~\cite{ivey2003cartan,gardner1989method}. When admitted symmetries are known, they can reduce the effective dimensionality or differential complexity of a system, for example by lowering the order of an ordinary differential equation (ODE) or reducing the number of independent variables of a partial differential equation (PDE), thereby transforming the original problem into a more tractable reduced system~\cite[Example~2.65]{olver2000applications}. Beyond dimensional and order reduction, symmetries can also reveal invariant quantities and structures carrying direct physical meaning. For variational systems, for example, Noether's theorem establishes a fundamental correspondence between continuous symmetries and conservation laws: invariance under time translations, spatial translations, and rotations is associated with conservation of energy, linear momentum, and angular momentum, respectively~\cite{hassani2013mathematical}.

Standard infinitesimal Lie-symmetry analysis begins with the infinitesimal invariance condition, reviewed in Section~\ref{sec:preliminaries}. In the classical approach, a candidate \textit{symmetry generator} is first prolonged to the required jet order and then restricted to the equation manifold. The resulting symbolic expression is subsequently split with respect to the unconstrained jet-fiber coordinates, yielding an overdetermined system of linear PDEs for the unknown infinitesimal coefficients of the candidate generator~\cite{ibragimov1995crc,olver2000applications}. The resulting system is known as the determining system of the given DE system. A first limitation of this approach is that it operates on an explicit symbolic representation of the governing equations and therefore cannot be applied directly to data sampled from the corresponding equation manifold. Moreover, systematic treatment of complicated determining systems may involve exact differential-elimination and involutive-completion algorithms~\cite{schwarz1992algorithm}, which may be viewed as differential analogues of Gaussian elimination in linear algebra. Such algorithms are naturally suited to differential-polynomial or rational systems. Non-polynomial determining systems can still be simplified symbolically, but their treatment may require auxiliary variables, additional analytic assumptions, heuristic simplification, symbolic integration, or generalized separation~\cite{carminati2000symbolic}. Finally, the size and complexity of the determining system can grow rapidly as the numbers of independent variables, dependent variables, equations, or differential orders increase. The resulting symbolic expression growth, hidden integrability conditions, and possible case distinctions can make the analysis substantially more difficult.

More structural symbolic approaches seek to extract information directly from an involutive or standard form of the determining system without explicitly integrating it. In particular, the associated \textit{initial data} can be used to determine the number of arbitrary constants and functions in the local solution space and, consequently, the dimension or functional freedom of the admitted symmetry algebra~\cite{reid1991finding}. Although this avoids the need to obtain closed-form expressions for the infinitesimal coefficients, such methods still require an explicit symbolic representation of the governing equations. Moreover, the resulting initial-data representation does not by itself provide an explicit symmetry vector field that can be evaluated throughout a prescribed region. To obtain such a regional representation, the infinitesimal coefficients must be reconstructed, for example through local Taylor expansions generated from the involutive system and its initial data~\cite{reid1991algorithms}, through symbolic integration when available, or through an appropriate numerical continuation of the locally specified symmetry data. The reconstruction becomes increasingly burdensome as the amount of free initial data grows. This distinction is particularly important for the present work, whose intended downstream applications require readily evaluable symmetry generators over sampled regions, including their eventual use in the construction of invariant $G$-coframes~\cite{fels1998moving}.

On the machine-learning (ML) side, symmetry has also become increasingly important for understanding and improving learned representations. Mallat interprets deep convolutional neural-network (CNN) representations through multiscale contractions and the progressive linearization of local or hierarchical symmetry transformations, thereby reducing irrelevant variability while preserving discriminative information~\cite{mallat2016understanding}. More generally, incorporating previously known symmetries into neural architectures or learning constraints can improve sample efficiency and generalization~\cite{finzi2021practical,rezende2019equivariant}. In particular, equivariant architectures exploit prescribed group actions to restrict the admissible hypothesis space, allowing the learned model to respect transformations that are known a priori to preserve the underlying problem structure.

In parallel, emerging methods for representing and solving differential-equation systems have become increasingly data-driven. Neural ordinary differential equation (Neural ODE) models parameterize an unknown continuous-time vector field by a neural network and obtain the corresponding trajectories through numerical integration~\cite{chen2018neural}, in contrast to conventional residual networks, whose layerwise transformations define a discrete-depth evolution~\cite{behrmann2019invertible}. Physics-informed neural networks (PINNs), on the other hand, parameterize an approximate solution $\tilde{u}_{\theta}$ and train the learnable parameters $\theta$ by penalizing the governing differential-equation residual together with prescribed initial and/or boundary conditions~\cite{raissi2019physics}; their use of neural representations is supported by the universal approximation capabilities of neural networks~\cite{kidger2020universal}. Koopman operator theory provides a complementary operator-theoretic description of nonlinear dynamical systems through the linear evolution of observable functions on the state space~\cite{brunton2021modern}. Originally introduced by Bernard Koopman for Hamiltonian systems in 1931~\cite{koopman1931hamiltonian}, the exact Koopman operator is generally infinite-dimensional, motivating modern data-driven approaches that seek finite-dimensional observable subspaces or embeddings in which the dynamics can be represented approximately, or in special cases exactly, by linear evolution~\cite{budivsic2012applied,brunton2021modern}.

Continuous symmetries have also recently been incorporated into data-driven methods for differential equations. In many existing approaches, the relevant Lie group or Lie algebra is assumed to be known in advance and is subsequently encoded into the learning architecture, training data, or optimization constraints~\cite{brandstetter2022lie,mialon2023self,akhound2023lie}. Of particular relevance to the present work are recent methods that extend beyond enforcing prescribed symmetries to discovering previously unspecified symmetries from models or dynamical-system data~\cite{otto2023unified,tahmasebi2026adaptive,liu2026liestonet,liu2026liedynnet}. LieStoNet targets projectable Lie point symmetries of stochastic differential equations by first learning neural surrogates for the drift and diffusion and then jointly learning generators under determining-equation and Lie-algebraic regularizers; its associated Fokker--Planck formulation also removes the infinite-dimensional additive solution sector structurally from the learned generator class~\cite{liu2026liestonet}. LieDynNet similarly learns multiple Lie point-symmetry generators jointly from spatiotemporal ODE/PDE data through differentiable neural surrogates, combining prolonged infinitesimal-invariance and finite-flow conditions with Lie-algebraic constraints~\cite{liu2026liedynnet}. These developments demonstrate growing interest in learning continuous symmetry structure directly from dynamical data, while also highlighting that the computational formulation depends strongly on the target system class and on how admissibility and algebraic structure are imposed.

The present work takes a different route. Rather than first reconstructing a dynamical surrogate and then learning a jointly constrained generator basis, we formulate the determining problem directly on sampled jet-space data associated with the differential system. The sampled determining operator is used to construct a finite represented admissible space before nonlinear optimization begins; within the linear-DE classes treated here, the additive solution-dependent sector is removed at the represented-space level, and the remaining intrinsic directions are recovered sequentially under global novelty and geometric nontriviality restrictions. In this sense, the structural exclusion used for the Fokker--Planck superposition sector in LieStoNet and the represented additive-sector removal developed here share the same practical principle---a large solution-dependent symmetry sector should be removed structurally rather than left to compete with the target directions through soft optimization---but they realize that principle in different problem settings.

\begin{figure}[t]
    \centering
    \includegraphics[width=\columnwidth]{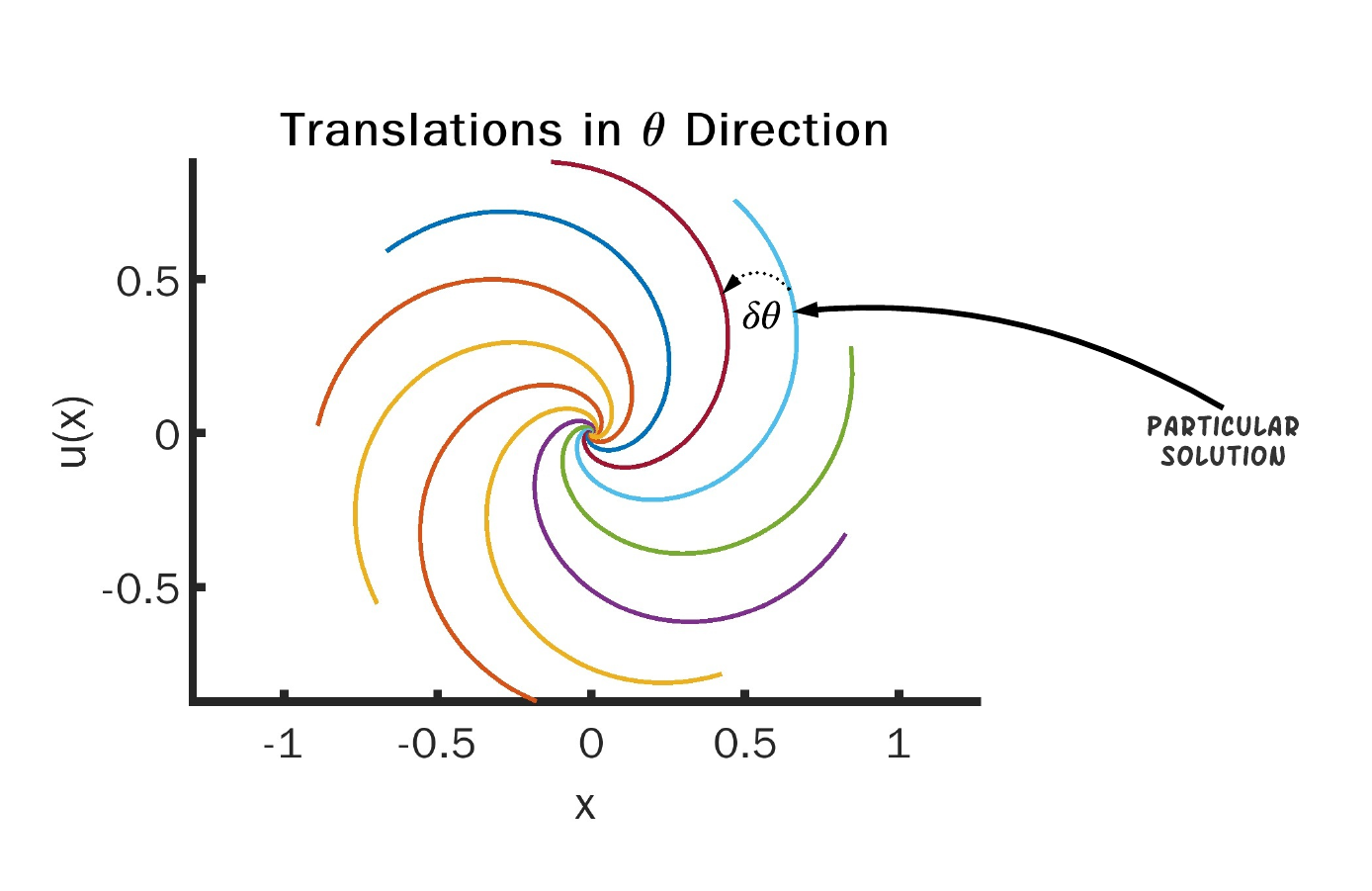}
    \caption{Several solution curves of $\Delta^{(1)}$ in \eqref{eqn:sampleODEforMotiv} related by its rotational symmetry. The remaining solution curves are obtained by rotating the particular solution shown in light blue.}
    \label{fig:SampleDERotSymm}
\end{figure}

To illustrate the potential benefit of discovering symmetries directly from data, consider the following nonlinear first-order ODE in the first jet bundle~\cite{starrett2007solving}:
\begin{equation}
\label{eqn:sampleODEforMotiv}
    \Delta^{(1)}\paraBrTerm{x,u,u_x} = u_x-\frac{u^3+x^2u-u-x}{x^3+u^2x+u-x} = 0.
\end{equation}
At first glance, \eqref{eqn:sampleODEforMotiv} is neither separable nor suggestive of an obvious simplifying change of coordinates. If only jet-space data sampled from its equation manifold were available, identifying such a transformation would be even less direct. Fortunately, the equation admits $\mathrm{SO}(2)$ as a point-symmetry group. Under the polar-coordinate transformation
\begin{equation}
    x=r\cos\theta,
    \qquad
    u=r\sin\theta,
\end{equation}
where $(r,\theta)$ form local coordinates on the total space $E$ and $r$ is locally taken as the transformed independent variable, \eqref{eqn:sampleODEforMotiv} becomes
\begin{equation}
    \label{eqn:sampleDESymmApp}
    \overline{\Delta^{(1)}}\paraBrTerm{r,\theta_r} = \frac{d\theta}{dr} - \frac{1}{r(1-r^2)} = 0.
\end{equation}
A crucial observation is that rotations in the original $(x,u)$ coordinates become translations in the transformed angular coordinate $\theta$. Indeed, the infinitesimal rotational generator
\begin{equation}
    v = -u\partial_x+x\partial_u
\end{equation}
is straightened to $\partial_\theta$ in polar coordinates. Consequently, the transformed equation is independent of $\theta$: the group coordinate has become cyclic. This is the local manifestation of symmetry reduction by a one-parameter group and, for the present first-order ODE, reduces the problem to a quadrature. Moreover, the originally nonseparable equation becomes explicitly separable in the symmetry-adapted coordinates. Thus, knowledge of a single admitted symmetry converts a comparatively complicated differential relation into a substantially simpler representation for analytical, numerical, or learning-based treatment. The natural question motivating the present work is then: \textit{what if such a useful symmetry were not known a priori and only sampled differential-system data were available?}

The limitations of the preceding approaches, together with the motivating example above, capture the central objective of this work: \textit{data-driven recovery of the intrinsic Lie point symmetries of a DE system from sampled data}. The term \textit{intrinsic}, as used here, is not intended as a canonical definition from the existing literature; its precise meaning is introduced in Section~\ref{sec:preliminaries}. A distinguishing feature of the proposed framework, developed in Section~\ref{sec:framework}, is its explicit use of the jet-space formalism, which provides the natural geometric setting for DEs and their prolongations~\cite[Chapter~9]{olver1995equivalence}. Within this setting, the objective is to recover a finite representation of the intrinsic Lie algebra $\mathfrak{g}_{\mathrm{int}}$ directly from sampled jets. Our earlier numerical experiments indicated that imposing all admissibility, nontriviality, solution-independence, and novelty requirements simultaneously leads to a difficult multi-objective optimization problem. We therefore adopt an operator-theoretic, feasible-space-first strategy: rather than asking an optimizer to discover admissible symmetry data from an unconstrained parameterization, we first construct a finite-dimensional admissible search space and subsequently optimize only within that space. Successive restrictions exclude directions associated with sampled geometric triviality, additive solution-dependence, and previously discovered generators, thereby progressively isolating admissible intrinsic candidates.

The structure of the admissible generator space also depends strongly on the class of differential system under consideration. As discussed in Section~\ref{sec:discApplAndLimit}, such structural information can be incorporated directly into the chosen finite representation, substantially simplifying the generator model. At the same time, the proposed framework is not intended to provide complete symmetry discovery for every class of DE systems, and we therefore make its present applicability boundaries explicit. In particular, even within the current practical scope, some systems may possess infinite-dimensional intrinsic symmetry algebras arising from arbitrary functional freedom. In the formal theory of DEs, such freedom is characterized through involutive completion, Cartan characters, and ultimately the Cartan--K\"{a}hler framework~\cite[Chapters~11 and~15]{olver1995equivalence}. A fully data-driven analogue of these involutivity tests has not yet been established here. Instead, we introduce a conservative rank-stabilization preflight (RSP) test: over a nested sequence of increasingly rich finite dictionaries, we monitor the dimension of the represented non-additive admissible symmetry space. If this dimension stabilizes persistently under successive refinements, the sequential discovery stage is allowed to proceed. If stabilization fails, the algorithm terminates without making a completeness claim for the intrinsic symmetry algebra of the supplied differential system.

Finally, the recovered generators are subjected to several complementary numerical validations. First, we test whether the admissible discovery space is exhausted once the known dimension of the benchmark intrinsic algebra has been reached. Second, we evaluate whether the discovered generators reproduce closure under the Lie-bracket operation to numerical precision. Unlike recent joint-learning approaches that impose Lie-bracket closure during symmetry discovery~\cite{liu2026liestonet,liu2026liedynnet}, closure is deliberately treated here only as a post-recovery validation. In a sequential setting, the generators recovered at an intermediate stage can span a proper subset of the target algebra that is not itself closed; enforcing closure prematurely can therefore exclude legitimate represented directions, bias the search toward a smaller closed subalgebra, or induce premature termination. \ref{app:whyNotImposingSolvability} develops the corresponding recovery-path obstruction in detail. Third, we impose a stronger global representation test: \textit{does a known analytic intrinsic symmetry generator of the benchmark system lie in the global linear span over $\mathbb{R}$ of the discovered generators throughout the sampled region?} Together, these tests assess not only whether individual candidates satisfy the sampled infinitesimal symmetry conditions, but also whether the recovered collection behaves as a coherent finite-dimensional representation of the target intrinsic Lie algebra.

Before concluding this section, we summarize the organization of the remainder of the paper. Section~\ref{sec:preliminaries} establishes the notation and mathematical prerequisites used throughout the manuscript. Readers already familiar with differential geometry, Lie groups, and Cartan's method of equivalence may skim this section primarily for notation before proceeding to the subsequent development. Section~\ref{sec:framework} presents the core of the proposed methodology. There, we formulate the optimization requirements associated with recovering the intrinsic symmetry algebra $\mathfrak{g}_{\mathrm{int}}$ and develop the operator-theoretic approach that replaces the resulting multi-objective formulation with a substantially simpler optimization problem posed within an admissible search space. Section~\ref{sec:implAndCompConsid} discusses the construction of our Kolmogorov--Arnold Network (KAN)-inspired spline-based dictionary, explains why the so-called ``nontriviality test'' must remain loss-based rather than being imposed entirely through linear space restrictions, describes the implementation and limitations of the RSP test, and establishes sequential stopping criteria concluding that the admissible represented algebra has been exhausted. Section~\ref{sec:experiments} presents the most recent numerical results of the framework, including
\begin{enumerate*}[label=\textit{(\roman*)}]
    \item the benchmark DE system and the \textit{generic random shell sampling} procedure,
    \item reconstruction accuracy of the target $\mathfrak{g}_{\mathrm{int}}$,
    \item the speed-tail ratios of the discovered generators,
    \item approximation of a known generator through a global constant-coefficient span fit, and
    \item the locality of both the infinitesimal symmetry-condition quality and the reconstruction accuracy.
\end{enumerate*}
Section~\ref{sec:discApplAndLimit} discusses the practical applicability of the framework, its present limitations, and the structural properties of DE systems underlying these boundaries. Finally, Section~\ref{sec:conclusion} summarizes the principal findings of this study and outlines the research directions that follow naturally from the current state of the proposed architecture.

\section{Mathematical Preliminaries and Problem Formulation}
\label{sec:preliminaries}
The purpose of this section is not to reiterate material already established in the literature, but rather to introduce the mathematical ingredients required to formulate the research problem. We refer the reader to the cited literature for more comprehensive treatments of the underlying theory. This section also establishes the mathematical notation used throughout the paper. Proofs of the theorems, lemmas, and propositions introduced in the manuscript are provided in \ref{app:proofs}. Furthermore, a \textquote{Table of Symbols} for reader convenience is included in \ref{app:tableOfSymbols}.

In the first subsection, we formulate a DE system within the \textit{jet formalism}, since the associated \textit{jet bundle} provides the local geometric setting in which the research problem is posed. We then introduce Lie point transformations and their prolongations to jet space, together with the relation between the corresponding Lie group of continuous transformations and its Lie algebra, identified with the tangent space at the identity element $e$. These two subsections provide the mathematical foundation for stating the central infinitesimal criterion, namely the \textit{infinitesimal symmetry condition}, in the third subsection, where the notion of a \textit{Lie point symmetry} is formally introduced together with the associated determining equations. The following subsection distinguishes a solution section of the underlying fiber bundle from its prolonged solution graph and from the equation manifold in jet space. The prolongation and projection maps are introduced there to clarify the geometric distinction between sampling a particular solution graph and sampling the equation manifold more generically. We subsequently formalize what is meant in this work by an \textit{intrinsic symmetry generator} and distinguish the intrinsic symmetry sector from solution-dependent symmetry sectors. These intrinsic generators are the smooth infinitesimal generators that constitute the principal objects to be recovered in the present study. Finally, we provide in \ref{app:geomViewPntAndFormalInteg} a concise introduction to \textit{Cartan's equivalence framework} and \textit{formal integrability}, both to
place the RSP test within its broader geometric context and to motivate the future research direction discussed in Section~\ref{sec:conclusion}. The following five subsections, however, contain all mathematical prerequisites required to understand the proposed discovery framework.

\subsection{Jet-Space Geometry and Differential Equations}
\label{subsec:deAndjetSpaceGeom}

\begin{figure}[t]
    \centering
    \includegraphics[width=0.82\columnwidth]{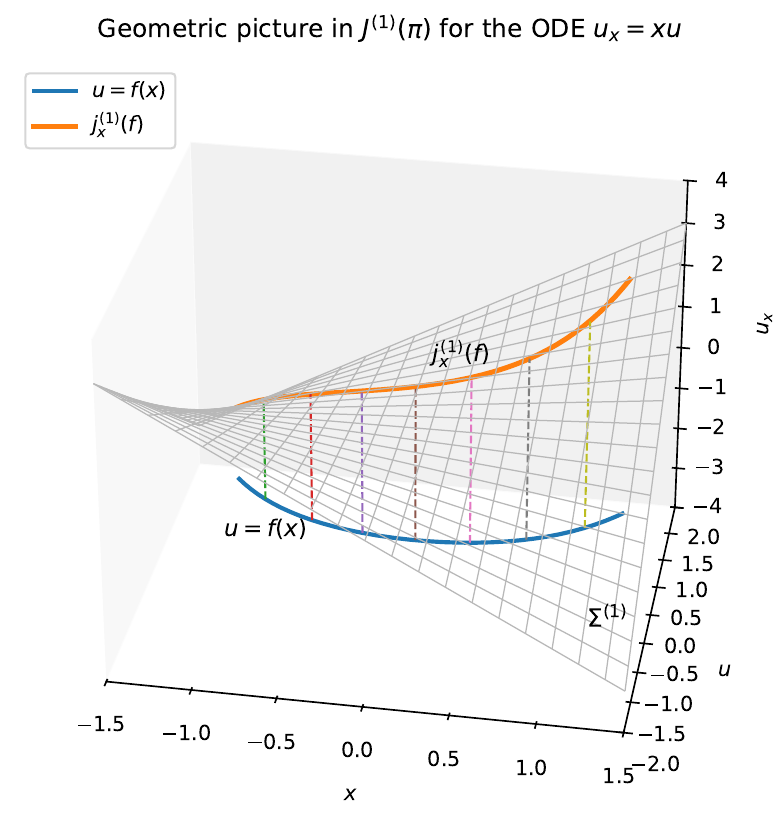}
    \caption{The equation manifold $\Delta^{(1)}=\curBarTerm{(x,u;u_x)\mid u_x-xu=0}$ in $J^{(1)}(\pi)$, together with a solution section $u=f(x)$ and its first jet lift $x\mapsto j^1 i_f(x)$.}
    \label{fig:jetSpaceODEGeometry}
\end{figure}

Let $\mathbf{X}$ denote the \textit{base space}, locally coordinatized by the independent variables $\mathbf{x}=\paraBrTerm{x^1,\dotsc,x^p}$, and let $\pi:E\to\mathbf{X}$ be a fiber bundle whose fibers are locally coordinatized by the dependent variables $\mathbf{u}=\paraBrTerm{u^1,\dotsc,u^q}$. The total space $E$ is the space in which the graphs of solutions to a given DE system are represented. Locally, $E$ admits the usual product-coordinate representation $\mathbf{X}\times\mathbf{U}$, with points denoted by the coordinate pair $(\mathbf{x},\mathbf{u})$. Accordingly, a \textit{local solution section} is represented as $\Gamma_f \coloneqq \paraBrTerm{\mathbf{x},\mathbf{f}(\mathbf{x})}$, or equivalently by $\mathbf{u}=\mathbf{f}(\mathbf{x})$. Assuming sufficient smoothness so that mixed partial derivatives commute, we denote derivatives of a dependent variable $u^\alpha$ by

\begin{equation}
    \label{eqn:derivsOfDepVar}
    u^\alpha_{x^I} \coloneqq \frac{\partial^{\vert I\vert}u^\alpha}{\partial x^{i_1}\cdots\partial x^{i_k}},
    \quad
    I=\paraBrTerm{i_1,\dotsc,i_k},
    \quad
    \vert I\vert=k,
\end{equation}
where $I$ is a multi-index specifying the independent coordinates with respect to which differentiation is performed. The corresponding quantities $f^\alpha_{x^I}$ denote the partial derivatives of the local representative $\mathbf{f}:\mathbf{X}\to\mathbf{U}$ of a solution section. These derivative variables will serve as coordinates on the jet spaces introduced below.

The formal definition of a jet bundle is based on the equivalence of local Taylor expansions, up to a prescribed order $s$, of smooth local sections at a fixed base point $\mathbf{x}\in\mathbf{X}$~\cite[Chapter~2]{Seiler2010}.
\begin{definition}[Jet bundle]
    \label{defn:jetBundle}
    Let
    \begin{equation}
        \label{eqn:taylorExpOfAFunc}
        f^\alpha(\mathbf{x}) = 
            \sum_{0\leq \vert J\vert\leq s} \frac{u^\alpha_{x^J}}{J!}\paraBrTerm{\mathbf{x}-\mathbf{x}_0}^{J} +
            R^\alpha(\mathbf{x},\mathbf{x}_0)
    \end{equation}
    denote the order-$s$ Taylor expansion of a smooth scalar component $f^\alpha$ of a local section about a fixed point $\mathbf{x}_0\in\mathbf{X}$, where
    \begin{equation}
        u^\alpha_{x^J} \coloneqq 
            \left.
                \frac{\partial^{\vert J\vert} f^\alpha}{\partial x^J}
            \right\vert_{\mathbf{x}=\mathbf{x}_0}
    \end{equation}
    are its derivative coefficients and $R^\alpha(\mathbf{x},\mathbf{x}_0)$ denotes the corresponding remainder. For a vector-valued local representative $\mathbf{f}$, let
    \begin{equation}
    \mathbf{u}_k \coloneqq \curBarTerm{\mathbf{u}_I:\vert I\vert=k},
    \quad
    \mathbf{u}_{(s)} \coloneqq \curBarTerm{\mathbf{u}_0,\mathbf{u}_1,\dotsc,\mathbf{u}_s},
    \end{equation}
    where $\mathbf{u}_0\equiv\mathbf{u}$ and $\mathbf{u}_k$ collects all derivative coordinates of total order $k$ for all dependent-variable components. 
    
    Two smooth local sections $\mathbf{f}$ and $\mathbf{g}$ are said to be equivalent to order $s$ at $\mathbf{x}_0$, denoted by $\mathbf{f}\sim_s\mathbf{g}$ at $\mathbf{x}_0$, if
    \begin{equation}
        \left.
            \frac{\partial^{\vert I\vert}f^\alpha}{\partial x^I}
        \right\vert_{\mathbf{x}=\mathbf{x}_0}
        =
        \left.
            \frac{\partial^{\vert I\vert}g^\alpha}{\partial x^I}
        \right\vert_{\mathbf{x}=\mathbf{x}_0}
    \end{equation}
    for every $\alpha=1,\dotsc,q$ and every multi-index $I$ satisfying $0\leq\vert I\vert\leq s$. Equivalently, their coordinate sets $\mathbf{u}_{(s)}$ coincide at $\mathbf{x}_0$. The corresponding equivalence class is
    \begin{equation}
        \label{eqn:equivOfMaps}
        \left. [\mathbf{f}]_{(s)} \right\vert_{\mathbf{x}_0} =
        \curBarTerm{\mathbf{g} \mid \mathbf{g}\sim_s\mathbf{f} \;\text{at}\; \mathbf{x}_0}.
    \end{equation}
    The order-$s$ jet bundle $J^{(s)}(\pi)$ is the space of all such equivalence classes over all $\mathbf{x}\in\mathbf{X}$. Each element $\left. [\mathbf{f}]_{(s)} \right\vert_{\mathbf{x}}$ is called an $s$-jet of a local section at $\mathbf{x}$.
\end{definition}

In local coordinates, $J^{(s)}(\pi)$ is coordinatized by
\begin{equation}
    \label{eqn:localJetCoords}
    z^{(s)} = \paraBrTerm{\mathbf{x}, \mathbf{u}, \mathbf{u}_I},
    \quad
    1\leq\vert I\vert\leq s.
\end{equation}
Since mixed partial derivatives are assumed to commute, $u^\alpha_{x^{i_1}\cdots x^{i_k}}$ is unchanged under permutations of the differentiation indices. We therefore retain only one coordinate for each symmetric multi-index, equivalently taking $i_1\leq\cdots\leq i_k$. The order-$k$ derivative coordinates consequently have the local structure of $S^k(T_{\mathbf{x}}^*\mathbf{X})$ for each dependent-variable component. Hence, the number of derivative coordinates of order $k$ for each component is
\begin{equation}
    \binom{p+k-1}{k},
\end{equation}
and the local dimension of the order-$s$ jet bundle~\cite[Chapter~2]{olver2000applications} is
\begin{equation}
    \label{eqn:dimJetBundle}
    \dim J^{(s)}(\pi) = p + q\binom{p+s}{s}.
\end{equation}

The jet bundle carries the canonical projections
\begin{equation}
    \label{eqn:jetCanonicalProjections}
    \pi_0^s: J^{(s)}(\pi)\to E,
    \quad
    \pi^s: J^{(s)}(\pi)\to\mathbf{X},
\end{equation}
with $\pi^s=\pi\circ\pi_0^s$. More generally, for $0\leq r<s$, the projection
\begin{equation}
    \pi_r^s: J^{(s)}(\pi)\to J^{(r)}(\pi)
\end{equation}
forgets all derivative coordinates of order greater than $r$.

Within the local representation in~\eqref{eqn:localJetCoords}, the dependent-variable and derivative coordinates are treated as independent jet-space coordinates; they are constrained by derivative relations only after restriction to a prolonged section or another submanifold carrying the corresponding differential relations, as discussed in \textit{\nameref*{subsec:sltnGrphAndGenSmpl}}. This local jet-coordinate representation is the one employed throughout the proposed computational framework.

With the jet formalism established above, we can now define an order-$s$ DE system, which is naturally represented as a subset of $J^{(s)}(\pi)$.
\begin{definition}[Differential equation system]
    \label{defn:deSystem}
    Let $z^{(s)}\in J^{(s)}(\pi)$ denote an $s$-jet and let $\Delta^{(s)}: J^{(s)}(\pi) \to \mathbb{R}^{m}$ be a smooth vector-valued differential function, whose $m$ components represent the constituent differential equations. An order-$s$ DE system is defined by the zero locus
    \begin{equation}
        \label{eqn:deSystem}
        \Sigma^{(s)} = \curBarTerm{z^{(s)}\in J^{(s)}(\pi) \mid \Delta^{(s)}\paraBrTerm{z^{(s)}}=\mathbf{0}}.
    \end{equation}
    Here, $s$ denotes the highest derivative order effectively appearing in $\Delta^{(s)}$. We refer to $\Sigma^{(s)}$ as the \textit{equation manifold} whenever the regularity condition stated below is satisfied.

    Let $M\subset\mathbf{X}$ be an open subset and let
    \begin{equation}
        i_f: M \to E,
        \quad
        \mathbf{x} \longmapsto \left. \Gamma_f \right\vert_{\mathbf{x}} = \paraBrTerm{\mathbf{x},\mathbf{f}(\mathbf{x})},
    \end{equation}
    be the smooth local section corresponding to $\Gamma_f$, satisfying $\pi\circ i_f = \operatorname{id}_{M}$. The order-$s$ jet prolongation of this section is the smooth map
    \begin{equation}
        \label{eqn:localSecProlongation}
        j^s i_f: M \to J^{(s)}(\pi),
        \quad
        \mathbf{x} \longmapsto \paraBrTerm{
            \mathbf{x},
            \mathbf{f}(\mathbf{x}),
            \mathbf{f}_I(\mathbf{x})
        },
    \end{equation}
    where $1\leq\vert I\vert\leq s$. The section $i_f$, or equivalently its local representative $\mathbf{u}=\mathbf{f}(\mathbf{x})$, is a solution of the DE system on $M$ if
    \begin{equation}
        \label{eqn:solutionCondition}
        \Delta^{(s)}\paraBrTerm{\sqBrTerm{j^s i_f}(\mathbf{x})} =\mathbf{0},
        \quad
        \forall\,\mathbf{x}\in M.
    \end{equation}
    Consequently, its prolonged solution graph,
    \begin{equation}
        \label{eqn:prolongedSolutionGraph}
        \Gamma_f^{(s)} \coloneqq \curBarTerm{\sqBrTerm{j^s i_f}(\mathbf{x})\mid\mathbf{x}\in M},
    \end{equation}
    satisfies $\Gamma_f^{(s)} \subseteq \Sigma^{(s)}$. Since $\pi^s\circ j^s i_f = \operatorname{id}_{M}$, the map $j^s i_f$ is a \textit{smooth embedding}~\cite[Chapter~4]{lee2013smooth}, and $\Gamma_f^{(s)}$ is therefore an embedded $p$-dimensional submanifold of $J^{(s)}(\pi)$. Moreover, the restricted projection $\left.\pi^s\right|_{\Gamma_f^{(s)}}: \Gamma_f^{(s)} \to M$ is a diffeomorphism whose inverse is $j^s i_f$.
\end{definition}

Throughout this work, we restrict attention to regular portions of the equation set by assuming that $\Delta^{(s)}$ has maximal rank $m$ on its zero locus $\Sigma^{(s)}$. Under this assumption, $\Sigma^{(s)}$ is a regular embedded submanifold of $J^{(s)}(\pi)$ of codimension $m$~\cite[Theorem~1.13]{olver2000applications}. Hence,
\begin{equation}
    \label{eqn:dimEqnManifold}
    \dim\Sigma^{(s)} = \dim J^{(s)}(\pi)-m = p + q\binom{p+s}{s} - m.
\end{equation}
Accordingly, samples belonging to singular or rank-deficient portions of the equation set are excluded from the present formulation. Additional nondegeneracy conditions relevant to the infinitesimal symmetry criterion are introduced in \textit{\nameref*{subsec:infSymCndAndDetEqns}}.

Figure~\ref{fig:jetSpaceODEGeometry} summarizes the geometric concepts introduced in this subsection through the example of the first-order ODE $\Delta^{(1)}\paraBrTerm{x, u, u_x} = u_x-xu = 0$. The solution section $u=f(x)$ is prolonged to the curve $x\mapsto j^1 i_f(x)$ in the first-order jet space, and, when $f$ satisfies the ODE, this prolonged solution graph lies entirely within the equation manifold $\Sigma^{(1)}$.

\subsection{Lie Point Transformations and Prolongation}
\label{subsec:LiePntTransandProlongation}
\begin{figure*}[t]
    \centering
    \includegraphics[width=0.5\textwidth]{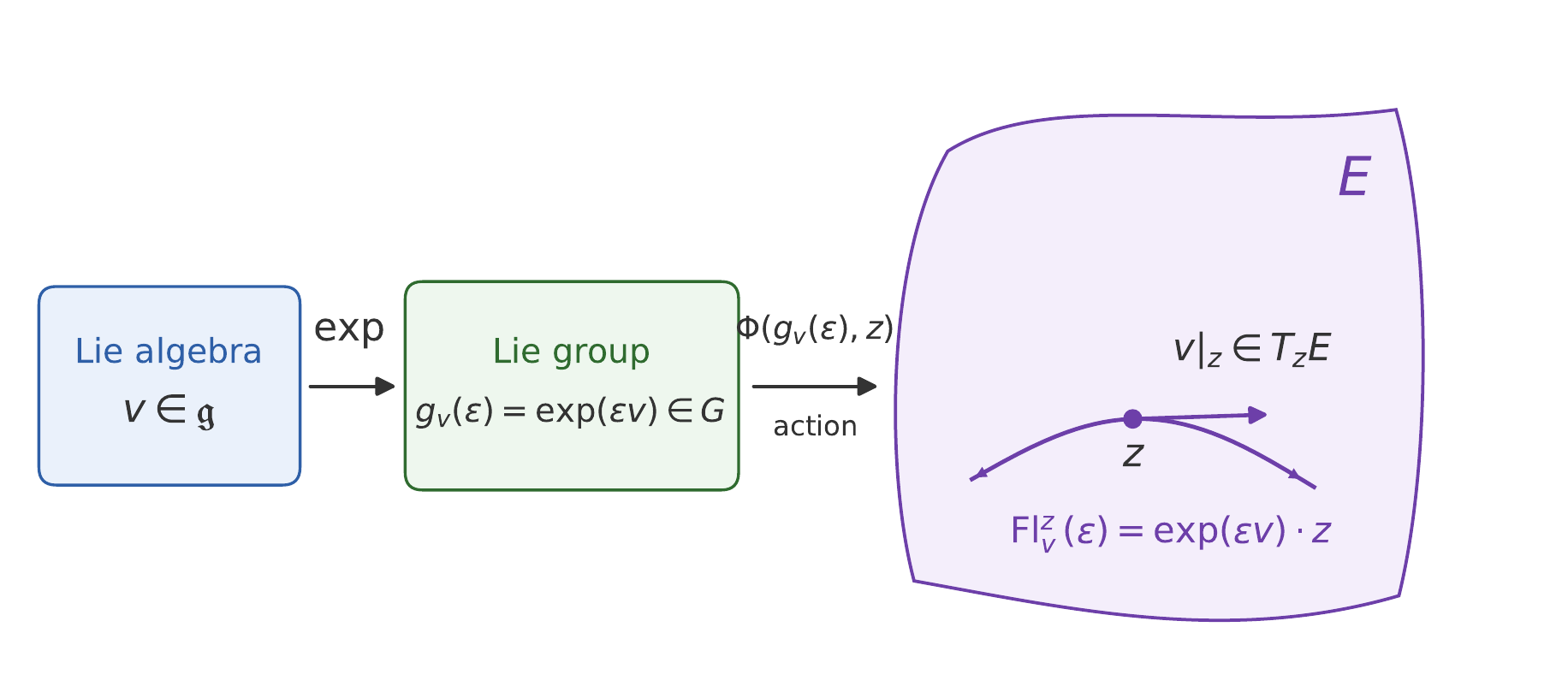}
    \caption{A schematic relation between a Lie algebra, its associated Lie group, and the induced action on $E$. The exponential map generates the one-parameter subgroup $g_v(\epsilon)=\exp(\epsilon v)\in G$, whose action produces the local flow $\operatorname{Fl}^{z}_{v}(\epsilon)=\exp(\epsilon v)\cdot z$, tangent to $v|_z\in T_zE$ at $\epsilon=0$.}
    \label{fig:LieAlgebraGrpAction}
\end{figure*}

\begin{definition}[Lie group]
    \label{defn:LieGrp}
    A \textit{differentiable manifold} is a topological space that locally admits smooth coordinate representations in Euclidean space, thereby permitting the usual tools of differential calculus to be defined intrinsically~\cite[Section~3.4]{renteln2013manifolds}. A \textit{Lie group} $G$ is a differentiable manifold endowed with a group structure for which the (left) multiplication and inversion maps
    \begin{equation}
        \label{eqn:grpOps}
        \begin{split}
            m_G &: G\times G\to G, \quad (g,h)\longmapsto g\cdot h,\\
            i_G &: G\to G, \quad g\longmapsto g^{-1},
        \end{split}
    \end{equation}
    are smooth. Here, $e\in G$ denotes the identity element, satisfying $e\cdot g=g\cdot e=g$ and $g\cdot g^{-1}=g^{-1}\cdot g=e$ for every $g\in G$.
\end{definition}

Besides multiplication on itself, a Lie group may act smoothly on another geometric space. In this work, $G$ acts on the total space $E$ through the smooth left action
\begin{equation}
    \label{eqn:grpAct}
    \Phi: G\times E\to E, \quad (g,z)\longmapsto g\cdot z,
\end{equation}
which satisfies $e\cdot z=z$ and $(g\cdot h)\cdot z=g\cdot(h\cdot z)$ for all $g,h\in G$ and $z\in E$. Such actions provide the geometric setting for Lie point transformations; for example, Olver considers the rotation group $\mathrm{SO}(2)$ acting on the $(x,u)$-space of a scalar ODE~\cite[Example~2.21]{olver2000applications}. The same abstract group may admit different \textit{realizations} through its actions on different spaces~\cite[Chapter~3, Section~II]{gilmore1974lie}. We use the symbol \textquote{$\cdot$} for both the group multiplication in $G$ and the action of $G$ on $E$ whenever the intended operation is clear from the participating objects.

For symmetry analysis of DEs, it is generally sufficient for the action to be defined only locally. Accordingly, the global action in~\eqref{eqn:grpAct} may be restricted to an appropriate open neighborhood of ${e}\times E$ in $G\times E$, yielding a \textit{local transformation group} whenever the identities of the group-action in~\eqref{eqn:grpAct} hold wherever the corresponding compositions are defined~\cite[Section~1.2]{olver2000applications}.

\begin{definition}[Lie algebra]
    \label{defn:LieAlg}
    A finite-dimensional real \textit{Lie algebra} $\mathfrak{g}$ is a real vector space equipped with a bilinear and skew-symmetric binary operation
    \begin{equation}
        \label{eqn:LieBracket}
        \sqBrTerm{\cdot,\cdot}: \mathfrak{g}\times\mathfrak{g} \longrightarrow \mathfrak{g},
        \quad
        \paraBrTerm{v_i,v_j} \longmapsto \sqBrTerm{v_i,v_j},
    \end{equation}
    called the \textit{Lie bracket}, which satisfies the Jacobi identity~\cite[Section~3.1]{hall2015lie}:
    \begin{equation}
        \label{eqn:JacobiIdentity}
        \sqBrTerm{v_i, \sqBrTerm{v_j, v_k}} + \sqBrTerm{v_j, \sqBrTerm{v_k, v_i}} + \sqBrTerm{v_k, \sqBrTerm{v_i, v_j}} = 0.
    \end{equation}

    For a Lie group $G$, we adopt the right-invariant realization of its Lie algebra. Let $R_g:G\to G$ denote right multiplication, $R_g(h)=h\cdot g$. A smooth vector field $v\in\mathfrak{X}(G)$ is \textit{right-invariant} if
    \begin{equation}
        \label{eqn:rightInvariantVF}
        d(R_g)_h
        \paraBrTerm{\left. v \right\vert_h} = \left. v \right\vert_{h\cdot g},
        \quad
        \forall\, g,h\in G.
    \end{equation}
    The space of all right-invariant vector fields on $G$ is closed under the ordinary vector-field Lie bracket and forms a finite-dimensional Lie algebra~\cite[Section~29.1.2]{hassani2013mathematical}. Evaluation at the identity, $\operatorname{ev}_e(v)=\left. v \right\vert_e$, provides a vector-space isomorphism between this space and $T_eG$; hence, we identify $\mathfrak{g}\cong T_eG$. Conversely, every tangent vector $\left. v \right\vert_e\in T_eG$ extends uniquely to a right-invariant vector field through $\left. v \right\vert_g=d(R_g)_e\paraBrTerm{\left. v \right\vert_e}$. In local coordinates $\paraBrTerm{g^1,\dotsc,g^r}$ on $G$, where $r=\dim G=\dim\mathfrak{g}$, such a vector field may be expressed as
    \begin{equation}
        \label{eqn:algEltDefn}
        \left. v_i \right\vert_g = \sum_{a=1}^{r} \eta_i^a(g) \partial_{g^a} \in T_gG.
    \end{equation}
    Thus, if $\curBarTerm{v_1,\dotsc,v_r}$ is a basis of $\mathfrak{g}$, its Lie-algebraic structure is encoded by the constants $c_{ij}^{k}\in\mathbb{R}$ defined through
    \begin{equation}
        \label{eqn:LieAlgStructConsts}
        \sqBrTerm{v_i, v_j} = \sum_{k=1}^{r} c_{ij}^{k}v_k.
    \end{equation}
\end{definition}

The smooth left action $\Phi$ in~\eqref{eqn:grpAct} induces a corresponding infinitesimal realization of $\mathfrak{g}$ in $\mathfrak{X}(E)$. For each $z\in E$, let $\Phi_z:G\to E$ denote the \textit{orbit map} $\Phi_z(g)=g\cdot z$. We identify the realization of $v\in\mathfrak{g}$ on $E$ through $\left. v \right\vert_z=d(\Phi_z)_e\paraBrTerm{\left.v\right\vert_e}$ and, when no ambiguity can arise, use the same symbol $v$ for the right-invariant vector field on $G$ and its induced vector field on $E$. In the local bundle coordinates $\paraBrTerm{\mathbf{x},\mathbf{u}}$, the latter takes the form
\begin{equation}
    \label{eqn:algEltRepOnE}
    v = \sum_{i=1}^{p} \xi^i\paraBrTerm{\mathbf{x},\mathbf{u}} \partial_{x^i} + 
        \sum_{\alpha=1}^{q} \varphi^\alpha\paraBrTerm{\mathbf{x},\mathbf{u}}\partial_{u^\alpha},
\end{equation}
where $\xi^i,\varphi^\alpha\in C^\infty(E)$. We refer to $\curBarTerm{\xi^i}$ and $\curBarTerm{\varphi^\alpha}$ as the base and fiber coefficients of $v$, respectively. The proposed framework does not reconstruct the Lie group $G$ itself; rather, it recovers the \textit{infinitesimal generators} represented as vector fields in $\mathfrak{X}(E)$. Accordingly, the local representation in~\eqref{eqn:algEltRepOnE} is the principal form used throughout this work. Since these generators act on the independent and dependent variables, rather than directly on derivative coordinates, the local transformations they generate on $E$ are \textit{Lie point transformations}; their induced action on jet space is obtained through prolongation below.

The connection between a Lie group $G$ and its Lie algebra $\mathfrak{g}$ is provided by a special map called the \textit{exponential map}~\cite[Section~1.4]{olver2000applications}:
\begin{equation}
    \label{eqn:expMap}
    \exp: \mathfrak{g} \to G,
    \quad
    v \longmapsto \exp(v).
\end{equation}
The exponential map is a local diffeomorphism from a neighborhood of $0\in\mathfrak{g}$ onto an open neighborhood of $e\in G$. Hence, every $g\in G$ sufficiently close to $e$ can be written as $g=\exp(v)$ for some $v\in\mathfrak{g}$. More generally, assuming $\mathfrak{g}=\operatorname{span}_{\mathbb{R}}\curBarTerm{v_1,\dotsc,v_r}$, every element $g$ of the connected identity component $G_0\subseteq G$ can be expressed as a finite product
\begin{equation}
    \label{eqn:anyGUsingExp}
    g = \exp\paraBrTerm{\epsilon^1v_{i_1}} \cdots \exp\paraBrTerm{\epsilon^kv_{i_k}},
\end{equation}
for some finite $k \in \mathbb{N}$, where $\epsilon^j\in\mathbb{R}$ and $1\leq i_j\leq r$ for $j=1,\dotsc,k$~\cite[(1.40)]{olver2000applications}. Thus, although $\exp:\mathfrak{g}\to G$ need not itself be globally surjective, finite products of one-parameter subgroups generated by a basis of $\mathfrak{g}$ recover the identity component $G_0$.

We exploit this correspondence in Section~\ref{sec:implAndCompConsid}, where neural ODE solvers~\cite{chen2018neural} are employed to approximate the \textit{flows} generated by elements of $\mathfrak{g}$ on $E$. For $v\in\mathfrak{g}$ and $z\in E$, the corresponding integral curve is
\begin{equation}
    \label{eqn:algFlowOnE}
    \operatorname{Fl}_v^z(\epsilon) \coloneqq \Phi_z\paraBrTerm{\exp(\epsilon v)},
    \quad
    \operatorname{Fl}_v^z: I_z\to E,
\end{equation}
where $I_z\subseteq\mathbb{R}$ is an interval containing $\epsilon=0$ on which the local action is defined. Its tangent at the identity satisfies
$\left. \nicefrac{d}{d\epsilon}\Phi_z\paraBrTerm{\exp(\epsilon v)} \right\vert_{\epsilon=0} \left.v\right\vert_z$, with $\left. v \right\vert_z$ given by~\eqref{eqn:algEltRepOnE}~\cite[Section~1.3]{olver2000applications}. Hence, $\operatorname{Fl}_v^z$ is simply the integral curve through $z$ generated by the \textit{infinitesimal action} of $v$. This correspondence between finite transformations and their infinitesimal generators yields the two equivalent formulations of the symmetry condition introduced in \textit{\nameref*{subsec:infSymCndAndDetEqns}}.

The diagram in Figure~\ref{fig:LieAlgebraGrpAction} summarizes the relationships introduced thus far. An infinitesimal generator $v\in\mathfrak{g}$ is mapped through the exponential map to a one-parameter subgroup of $G$, denoted by $g_v(\epsilon)$. The action of $g_v(\epsilon)$ on $E$, as defined in~\eqref{eqn:grpAct}, then induces the flow $\operatorname{Fl}_v^z(\epsilon)$ through each $z\in E$.

For consistency with the right-invariant realization of $\mathfrak{g}$ adopted above, we distinguish the left action of $G$ on $E$ from the inverse right action of $G$ on itself. In particular, for $g,h\in G$, let the latter be given by $h\longmapsto h\cdot g^{-1}$. A smooth map $\rho:E\to G$ is then $G$-\textit{equivariant} with respect to these two actions if
\begin{equation}
    \label{eqn:rightEquivariance}
    \rho\paraBrTerm{g\cdot z} = \rho(z)\cdot g^{-1}, 
    \quad
    g\in G,\; z\in E.
\end{equation}
Equivalently, $\rho$ intertwines the left action on $E$ with the inverse right action on $G$. This is the convention underlying the right regularization of a group action in the moving-frame framework of Fels and Olver, where the lifted action on $G\times E$ takes the form $(h,z)\longmapsto(h\cdot g^{-1},g\cdot z)$~\cite[Section~3]{fels1999moving}.

This convention also aligns the right-invariant Lie algebra on $G$ with the infinitesimal generators induced by the left action on $E$. In particular, Fels and Olver show that the correspondence between a right-invariant element $v\in\mathfrak{g}$ and its induced infinitesimal generator on the acted-upon manifold preserves the Lie-algebraic structure, while Theorem~3.10 expresses the associated group-direction differentials of the lifted invariants directly in terms of these infinitesimal-generator coefficients and the right Maurer--Cartan forms~\cite[Theorem~3.10]{fels1999moving}. We adopt the same left-action/right-invariant convention here both to avoid unnecessary sign changes in the infinitesimal realization in Section~\ref{sec:framework}, to remain compatible with the moving-coframe formulation revisited in \ref{app:geomViewPntAndFormalInteg}, and to provide a consistent foundation for the future research trajectory discussed in Section~\ref{sec:conclusion}.

Lastly, we introduce the \textit{prolongation of an infinitesimal generator} $v\in\mathfrak{X}(E)$ to a vector field on $J^{(s)}(\pi)$, in analogy with the prolongation of local sections introduced in~\eqref{eqn:localSecProlongation}. The infinitesimal action of $v$ on a local section $\Gamma_f\subset E$ must be \textquote{compatible} with the induced action on its prolonged graph $\Gamma_f^{(s)}\subset J^{(s)}(\pi)$. This compatibility underlies the general prolongation formula of Olver~\cite[Theorem~2.36]{olver2000applications}, which defines
\begin{equation}
    \label{eqn:prolongationOfGens}
    \begin{split}
        \mathrm{pr}^{(s)}(v) &\coloneqq v + \sum_{\alpha=1}^{q} \sum_{1\leq\vert J\vert\leq s} \varphi_J^\alpha\frac{\partial}{\partial u^\alpha_{x^J}}, \\
        \varphi_J^\alpha &\coloneqq D_J \paraBrTerm{\varphi^\alpha - \sum_{i=1}^{p} \xi^i u^\alpha_{x^i}} + \sum_{i=1}^{p} \xi^i u^\alpha_{x^Jx^i},
    \end{split}
\end{equation}
where $D_J=D_{x^{i_1}}\cdots D_{x^{i_{\vert J\vert}}}$ denotes the iterated total derivative associated with the symmetric multi-index $J$. For a smooth differential function $P(\mathbf{x},\mathbf{u}_{(s)})$ whose highest derivative order is $s$, the total derivative with respect to $x^i$ is
\begin{equation}
    \label{eqn:totalDerivative}
    D_{x^i}P = \frac{\partial P}{\partial x^i} +
        \sum_{\alpha=1}^{q} \sum_{0\leq\vert J\vert\leq s} u^\alpha_{x^Jx^i} \frac{\partial P}{\partial u^\alpha_{x^J}},
\end{equation}
where the case $\vert J\vert=0$ corresponds to $u^\alpha$ itself~\cite[Proposition~2.35]{olver2000applications}. Thus, $\mathrm{pr}^{(s)}(v)$ extends the infinitesimal action of $v$ from the coordinates $(\mathbf{x},\mathbf{u})$ to all derivative coordinates required through order $s$.

\subsection{Infinitesimal Symmetry Condition and Determining Equations}
\label{subsec:infSymCndAndDetEqns}
\textit{\nameref*{subsec:deAndjetSpaceGeom}} and \textit{\nameref*{subsec:LiePntTransandProlongation}} provide the mathematical machinery required to formulate the symmetry condition for a DE system.

\begin{definition}[Symmetry condition]
    \label{defn:symCnd}
    Let $z^{(s)}=\paraBrTerm{\mathbf{x},\mathbf{u}_{(s)}}$ denote an $s$-jet and let $\Delta^{(s)}(z^{(s)})=\mathbf{0}$ be an order-$s$ DE system consisting of $m$ equations. A Lie group $G$ acting locally on $E$ is a \textit{symmetry group} of the DE system if its prolonged action preserves the equation manifold $\Sigma^{(s)}$. Equivalently, for every $z^{(s)}\in\Sigma^{(s)}$ and every $g\in G$ for which the prolonged action is defined,
    \begin{equation}
        \label{eqn:symCondOverGrpAct}
        z^{(s)}\in\Sigma^{(s)} 
        \quad
        \Longrightarrow
        \quad
        g^{(s)}\cdot z^{(s)} \in \Sigma^{(s)}, 
        \quad \forall\, g\in G,
    \end{equation}
    where $g^{(s)}$ denotes the prolonged transformation induced by $g$ on $J^{(s)}(\pi)$.
    
    Under the regularity and local-solvability assumptions required by the infinitesimal invariance criterion, preservation of $\Sigma^{(s)}$ by the connected identity component $G_0$ is equivalently characterized by
    \begin{equation}
        \label{eqn:symCondOverInfiniActOnEqnMan}
        \begin{split}
            \left. \mathrm{pr}^{(s)}(v) \sqBrTerm{\Delta_\nu^{(s)}} \right\vert_{\Sigma^{(s)}} &= 0, \\
            &\nu=1,\dotsc,m, 
            \quad\forall\, v\in\mathfrak{g}, 
        \end{split}
    \end{equation}
    which is known as the \textit{infinitesimal symmetry condition}~\cite[Theorem~2.31]{olver2000applications}. Geometrically, \eqref{eqn:symCondOverInfiniActOnEqnMan} states that every prolonged infinitesimal generator $\mathrm{pr}^{(s)}(v)$ is tangent to the equation manifold $\Sigma^{(s)}$.
    
    Since $\Delta^{(s)}$ has maximal rank $m$ along $\Sigma^{(s)}$, the tangency condition can locally be expressed equivalently in the multiplier form
    \begin{equation}
        \label{eqn:symCondOverInfiniAct}
        \mathrm{pr}^{(s)}(v)\sqBrTerm{\Delta_\nu^{(s)}} = \sum_{\mu=1}^{m} F_\nu^\mu\paraBrTerm{z^{(s)}}\Delta_\mu^{(s)},
    \end{equation}
    for suitable smooth functions $F_\nu^\mu$ and $\nu=1,\dotsc,m$. Thus, the prolonged infinitesimal action preserves the differential constraints modulo the constraints themselves.
\end{definition}

The criterion in~\eqref{eqn:symCondOverInfiniActOnEqnMan} is the symmetry condition employed throughout the proposed framework. Its converse implication, namely that such infinitesimal generators integrate to a local symmetry group, requires the standard nondegeneracy and local-solvability assumptions on the DE system~\cite[Section~2.6]{olver2000applications}.

In Section~\ref{sec:introduction}, we noted that the classical approach to recovering $\mathfrak{g}$ applies symbolic-manipulation algorithms to a system known as the \textit{determining system}. After substituting the prolonged generator into~\eqref{eqn:symCondOverInfiniActOnEqnMan} and, when appropriate, eliminating dependent highest-order derivatives using the DE system itself, the resulting symbolic expressions can be split with respect to the remaining functionally independent jet coordinates $\curBarTerm{\mathbf{u}_I}$, where $1\leq\vert I\vert\leq s$. Equating the corresponding independent coefficients to zero yields an overdetermined linear homogeneous PDE system for the unknown infinitesimal coefficients $\boldsymbol{\xi}(\mathbf{x},\mathbf{u})$ and $\boldsymbol{\varphi}(\mathbf{x},\mathbf{u})$ introduced in~\eqref{eqn:algEltRepOnE}. We refer the reader to~\cite[Section~2.4]{olver2000applications} and~\cite[Section~3.2.4]{bluman2008symmetry} for representative examples in which determining systems are constructed and subsequently solved.

\subsection{Solution Graphs and Generic Sampling on the Equation Manifold}
\label{subsec:sltnGrphAndGenSmpl}
In \textit{\nameref*{subsec:deAndjetSpaceGeom}}, we introduced two geometric objects that are central to the proposed framework: the equation manifold $\Sigma^{(s)}\subset J^{(s)}(\pi)$ and the prolonged solution graph $\Gamma_f^{(s)}\subset J^{(s)}(\pi)$. Although every solution of the DE system determines a prolonged graph satisfying $\Gamma_f^{(s)}\subseteq\Sigma^{(s)}$, the two manifolds generally encode fundamentally different information. Since $\Gamma_f^{(s)}$ is parameterized only by the $p$ independent variables, $\dim\Gamma_f^{(s)}=p$. In contrast, $\Sigma^{(s)}$ retains all jet-space degrees of freedom compatible with the DE constraints and, under the maximal-rank assumption, has the dimension given in~\eqref{eqn:dimEqnManifold}. Consequently, $\Gamma_f^{(s)}$ is a proper positive-codimension submanifold of $\Sigma^{(s)}$ whenever
\begin{equation}
    \label{eqn:sltnGrpStrictIncl}
    \operatorname{codim}_{\Sigma^{(s)}}\paraBrTerm{\Gamma_f^{(s)}} = \dim\Sigma^{(s)}-p > 0.
\end{equation}
In particular, under~\eqref{eqn:sltnGrpStrictIncl}, $\Gamma_f^{(s)}\subsetneq\Sigma^{(s)}$.

A particular $\Gamma_f^{(s)}$ does not merely satisfy the DE constraints. Its jet coordinates additionally satisfy the relations imposed by the specific solution $\mathbf{f}(\mathbf{x})$; e.g. $u^\alpha=f^\alpha(\mathbf{x})$ and $u^\alpha_{x^i}=f^\alpha_{x^i}(\mathbf{x})$. These relations constitute solution-specific differential information and may artificially enlarge the apparent null space associated with the sampled infinitesimal symmetry condition in~\eqref{eqn:symCondOverInfiniActOnEqnMan}, thereby admitting superfluous infinitesimal generators. The mechanism used in the proposed framework to identify and eliminate such generators is introduced in Section~\ref{sec:framework}.

\begin{proposition}[Generic Shell Sampling]
    \label{pro:genericShellSampling}
    Let $\Sigma^{(s)}\subset J^{(s)}(\pi)$ be a regular equation manifold satisfying $\dim\Sigma^{(s)}>p$, and let $i_f:M\to E$ be any fixed local solution section of $\Delta^{(s)}=\mathbf{0}$, where $M\subset\mathbf{X}$ is open. Then, its prolonged solution graph
    \begin{equation}
    \Gamma_f^{(s)} = j^s i_f(M) \subset \Sigma^{(s)}
    \end{equation}
    is an embedded positive-codimension submanifold of $\Sigma^{(s)}$ and therefore has zero measure with respect to any smooth volume measure on $\Sigma^{(s)}$.
    
    Consequently, let
    \begin{equation}
        \label{eqn:onShellDataset}
        \mathbf{S}^{\mathrm{on}} \coloneqq \curBarTerm{z_l^{(s)}\in\Sigma^{(s)}}_{l=1}^{N_{\mathrm{on}}},
        \quad
        N_{\mathrm{on}}\geq 1,
    \end{equation}
    be a finite collection of jet-space samples drawn from a probability distribution $\mu_{\Sigma^{(s)}}$ that is absolutely continuous with respect to a smooth volume measure on $\Sigma^{(s)}$. Then, for every fixed local solution $f$,
    \begin{equation}
        \label{eqn:ensuringNonSltnSmpls}
        \mathbb{P} \paraBrTerm{\mathbf{S}^{\mathrm{on}}\subset \Gamma_f^{(s)}} = 0.
    \end{equation}
\end{proposition}

A proof of Proposition~\ref{pro:genericShellSampling} is provided in \ref{subapp:genericShellSampling}. We emphasize that the proposition does not imply that random sampling prevents samples from lying near a particular solution graph, nor does it imply uniform sampling over the family of all solutions. Rather, it states only that, under an absolutely continuous sampling law, exact concentration on any fixed positive-codimension solution graph occurs with probability zero.

\subsection{Intrinsic and Solution-Dependent Symmetry Sectors}
\label{subsec:intrAndSltnDepSymSectors}
\begin{figure*}[t]
    \centering
    \includegraphics[width=0.5\textwidth]{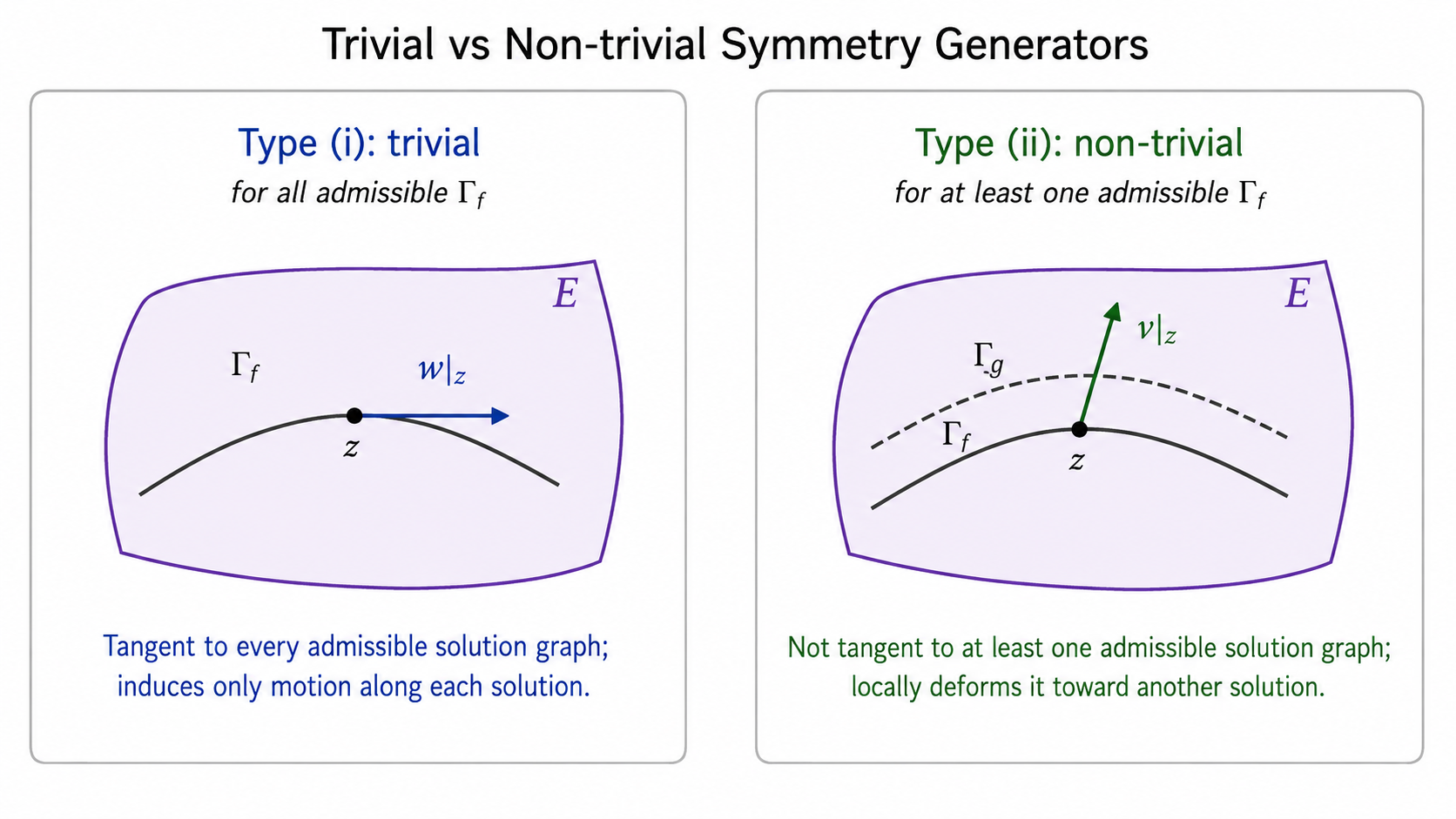}
    \caption{A trivial generator (left panel) is tangent to every admissible solution graph, whereas a nontrivial generator (right panel) maps at least one admissible solution graph to a distinct solution graph.}
    \label{fig:trivialVSnonTrivialSym}
\end{figure*}

The proposed architecture does not aim to recover the entire abstract Lie symmetry algebra $\mathfrak{g}$ of a given DE system $\Delta^{(s)}=\mathbf{0}$. Instead, we target a distinguished symmetry sector, termed the \textit{intrinsic symmetry algebra}, whose generators capture structural transformations of the DE system without importing a priori information from any particular solution. We emphasize that \textquote{intrinsic} is not a canonical classification in classical Lie symmetry theory, but terminology introduced specifically for the purposes of the present work.

\begin{definition}[Prolonged characteristic information]
    \label{defn:prolongedChar}
    Let $v\in\mathfrak{g}$ be represented in $\mathfrak{X}(E)$ as in~\eqref{eqn:algEltRepOnE}. Its \textit{characteristic vector} is defined by~\cite[Characteristics of Symmetries]{olver2000applications}
    \begin{equation}
    \label{eqn:genChar}
        \begin{split}
            \mathbf{Q}_v &= \sqBrTerm{Q_v^1,\dotsc,Q_v^q}, \\
            Q_v^\alpha\paraBrTerm{\mathbf{x},\mathbf{u}_{(1)}} &\coloneqq \varphi^\alpha - \sum_{i=1}^{p} u_{x^i}^\alpha\xi^i,
        \end{split}
    \end{equation}
    where $\alpha=1,\dotsc,q$. Thus, $\mathbf{Q}_v$ is naturally defined on $J^{(1)}(\pi)$ and, for $s\geq1$, can be evaluated on $J^{(s)}(\pi)$ through the canonical projection $\pi_1^s$.
    
    We refer to $\mathbf{E}_v^{(0)}\coloneqq\mathbf{Q}_v$ as the zeroth-order characteristic information. Motivated by the characteristic form of the prolongation formula~\cite[(2.50)--(2.51)]{olver2000applications}, we define the characteristic information available through an $s$-jet by
    \begin{equation}
        \label{eqn:genProlongedChar}
        \mathbf{E}_v^{(s-1)} \coloneqq \sqBrTerm{\mathbf{Q}_v,\, D_J\mathbf{Q}_v},
        \;\;
        1\leq\vert J\vert<s,
    \end{equation}
    where all symmetric multi-indices $J$ in the indicated range are included and $D_J$ denotes the total derivative operator introduced in~\eqref{eqn:totalDerivative}. We call $\mathbf{E}_v^{(s-1)}$ the \textit{prolonged characteristic information} of $v$ through order $s-1$.
\end{definition}

For a Lie point generator, $\mathbf{Q}_v$ depends on jet coordinates only through first order. Consequently, every $D_J\mathbf{Q}_v$ with $\vert J\vert\leq s-1$ can be evaluated using jet information available in $J^{(s)}(\pi)$. The choice in~\eqref{eqn:genProlongedChar} therefore incorporates all successive characteristic information accessible at the native order of the DE data without requiring prolongation to $J^{(s+1)}(\pi)$. Higher-order characteristic information could be introduced by prolonging the DE system and its sampled jets further, but at additional computational cost.

The characteristic $\mathbf{Q}_v$ measures the infinitesimal deformation of a solution graph modulo motion induced by reparameterization of its independent variables. In particular, $\mathbf{Q}_v$ vanishes along a prescribed solution graph precisely when the corresponding point generator is tangent to that graph. This observation motivates the following DE-level notion of triviality.

\begin{definition}[Nontriviality]
    \label{defn:nonTrivGen}
    Let $v$ be a Lie point symmetry generator with characteristic vector $\mathbf{Q}_v$. We call $v$ \textit{trivial} with respect to the DE system if
    \begin{equation}
        \label{eqn:trivGenCnd}
        \left. \mathbf{Q}_v \right\vert_{\Gamma_f^{(1)}} \equiv \mathbf{0}
    \end{equation}
    for every admissible local solution $f$. Equivalently, $v$ is tangent to every admissible local solution graph. Otherwise, $v$ is called \textit{nontrivial}. Figure~\ref{fig:trivialVSnonTrivialSym} illustrates this distinction geometrically. This definition is consistent with the classical equivalence of generalized symmetries through their evolutionary representatives and the associated notion of trivial symmetries~\cite[Section~5.1, Equivalence and Trivial Symmetries]{olver2000applications}.
    
    Moreover, if~\eqref{eqn:trivGenCnd} holds for a particular solution $f$, then all total derivatives of $\mathbf{Q}_v$ along the prolonged solution also vanish to the available order, and hence $\left.\mathbf{E}_v^{(s-1)}\right\vert_{\Gamma_f^{(s)}}\equiv\mathbf{0}$.
\end{definition}

\begin{remark}
    \label{rem:nonTrivConfusion}
    Definition~\ref{defn:nonTrivGen} does not imply that a nontrivial generator must be transverse to every admissible solution graph. A genuine nontrivial symmetry may be tangent to, and therefore have vanishing characteristic on, a particular subset of solutions. DE-level triviality requires this tangency property for the entire admissible solution family.
\end{remark}

\begin{remark}
    \label{rem:nonTrivFormalInteg}
    Definition~\ref{defn:nonTrivGen} is stated directly over the family of admissible local solutions and therefore does not require every point $z^{(s)}$ of the finite-order equation manifold $\Sigma^{(s)}$ to be realizable as the jet of an actual prolonged solution section. Under the regularity, formal-compatibility, involutivity, and local-existence assumptions discussed in \ref{app:geomViewPntAndFormalInteg}, the Cartan--K\"{a}hler framework ensures local integral manifolds through the relevant ordinary integral elements. On the corresponding solution-realizable regular locus, vanishing of $\mathbf{Q}_v$ on every admissible solution is therefore equivalent to $\left.\mathbf{Q}_v\right\vert_{\Sigma^{(s)}}\equiv\mathbf{0}$, with $\Sigma^{(s)}$ understood as the appropriately completed equation manifold. In the absence of these assumptions, Definition~\ref{defn:nonTrivGen} provides the less restrictive solution-level formulation adopted in this work.
\end{remark}

For a data-driven framework, evaluating only the zeroth-order characteristic at finitely many jet samples may provide insufficient evidence of nontriviality. At any $z^{(s)}\in\Sigma^{(s)}$,
$\left.\mathbf{E}_v^{(s-1)}\right\vert_{z^{(s)}}=\mathbf{0}$ implies $\left.\mathbf{Q}_v\right\vert_{z^{(s)}}=\mathbf{0}$ because $\mathbf{Q}_v$ is itself a component of $\mathbf{E}_v^{(s-1)}$, whereas the converse need not hold. Consider, for example,
\begin{equation}
    \label{eqn:nonTrivIllustrativeODE}
    \Delta^{(2)}(x,u_{(2)}) = u_{xx} = 0.
\end{equation}
The vector field $v=x\partial_x$ is a Lie point symmetry generator of~\eqref{eqn:nonTrivIllustrativeODE} by \eqref{eqn:symCondOverInfiniActOnEqnMan}, with characteristic $Q_v=-xu_x$. At the on-shell jet $z_0^{(2)}=(0,0,1,0)$, one obtains $\left. Q_v \right\vert_{z_0^{(2)}} = 0$, whereas $D_xQ_v=-u_x-xu_{xx}$ gives $\left.\mathbf{E}_v^{(1)} \right\vert_{z_0^{(2)}}=\sqBrTerm{0,-1}\neq\mathbf{0}$. Thus, the prolonged characteristic information can distinguish a pointwise zero of the characteristic from vanishing of the characteristic together with its available differential information.

To make this distinction explicit, define
\begin{equation}
    \label{eqn:characSets}
    \begin{split}
        \mathcal{Z}_{E}(v) &\coloneqq \curBarTerm{z^{(s)}\in\Sigma^{(s)} \mid \left. \mathbf{E}_v^{(s-1)} \right\vert_{z^{(s)}} = \mathbf{0}}, \\
        \mathcal{Z}_{Q}(v) &\coloneqq \curBarTerm{z^{(s)}\in\Sigma^{(s)} \mid \left. \mathbf{Q}_v \right\vert_{z^{(s)}} = \mathbf{0}}.
    \end{split}
\end{equation}
Since $\mathbf{Q}_v$ is contained in $\mathbf{E}_v^{(s-1)}$, it follows immediately that $\mathcal{Z}_{E}(v)\subseteq\mathcal{Z}_{Q}(v)$. Section~\ref{sec:framework} exploits this inclusion to strengthen, rather than render mathematically conclusive, the sampled optimization objective associated with \textit{nontriviality}.

Nontriviality is central to the symmetry sector targeted by the proposed framework and is also consistent with the future objective of constructing $G$-coframes from the recovered intrinsic generators and prescribed solution data, as discussed in Section~\ref{sec:conclusion}.

As discussed in Section~\ref{sec:discApplAndLimit}, certain classes of DE systems within the practical scope of the proposed framework may admit symmetry generators whose coefficients depend explicitly on particular solutions of the underlying DE system. We refer to such generators as \textit{solution-dependent} generators. We also note that \textquote{solution-dependent generator} is not a canonical term in classical Lie symmetry theory.

\begin{definition}[Solution-dependent vertical generator]
    \label{defn:sltnDepGen}
    Let $\Delta^{(s)}=\mathbf{0}$ be a DE system and let $v_{\varphi}$ be one of its Lie point symmetry generators having vanishing base coefficients; i.e.,
    \begin{equation}
    \label{eqn:fiberGen}
        v_{\varphi} = \sum_{\alpha=1}^{q} \varphi^\alpha\paraBrTerm{\mathbf{x}}\partial_{u^\alpha},
    \end{equation}
    where $\boldsymbol{\varphi}=\paraBrTerm{\varphi^1,\dotsc,\varphi^q}$ depends only on the independent coordinates $\mathbf{x}$. If $\boldsymbol{\varphi}$ itself constitutes an admissible local solution of the same DE system, equivalently,
    \begin{equation}
        \label{eqn:sltnDepCoeffCnd}
        \Gamma_{\varphi}^{(s)} \subseteq \Sigma^{(s)},
    \end{equation}
    then $v_{\varphi}$ is called a solution-dependent vertical generator. Its so-called \textquote{verticality} follows from the fact that $v_\varphi$ takes values in the vertical bundle $VE \coloneqq \ker(d\pi) \subset TE$.
\end{definition}

Such generators are excluded from the target symmetry sector for several reasons. First, in the canonical linear case, their flows act by additive translations within the solution family and therefore do not provide the structural transformations sought for the $G$-coframe construction discussed in Section~\ref{sec:conclusion}. Second, as shown in Section~\ref{sec:framework}, whenever the underlying linear PDE admits an infinite-dimensional local solution space, these generators constitute an infinite-dimensional Abelian ideal of its Lie symmetry algebra, which cannot be exhaustively represented by a finite computational architecture. Third, their construction requires a priori information about solutions of the DE system. This introduces a conceptual circularity for the proposed architecture, whose eventual purpose is precisely to exploit the recovered intrinsic symmetries to simplify and integrate the given DE system without presupposing its particular solutions. We refer collectively to such unwanted generators as the \textit{solution-dependent sector}. For linear DE systems, the corresponding family induced by the superposition principle will be termed \textit{the additive solution subalgebra} in Section~\ref{sec:framework}.

Definitions~\ref{defn:nonTrivGen} and~\ref{defn:sltnDepGen} provide the necessary ingredients to define the intrinsic symmetry algebra.

\begin{definition}[Intrinsic symmetry algebra]
    \label{defn:intrSymAlg}
    Let $\Delta^{(s)}=\mathbf{0}$ be a DE system with Lie symmetry algebra
    $\mathfrak{g}$. A Lie subalgebra
    $\mathfrak{g}_{\mathrm{int}}\leq\mathfrak{g}$ is called an
    \textit{intrinsic symmetry algebra} if every nonzero
    $v\in\mathfrak{g}_{\mathrm{int}}$
    \begin{enumerate*}[label=\textit{(\roman*)}]
        \item is nontrivial, and
        \item is solution-independent.
    \end{enumerate*}
\end{definition}

\begin{remark}
    \label{rem:mixedSltnDepGen}
    The presence of a solution-dependent vertical component does not, by itself, render the full symmetry generator solution-dependent. In particular, if $v=v_{\xi}+v_{\varphi}$ with $v_{\xi}\not\equiv\mathbf{0}$, then $v$ is not classified as solution-dependent in the sense of Definition~\ref{defn:sltnDepGen}, even if $v_{\varphi}$, considered separately, belongs to the solution-dependent vertical sector. Accordingly, the term \textit{solution-dependent generator} is reserved here for the purely vertical class characterized in Definition~\ref{defn:sltnDepGen}.
\end{remark}

\section{Data-Driven Symmetry Discovery Framework}
\label{sec:framework}
Section~\ref{sec:preliminaries} established the infinitesimal symmetry condition, nontriviality, solution independence, and the ultimate target of this work: the intrinsic symmetry algebra $\mathfrak{g}_\mathrm{int}$ introduced in Definition~\ref{defn:intrSymAlg}. This mathematical formulation must now be translated into a computational framework operating on sampled jet data.

We first formulate the requirements for recovering $\mathfrak{g}_\mathrm{int}$ as a multi-objective optimization problem. However, as discussed in the subsequent subsection, the resulting optimization becomes increasingly difficult to control as these requirements are imposed simultaneously. This motivates our \textit{operator-theoretic} approach, in which the determining condition and the subsequent admissibility restrictions are made explicit through the underlying algebraic structure. The remainder of this section develops this formulation and its sequential recovery procedure.

\subsection{Initial Multi-Objective Optimization Formulation}
\label{subsec:initialFormulation}
The initial approach was to design an architecture that is entirely optimization-based and relies on a standard neural-network parameterization. The requirements that each successively discovered generator $v_i\in\mathfrak{g}_\mathrm{int}$ must satisfy are
\begin{enumerate*}[label=\textit{(\roman*)}]
    \item the infinitesimal symmetry condition,
    \item nontriviality,
    \item solution independence, and
    \item global linear independence.
\end{enumerate*}
We formulate and discuss the corresponding optimization objectives in this order.

In Definition~\ref{defn:symCnd}, we stated what it means for a transformation group to constitute a symmetry of a given DE system $\Delta^{(s)}=0$. For computational discovery, however, we adopt the infinitesimal symmetry condition in \eqref{eqn:symCondOverInfiniActOnEqnMan} rather than attempting to learn the finite group action in \eqref{eqn:grpAct} directly. A finite group action must satisfy not only pointwise invariance, but also the nonlinear compatibility conditions associated with group composition, invertibility, and the existence of a local action on an open neighborhood of each $z\in E$. Consequently, learning such transformations directly does not naturally expose the linear structure underlying the symmetry problem.

By contrast, the infinitesimal generators form the Lie algebra $\mathfrak{g}$, which is a vector space, and the infinitesimal symmetry condition yields an overdetermined linear homogeneous determining system for the coefficient functions of each generator. This formulation therefore permits the use of ordinary calculus and, crucially for the framework developed below, linear-algebraic operations on represented generator spaces. Once a realization of the relevant Lie algebra has been recovered, its generators may be integrated through the exponential map to obtain the corresponding local one-parameter group actions, as described in \eqref{eqn:anyGUsingExp}. Thus, for the connected symmetry transformations generated locally from the identity, the infinitesimal formulation retains the group-level symmetry information while providing a substantially more tractable computational representation.

The condition \eqref{eqn:symCondOverInfiniActOnEqnMan} converts symmetry discovery into a local differential condition that can be evaluated directly on jet-space samples. Moreover, the \texttt{autograd} facility of modern ML tools is particularly convenient for computing the prolongation of a candidate generator given in \eqref{eqn:prolongationOfGens}. For an on-shell sample $z_l^{(s)}\in\mathbf{S}^{\mathrm{on}}$ (see \eqref{eqn:onShellDataset}), define the determining residual vector via \eqref{eqn:symCondOverInfiniActOnEqnMan}
\begin{equation}
\label{eqn:determiningResidual}
    \mathbf{r}_l(v) \coloneqq \paraBrTerm{
        \left. \mathrm{pr}^{(s)}(v) \sqBrTerm{\Delta_\nu^{(s)}} \right\vert_{z_l^{(s)}}
    }_{\nu=1}^{m}.
\end{equation}
A naive optimization objective is then the \textit{squared determining residual}
\begin{equation}
    \label{eqn:infSymCost}
    \mathcal{L}_{\mathrm{sym},l}^{\mathrm{raw}} \coloneqq \left\Vert \mathbf{r}_l(v) \right\Vert_2^2.
\end{equation}
Although \eqref{eqn:infSymCost} directly encodes the infinitesimal symmetry condition, our preliminary experiments showed that its raw magnitude is not geometrically balanced. In particular, it depends both on the local scale of the defining equations and distribution of the prolonged generator coefficients. We therefore introduce two normalization factors.

The first accounts for the local scale of the equation manifold $\Sigma^{(s)}$. Since
\begin{equation}
    \label{eqn:determiningResidualJac}
    \mathbf{r}_l(v) = \mathrm{J}_{\Delta^{(s)}}\paraBrTerm{z_l^{(s)}} \, \mathbf{b}\paraBrTerm{z_l^{(s)}}.
\end{equation}
where $\mathbf{b}$ denotes the coordinate representation of $\mathrm{pr}^{(s)}(v)$ in \eqref{eqn:prolongationOfGens}, the Jacobian $\mathrm{J}_{\Delta^{(s)}}$ represents the local normal information of $\Sigma^{(s)}$. We therefore use
\begin{equation}
    \label{eqn:localNormalMagOfEqnMan}
    \left\Vert \mathrm{J}_{\Delta^{(s)}}\paraBrTerm{z_l^{(s)}} \right\Vert_{\mathrm{F}}
\end{equation}
as the first normalization factor, where $\Vert\cdot\Vert_{\mathrm{F}}$ denotes the Frobenius norm. This prevents a large local scaling of the defining equations from disproportionately increasing the determining residual.

The second normalization concerns the prolonged generator itself. Suppose that any $v\in\mathfrak{g}_\mathrm{int}$, lifted to the order-$s$ jet bundle, is realized as in \eqref{eqn:prolongationOfGens} inside $\mathfrak{X}\paraBrTerm{J^{(s)}(\pi)}$. We call each subset of prolonged coefficients grouped according to its jet order $0\leq\vert J\vert\leq s$ a \textit{jet block}. At order zero, we further separate the horizontal and vertical coefficients,
\begin{equation}
    b_0^h\coloneqq\curBarTerm{\xi^i}, 
    \quad
    b_0^v\coloneqq\curBarTerm{\varphi^\alpha},
\end{equation}
while, for $\vert J\vert\geq1$, $b_{\vert J\vert}$ collects the prolonged coefficients $\varphi_J^\alpha$ of the corresponding order. Accordingly, we write
\begin{equation}
    \label{eqn:jetBlockRep}
    \mathbf{b} \coloneqq \paraBrTerm{b_k}_{k=1}^{s+2} =
    \paraBrTerm{
        b_0^h,b_0^v;
        b_{\vert J\vert=1},\ldots,b_{\vert J\vert=s}
    }.
\end{equation}

The magnitudes of these blocks may differ substantially across jet orders, causing the residual to be dominated by a particular block rather than by the geometric direction represented by the prolonged generator as a whole. To compensate for this imbalance, for each block $k$ we compute an instantaneous RMS scale at training step $t$,
\begin{equation}
    \label{eqn:blockRMSScale}
    \widetilde{s}_k^t \coloneqq \sqrt{\operatorname{mean}\sqBrTerm{\paraBrTerm{b_k^t}^2} + \epsilon},
\end{equation}
where the mean is taken over the corresponding block entries in the current sample batch and $\epsilon > 0$ is a sufficiently small numerical regularizer. The running scale is then tracked through the exponential moving average (EMA)
\begin{equation}
    \label{eqn:blockRMSScaleEMA}
    s_k^t \gets \alpha s_k^{t-1} + \paraBrTerm{1-\alpha}\widetilde{s}_k^t,
    \quad
    \alpha=0.9,
\end{equation}
with $s_k^0$ initialized from the first sample batch.

Using these running scales, we define the block-normalized magnitude of the prolonged generator at $z_l^{(s)}$ as
\begin{equation}
    \label{eqn:blockNrmMag}
    \left\Vert \left. \mathrm{pr}^{(s)}(v) \right\vert_{z_l^{(s)}}\right\Vert_{\mathrm{blk}} \coloneqq
    \sum_{k=1}^{s+2} \frac{ \left\Vert b_k\paraBrTerm{z_l^{(s)}} \right\Vert_2}{\paraBrTerm{s_k+\epsilon}\sqrt{D_k}},
\end{equation}
where $D_k\coloneqq\dim b_k$. The factor $\sqrt{D_k}$ compensates for the natural $\ell_2$-norm growth of a block with its dimension; without this correction, a higher-dimensional jet block could remain dominant even after RMS normalization.

Combining the local normal magnitude in \eqref{eqn:localNormalMagOfEqnMan} with the block-normalized generator magnitude in \eqref{eqn:blockNrmMag}, we obtain the \textit{normalized determining residual}
\begin{equation}
    \label{eqn:nrmResTerm}
    \boldsymbol{\rho}_l(v) \coloneqq
    \frac{\mathbf{r}_l(v)}{
        \left\Vert \mathrm{J}_{\Delta^{(s)}}\paraBrTerm{z_l^{(s)}} \right\Vert_{\mathrm{F}}
        \left\Vert \mathrm{pr}^{(s)}(v)\paraBrTerm{z_l^{(s)}} \right\Vert_{\mathrm{blk}}
        +\epsilon
    }.
\end{equation}
The two factors in the denominator render the residual \textit{angle-like}: the first normalizes the local normal scale of the equation manifold, while the second normalizes the magnitude of the prolonged generator in a block-balanced metric. In particular, the latter also makes the symmetry objective insensitive to an overall rescaling of a nonzero candidate generator, so reducing the magnitude of $v$ alone cannot improve the normalized determining residual.

For a sample batch $\mathbf{S}^{\mathrm{on}}_B\subset\mathbf{S}^{\mathrm{on}}$ of cardinality $B$, the infinitesimal-symmetry objective is finally taken as
\begin{equation}
    \label{eqn:nrmInfSymCost}
    \mathcal{L}_\mathrm{sym} \coloneqq \frac{1}{B} \sum_{z_l^{(s)}\in\mathbf{S}^{\mathrm{on}}_B}
    \left\Vert \boldsymbol{\rho}_l(v)\right\Vert_2^2.
\end{equation}
Thus, $\mathcal{L}_\mathrm{sym}$ penalizes the relative failure of the candidate prolonged vector field to remain tangent to the sampled equation manifold rather than the unnormalized magnitude of the determining residual itself.

Definition~\ref{defn:intrSymAlg} requires every nonzero element of $\mathfrak{g}_\mathrm{int}$ to be nontrivial in the sense of Definition~\ref{defn:nonTrivGen}. Accordingly, every candidate accepted as a new direction of the recovered intrinsic algebra must satisfy a corresponding nontriviality requirement. One practical motivation for this restriction is discussed in Section~\ref{sec:conclusion}, where the discovered intrinsic algebra is proposed for constructing a $G$-coframe on sampled data associated with a particular local solution section.

For this purpose, we use the prolonged characteristic information $\mathbf{E}_v^{(s-1)}$ introduced in \eqref{eqn:genProlongedChar}, rather than the zeroth-order characteristic $\mathbf{Q}_v$ alone. From the characteristic zero sets defined in \eqref{eqn:characSets}, we have $\mathcal{Z}_{E}(v)\subseteq\mathcal{Z}_{Q}(v)$. Therefore, $\mathbf{E}_v^{(s-1)}$ provides a stronger finite-sample discriminator: a sample at which $\mathbf{Q}_v$ vanishes may still carry nonzero higher-order characteristic information. Since $\mathbf{Q}_v$ already depends on first-order jet coordinates, evaluating $D_J\mathbf{Q}_v$ for $1\leq\vert J\vert<s$ requires successive total differentiation only through order $s-1$, while using jet coordinates no higher than the native data order $s$. Thus, the stronger nontriviality diagnostic does not require lifting the data beyond $J^{(s)}(\pi)$.

There are nevertheless two distinct numerical degeneracies to prevent. First, since $\mathfrak{g}_\mathrm{int}$ is a vector space, it necessarily contains the zero element, although Definition~\ref{defn:intrSymAlg} imposes nontriviality only on its nonzero elements. Moreover, the normalized symmetry objective in \eqref{eqn:nrmInfSymCost} is insensitive to an overall rescaling of a nonzero candidate. A separate magnitude condition is therefore required to prevent optimization from collapsing toward the \textit{zero vector field}. Second, among candidates of visible magnitude, the optimization must discourage directions whose characteristic information vanishes, or becomes negligibly small relative to the prolonged response, on the sampled equation manifold.

For a batch $\mathbf{S}^{\mathrm{on}}_B$ of cardinality $B$, let
\begin{equation}
    \label{eqn:batchGenMagnitude}
    \Vert v\Vert_B \coloneqq
    \sqrt{
        \frac{1}{B} \sum_{z_l^{(s)}\in\mathbf{S}^{\mathrm{on}}_B}
        \left\Vert \left.\mathrm{pr}^{(s)}(v)\right\vert_{z_l^{(s)}} \right\Vert_{\mathrm{blk}}^2
    }.
\end{equation}
We then impose the minimum-magnitude constraint
\begin{equation}
\label{eqn:minMagConstraint}
    c_\mathrm{min} \coloneqq \tau_\mathrm{min}-\Vert v\Vert_B \leq 0,
\end{equation}
where $\tau_\mathrm{min} > 0$. This condition forces the candidate generator to retain a numerically visible magnitude.

To detect sampled tangency to the corresponding solution structure, we define the \textit{relative prolonged-characteristic response} at $z_l^{(s)}$ by
\begin{equation}
    \label{eqn:relativeNonTrivResponse}
    \mathcal{R}_{\mathrm{ntr},l}(v) \coloneqq \frac{
        \left\Vert \left.\mathbf{E}_v^{(s-1)}\right\vert_{z_l^{(s)}} \right\Vert_2
    }{
    \left\Vert
        \left.\mathrm{pr}^{(s)}(v) \right\vert_{z_l^{(s)}}
    \right\Vert_{\mathrm{blk}}
    +
    \epsilon_\mathrm{ntr}
},
\end{equation}
where $\epsilon_\mathrm{ntr} > 0$ is a numerical regularizer. The corresponding inequality constraint is
\begin{equation}
    \label{eqn:nonTrivConstraint}
    c_{\mathrm{ntr},l} \coloneqq \tau_\mathrm{ntr} - \mathcal{R}_{\mathrm{ntr},l}(v) \leq 0,
\end{equation}
with $\tau_\mathrm{ntr}>0$. The normalization in \eqref{eqn:relativeNonTrivResponse} makes this criterion relative to the magnitude of the prolonged response, rather than imposing an absolute characteristic threshold that depends on the overall scale of $v$.

The condition \eqref{eqn:nonTrivConstraint} should be interpreted as a sampled nontriviality gate rather than a global certification of Definition~\ref{defn:nonTrivGen}. The use of $\mathbf{E}_v^{(s-1)}$ strengthens the available finite-order evidence through $\mathcal{Z}_{E}(v)\subseteq\mathcal{Z}_{Q}(v)$; nevertheless, satisfaction of a numerical threshold on finitely many sampled jets does not by itself establish the corresponding property over the entire solution-realizable equation manifold. Stronger conclusions may be drawn under additional regularity, solution-realizability, and analyticity assumptions, but these are not required by the optimization criterion employed here.

To incorporate the inequality constraints into the optimization, we use the quadratic exterior penalty
\begin{equation}
    \label{eqn:quadExtPenalty}
    \mathcal{P}\paraBrTerm{c} \coloneqq \frac{1}{2}\gamma\max\paraBrTerm{0,c}^2,
\end{equation}
where $\gamma > 0$ is increased during training according to the continuation schedule specified in Section~\ref{sec:implAndCompConsid}. The nontriviality objective over the current batch is therefore
\begin{equation}
    \label{eqn:nonTrivCost}
    \mathcal{L}_\mathrm{ntr} \coloneqq \mathcal{P}\paraBrTerm{c_\mathrm{min}} +
    \frac{1}{B} \sum_{z_l^{(s)}\in\mathbf{S}^{\mathrm{on}}_B}\mathcal{P}\paraBrTerm{c_{\mathrm{ntr},l}}.
\end{equation}

For the reasons discussed in \textit{\nameref*{subsec:intrAndSltnDepSymSectors}}, the solution-dependent sector consisting of generators of the form specified in Definition~\ref{defn:sltnDepGen} must be excluded from $\mathfrak{g}_\mathrm{int}$. Among the requirements considered in the present optimization formulation, however, our preliminary experiments showed that solution independence is the most difficult to encode reliably through a differentiable objective, as detailed in \textit{\nameref*{subsec:limitOfOptimFormulation}}.

Detecting a solution-dependent generator in the sense of Definition~\ref{defn:sltnDepGen} requires three simultaneous conditions. The candidate must be approximately vertical, its vertical coefficients must be approximately independent of the dependent coordinates $\mathbf{u}$, and those coefficients, when interpreted as a local section over $\mathbf{x}$, must approximately satisfy the original DE system. Symbolically, these conditions can be summarized as
\begin{equation}
\label{eqn:sltnIndepCnds}
    \begin{split}
        v &\approx v_\varphi,\\
        \frac{\partial\boldsymbol{\varphi}}{\partial\mathbf{u}} &\approx \mathbf{0}
        \;\Longrightarrow\;
        \boldsymbol{\varphi}\approx\boldsymbol{\varphi}\paraBrTerm{\mathbf{x}},\\
        \left.\Delta^{(s)}\right\vert_{\Gamma_\varphi^{(s)}} &\approx\mathbf{0},
        \quad
        \Gamma_\varphi^{(s)} = \sqBrTerm{j^s i_\varphi}\paraBrTerm{M}.
    \end{split}
\end{equation}
where $j^s i_\varphi$ is the order-$s$ prolongation map of a local section in~\eqref{eqn:localSecProlongation} and $M\subset X$ is an open subset. The first two conditions identify whether the candidate lies near the purely vertical, $\mathbf{u}$-independent class considered in Definition~\ref{defn:sltnDepGen}; only in that regime should satisfaction of the third condition be penalized. The requirement that $v_\varphi$ itself constitute a Lie point symmetry need not be tested again here, since it is imposed simultaneously through the infinitesimal-symmetry objective $\mathcal{L}_\mathrm{sym}$.

We implement the first two decisions through smooth sigmoid gates. For each $z_l^{(s)}\in\mathbf{S}^{\mathrm{on}}_B$, let
\begin{equation}
    \label{eqn:verticalGate}
    g_{\mathrm{ver},l} \coloneqq 
        \sigma\paraBrTerm{
            \kappa_\mathrm{ver}
            \paraBrTerm{\tau_\mathrm{ver} - \left\Vert \boldsymbol{\xi}\paraBrTerm{z_l^{(s)}}\right\Vert_2}
        }
\end{equation}
and
\begin{equation}
    \label{eqn:uIndGate}
    g_{\mathrm{uInd},l} \coloneqq
        \sigma\paraBrTerm{
            \kappa_\mathrm{uInd}
            \paraBrTerm{
                \tau_\mathrm{uInd} - 
                \left\Vert \frac{\partial\boldsymbol{\varphi}}{\partial\mathbf{u}}\paraBrTerm{z_l^{(s)}} \right\Vert_\mathrm{F}
            }
        },
\end{equation}
where $\sigma$ denotes the \textit{logistic sigmoid}, $\tau_\mathrm{ver},\tau_\mathrm{uInd } > 0$ are activation thresholds, and $\kappa_\mathrm{ver},\kappa_\mathrm{uInd} > 0$ control the corresponding gate sharpness. Hence, $g_{\mathrm{ver},l}\approx1$ when the horizontal component is negligible, while $g_{\mathrm{uInd},l}\approx1$ when the vertical coefficients exhibit negligible dependence on $\mathbf{u}$.

Once both gates are active, we test whether $\boldsymbol{\varphi}$ behaves as a local solution section. Let
\begin{equation}
    \label{eqn:relativeSltnDepResponse}
    \mathcal{R}_{\mathrm{dep},l}(v) \coloneqq
        \frac{
            \left\Vert \Delta^{(s)}\paraBrTerm{\sqBrTerm{j^s i_\varphi}\paraBrTerm{\mathbf{x}_l}} \right\Vert_2
        }{
            \left\Vert \left. \mathrm{pr}^{(s)}(v) \right\vert_{z_l^{(s)}} \right\Vert_{\mathrm{blk}} 
            + \epsilon_\mathrm{dep}
        },
\end{equation}
where $\epsilon_\mathrm{dep} > 0$ is a numerical regularizer. A small value of $\mathcal{R}_{\mathrm{dep},l}$ indicates that the vertical coefficient, interpreted as an approximately $\mathbf{u}$-independent section, lies close to the solution manifold of the DE system relative to the magnitude of the candidate prolonged response.

The corresponding solution-dependence constraint is
\begin{equation}
    \label{eqn:sltnIndepConstraint}
    c_{\mathrm{dep},l} \coloneqq 
        g_{\mathrm{ver},l} 
        g_{\mathrm{uInd},l} 
        \paraBrTerm{\tau_\mathrm{dep} - \mathcal{R}_{\mathrm{dep},l}(v)}
    \leq0,
\end{equation}
where $\tau_\mathrm{dep} > 0$. Thus, when either verticality or $\mathbf{u}$-independence is absent, the corresponding gate suppresses the constraint. When both gates are active, however, a sufficiently small DE residual yields $c_{\mathrm{dep},l}>0$ and the candidate is repelled from the detected solution-dependent sector.

Using the quadratic exterior penalty in \eqref{eqn:quadExtPenalty}, the solution-dependence loss over a batch of cardinality $B$ is
\begin{equation}
\label{eqn:sltnIndepCost}
    \mathcal{L}_\mathrm{dep} \coloneqq \frac{1}{B} \sum_{z_l^{(s)}\in\mathbf{S}^{\mathrm{on}}_B}
    \mathcal{P}\paraBrTerm{c_{\mathrm{dep},l}}.
\end{equation}

\begin{remark}
\label{rem:mixedSltnDepGenNotViolated}
    The construction above is consistent with Remark~\ref{rem:mixedSltnDepGen}. In particular, the general definition of a solution-dependent generator is restricted to the purely vertical class. Accordingly, a mixed generator with $\boldsymbol{\xi}\neq\mathbf{0}$ is not classified as solution-dependent merely because its vertical component is $\mathbf{u}$-independent and satisfies the DE system. Up to the numerical tolerance encoded by $g_{\mathrm{ver},l}$, the verticality gate suppresses $\mathcal{L}_\mathrm{dep}$ for such mixed generators.
\end{remark}

The final requirement differs fundamentally from the previously defined objectives because it is not a property of a single candidate in isolation. A finite-dimensional real Lie algebra is a vector space over the field of real constants, not over the ring of smooth functions (see Definition~\ref{defn:LieAlg}). Therefore, for any $w\in\mathfrak{g}$,
\begin{equation}
    \label{eqn:anyGenRepInLieAlg}
    w = \sum_{i=1}^{r} c^i v_i,
    \quad
    \mathbf{C}\coloneqq\sqBrTerm{c^1,\dotsc,c^r}\in\mathbb{R}^r,
\end{equation}
where
\begin{equation}
    \label{eqn:spanOfLieAlg}
    \mathfrak{g} = \operatorname{span}_{\mathbb{R}}\curBarTerm{v_1,\dotsc,v_r},
    \quad
    \dim\mathfrak{g}=r.
\end{equation}
Equivalently, for every $z\in E$,
\begin{equation}
    \label{eqn:anyGenRepInLieAlgEval}
    \left.w\right\vert_z = \sum_{i=1}^{r} c^i \left.v_i\right\vert_z,
\end{equation}
with the same coefficient vector $\mathbf{C}$ holding over the entire domain. Hence, linear dependence in $\mathfrak{g}$ cannot be established from an arbitrary pointwise relation whose coefficients are allowed to vary with $z$. The corresponding discovery objective must instead test whether one common vector of real constants explains the candidate over a sufficiently large collection of evaluation points.

Global linear independence is an ambient vector-field property of the realized generators in $\mathfrak{X}(E)$ and should therefore not be inferred solely from their values over the constrained sample distribution used to enforce the symmetry condition. In the present optimization formulation, the independence test is evaluated from the zeroth-order generator coefficients $\paraBrTerm{\boldsymbol{\xi},\boldsymbol{\varphi}}$, rather than from their prolonged responses.

Let $\pi_0^s:J^{(s)}(\pi)\to E$ denote the canonical projection onto the zeroth-order bundle. Depending on the structure of the DE system and the projection used to construct $\mathbf{S}^{\mathrm{on}}$, two situations may arise. First,
\begin{equation}
    \pi_0^s\paraBrTerm{\Sigma^{(s)}} \subsetneq E,
\end{equation}
in which case the on-shell samples probe only a proper subset of the ambient space on which the realized vector fields are defined. Second,
\begin{equation}
    \pi_0^s\paraBrTerm{\Sigma^{(s)}} = E,
\end{equation}
in which case the zeroth-order coordinates are not geometrically restricted to a proper subset of $E$. Even in the latter case, however, projection onto $\Sigma^{(s)}$ may induce correlations or nonuniform sampling among the zeroth-order coordinates in a finite dataset. For example, whenever the DE system locally determines a dependent coordinate in the form $u^\alpha=F^\alpha\paraBrTerm{\mathbf{x},\mathbf{u}_I}$ with $0 \leq \vert I \vert \leq s$, the corresponding on-shell values of $u^\alpha$ are no longer sampled independently of the remaining jet coordinates after projection. Consequently, restricting the independence test to the finite on-shell sample cloud may produce sample-induced rank loss or poor conditioning, whereby globally distinct vector fields exhibit dependent, or nearly dependent, responses over the sampled region.

To reduce this degeneracy, we construct an auxiliary off-shell dataset
\begin{equation}
    \label{eqn:offShellDataset}
    \mathbf{S}^{\mathrm{off}}\coloneqq \curBarTerm{
        z_l^{(s)} \notin \Sigma^{(s)} \mid
        \left\Vert \Delta^{(s)}\paraBrTerm{z_l^{(s)}} \right\Vert_2 > \epsilon_{\mathrm{off}}
    },
\end{equation}
where $\epsilon_{\mathrm{off}}>0$ and $l=1,\dotsc,N_{\mathrm{off}}=N_{\mathrm{on}}$. Each off-shell sample is generated randomly within a prescribed neighborhood of its corresponding originator sample from $\mathbf{S}^{\mathrm{on}}$ defined in \eqref{eqn:onShellDataset}, subject to the condition in \eqref{eqn:offShellDataset}. In particular, the fiber coordinates are perturbed away from their on-shell values, thereby relaxing correlations that may have been induced by projection onto the equation manifold. The auxiliary samples therefore remain computationally tied to the data-supported region while probing a less constrained neighborhood in the ambient jet space.

From $\mathbf{S}^{\mathrm{off}}$, we then select a sufficiently large fixed subset
\begin{equation}
    \label{eqn:anchorDataset}
    \mathbf{S}^\mathrm{off}_{\mathrm{anc}} \subset \mathbf{S}^{\mathrm{off}},
    \quad
    A_\mathrm{off} \coloneqq \left\vert \mathbf{S}^\mathrm{off}_{\mathrm{anc}}\right\vert,
\end{equation}
which we call the \textit{off-shell anchor dataset}. Although the anchor samples themselves belong to $J^{(s)}(\pi)$, the independence objective evaluates only the corresponding zeroth-order generator responses in $TE$. The same anchor set is reused when comparing a candidate with the generators already discovered, so that the resulting independence test probes one common constant-coefficient relation over the entire sampled anchor domain rather than allowing unrelated pointwise dependencies. Off-shell anchoring therefore provides a more robust finite-sample surrogate for ambient linear independence while remaining localized to the neighborhood of the observed equation manifold.

Starting with the second discovered generator $v_2$, global linear independence is enforced by separating a candidate $v_i$, for $i>1$, into a component represented by the previously discovered generators and a residual component outside that represented span. Accordingly, we write
\begin{equation}
    \label{eqn:decomposeVi}
    v_i = v_{\perp,i} + \underbrace{\sum_{j=1}^{i-1} c_i^j v_j}_{v_{\parallel,i}},
    \quad
    c_i^j\in\mathbb{R},
\end{equation}
where $v_{\parallel, i}\in\operatorname{span}_{\mathbb{R}}\curBarTerm{v_1,\dotsc,v_{i-1}}$ denotes the component already represented by the previously discovered generators, while
\begin{equation}
    \label{eqn:novelResidual}
    v_{\perp,i} \coloneqq v_i-v_{\parallel,i}
\end{equation}
denotes the residual component associated with the candidate. The constant coefficient vector determining $v_{\parallel,i}$ is obtained globally over the off-shell anchor dataset and is specified below.

Since the independence criterion is evaluated through the zeroth-order generator responses, we introduce the anchor-induced norm
\begin{equation}
    \label{eqn:anchorInducedNorm}
    \Vert v\Vert_{\mathrm{anc}}^2 \coloneqq
    \frac{1}{A_\mathrm{off}}
    \sum_{z_a^{(s)}\in\mathbf{S}_{\mathrm{anc}}^\mathrm{off}}
    \left\Vert \left.v\right\vert_{z_a} \right\Vert_2^2,
\end{equation}
where $z_a\coloneqq\pi_0^s\paraBrTerm{z_a^{(s)}}\in E$ denotes the zeroth-order projection of an anchor jet. Thus, a single norm aggregates the response of a generator over the entire fixed anchor domain rather than assessing independence pointwise.

Using \eqref{eqn:anchorInducedNorm}, we define the \textit{relative novelty response} of $v_i$ as
\begin{equation}
    \label{eqn:relativeNoveltyResp}
    \mathcal{R}_{\mathrm{nov}}\paraBrTerm{v_i} \coloneqq
    \frac{\Vert v_{\perp,i}\Vert_{\mathrm{anc}}}{\Vert v_i\Vert_{\mathrm{anc}} + \epsilon_{\mathrm{nov}}},
    \quad
    i>1,
\end{equation}
where $\epsilon_{\mathrm{nov}}>0$ is a numerical regularizer. A small value of $\mathcal{R}_{\mathrm{nov}}\paraBrTerm{v_i}$ indicates that the candidate is largely represented by the real span of the previously discovered generators, whereas a larger value indicates the presence of a substantial globally novel component over the anchor domain.

The corresponding linear-independence constraint is
\begin{equation}
    \label{eqn:linIndConstraint}
    c_{\mathrm{ind}} \coloneqq \tau_{\mathrm{nov}} - \mathcal{R}_{\mathrm{nov}}\paraBrTerm{v_i} \leq0,
\end{equation}
where $\tau_{\mathrm{nov}}>0$ specifies the minimum required \textit{relative novelty}. Applying the quadratic exterior penalty in \eqref{eqn:quadExtPenalty} yields
\begin{equation}
    \label{eqn:linIndCost}
    \mathcal{L}_{\mathrm{ind}} \coloneqq \mathcal{P}\paraBrTerm{c_{\mathrm{ind}}}.
\end{equation}
Hence, $\mathcal{L}_{\mathrm{ind}}$ penalizes candidates whose globally evaluated response is insufficiently distinct from the span already represented by the previously discovered generators.

To determine the constant coefficient vector $\mathbf{C}_i\coloneqq\sqBrTerm{c_i^1,\dotsc,c_i^{i-1}}\in\mathbb{R}^{i-1}$ in \eqref{eqn:decomposeVi} in a \textquote{global} manner, we employ a single least-squares projection over the complete off-shell anchor dataset $\mathbf{S}^\mathrm{off}_{\mathrm{anc}}$. For each previously discovered generator $v_j$, define its stacked anchor response vector as
\begin{equation}
    \label{eqn:oldAnchorRespVec}
    Y_j \coloneqq \operatorname{vec}\paraBrTerm{
        \paraBrTerm{
            \left.v_j\right\vert_{z_a}}_{z_a^{(s)} \in \mathbf{S}^\mathrm{off}_{\mathrm{anc}}
        }
    }
    \in
    \mathbb{R}^{A_\mathrm{off}D},
\end{equation}
where $z_a\coloneqq\pi_0^s\paraBrTerm{z_a^{(s)}}\in E$ and $D=\dim E$. The \textit{old anchor response matrix} is then
\begin{equation}
    \label{eqn:oldRespMatrix}
    R_i \coloneqq
    \begin{bmatrix}
        \,|\, &        & \,|\, \\
        Y_1 & \cdots & Y_{i-1} \\
        \,|\, &        & \,|\,
    \end{bmatrix}
    \in \mathbb{R}^{A_\mathrm{off}D\times(i-1)}.
\end{equation}
Each column of $R_i$ therefore represents one previously discovered generator over the entire off-shell anchor domain. Similarly, let
\begin{equation}
    \label{eqn:newAnchorRespVec}
    Y_i \coloneqq \operatorname{vec}\paraBrTerm{
        \paraBrTerm{\left.v_i\right\vert_{z_a}}_{z_a^{(s)} \in \mathbf{S}^\mathrm{off}_{\mathrm{anc}}}
    }
    \in \mathbb{R}^{A_\mathrm{off}D}
\end{equation}
denote the stacked anchor response of the candidate generator under training.

The global coefficient vector is obtained from
\begin{equation}
    \label{eqn:leastSqSltnOfLinIndCoeffs}
    \mathbf{C}_i \coloneqq
    \underset{\mathbf{C}\in\mathbb{R}^{i-1}}{\operatorname{argmin}}
    \left\Vert Y_i-R_i\mathbf{C} \right\Vert_2^2 = R_i^\dagger Y_i,
\end{equation}
where $R_i^\dagger$ denotes the \textit{Moore--Penrose pseudoinverse}. If $R_i$ has full column rank, \eqref{eqn:leastSqSltnOfLinIndCoeffs} reduces to
\begin{equation}
    \mathbf{C}_i = \paraBrTerm{R_i^\top R_i}^{-1} R_i^\top Y_i.
\end{equation}
Thus, $R_i\mathbf{C}_i$ is the least-squares projection of $Y_i$ onto the column space of $R_i$, which represents the sampled real span of the previously discovered generators. The corresponding residual
\begin{equation}
    \label{eqn:complAnchorRespVec}
    Y_{\perp,i} \coloneqq Y_i-R_i\mathbf{C}_i = \paraBrTerm{I-R_iR_i^\dagger}Y_i
\end{equation}
is the anchor-space representation of the complementary component $v_{\perp,i}$ introduced in \eqref{eqn:decomposeVi}. In particular, the anchor-induced norms in \eqref{eqn:anchorInducedNorm} satisfy
\begin{equation}
    \begin{split}
        \left\Vert v_i\right\Vert_{\mathrm{anc}} &= \frac{1}{\sqrt{A_\mathrm{off}}} \left\Vert Y_i\right\Vert_2, \\
        \left\Vert v_{\perp,i}\right\Vert_{\mathrm{anc}} &= \frac{1}{\sqrt{A_\mathrm{off}}} \left\Vert Y_{\perp,i}\right\Vert_2.
    \end{split}
\end{equation}

For every $i>1$, the remaining optimization requirements are imposed on the complementary component $v_{\perp,i}$ rather than directly on the raw candidate $v_i$. Indeed, $v_{\parallel,i}$ belongs to the real span of the previously accepted generators and therefore represents information already recovered by the framework. Allowing this component to contribute to the remaining objectives could mask a deficiency of the genuinely novel direction. Accordingly, the normalized determining residual $\boldsymbol{\rho}_l(v)$ in \eqref{eqn:nrmResTerm}, the minimum-magnitude and relative nontriviality constraints defined through \eqref{eqn:minMagConstraint} and \eqref{eqn:relativeNonTrivResponse}, the relative solution-dependence response in \eqref{eqn:relativeSltnDepResponse}, and the gates in \eqref{eqn:verticalGate} and \eqref{eqn:uIndGate} are evaluated using $v_{\perp,i}$ for $i>1$. For the first discovered generator $v_1$, no previously recovered subspace exists and hence the raw realization is used directly.

\subsection{Limitations of the Optimization Formulation}
\label{subsec:limitOfOptimFormulation}
Over several years, we attempted to impose all requirements introduced in \textit{\nameref*{subsec:initialFormulation}} simultaneously through different multi-objective optimization strategies. In practice, the resulting optimization became increasingly difficult to control. We summarize below the principal difficulties encountered, both to motivate the transition to the operator-theoretic formulation and to explain the need for the admissible restrictions introduced by the new methodology.

The most fundamental difficulty originates from the symmetry objective itself. Let a candidate generator be represented by a finite-dimensional neural-network parameter vector $\boldsymbol{\Theta}\in\mathcal{P}$ through the realization map
\begin{equation}
    \label{eqn:paramEvalMap}
    \Psi: \mathcal{P} \to \mathfrak{X}(E),
    \quad
    \boldsymbol{\Theta} \xmapsto{\mathrm{ev}} v_{\boldsymbol{\Theta}}.
\end{equation}
For a standard neural-network parameterization, $\Psi$ is nonlinear in $\boldsymbol{\Theta}$. Let $\mathcal{D}_{\Delta}$ denote the determining operator associated with the infinitesimal symmetry condition in \eqref{eqn:symCondOverInfiniAct}. Although $\mathcal{D}_{\Delta}$ is linear in the generator $v_{\boldsymbol{\Theta}}$ (see \textit{\nameref*{subsec:operatorTheoreticSymRep}}), the composition $\mathcal{D}_{\Delta}\circ\Psi$ is generally nonlinear in the trainable parameters. Consequently, the set
\begin{equation}
    \label{eqn:kerOfDeterminingOperator}
    \curBarTerm{\boldsymbol{\Theta}\in\mathcal{P} \mid \mathcal{D}_{\Delta} \sqBrTerm{v_{\boldsymbol{\Theta}}} = \mathbf{0}}
\end{equation}
does not, in general, constitute a fixed linear subspace of the parameter space.

The same issue appears after sampling. Let $\widehat{\mathcal{D}}_{\Delta}$ denote the determining operator evaluated over the finite on-shell dataset $\mathbf{S}^\mathrm{on}$. Its composition with the neural realization has the parameter-dependent Jacobian
\begin{equation}
    \mathrm{J}_{\boldsymbol{\Theta}}\sqBrTerm{\widehat{\mathcal{D}}_{\Delta}\circ\Psi}\paraBrTerm{\boldsymbol{\Theta}},
\end{equation}
whose row and null spaces vary as $\boldsymbol{\Theta}$ changes during training. Hence, an singular value decomposition (SVD) of this Jacobian characterizes only the local linearization of the sampled determining residual around the current parameter iterate; it does not expose a fixed symmetry nullspace that can be isolated once and preserved throughout optimization.

This behavior was also observed empirically in our preliminary experiments with $\mathcal{L}_{\mathrm{sym}}$ in \eqref{eqn:nrmInfSymCost}. SVD analysis of the sampled residual Jacobian revealed weakly constrained right-singular directions associated with small singular values, typically with $\sigma_i^2$ approaching $10^{-6}$ and below. These directions were difficult to suppress consistently during gradient-based training and produced persistent row-space contamination, whereby components outside the desired local nullspace remained mixed with the learned representation. Thus, even before the remaining objectives were imposed, the nonlinear parameterization prevented the determining condition from being treated as a \textit{fixed linear-algebraic admissibility constraint}.

A second difficulty arises from the direct competition between the infinitesimal-symmetry objective $\mathcal{L}_{\mathrm{sym}}$ and the global-independence objective $\mathcal{L}_{\mathrm{ind}}$. Once several generators have been discovered, their span already constitutes a region of the represented function space in which the determining residual is small. Consequently, gradient descent on $\mathcal{L}_{\mathrm{sym}}$ tends to favor directions close to the previously recovered subspace. In contrast, $\mathcal{L}_{\mathrm{ind}}$ explicitly penalizes such directions and pushes the new candidate toward a complementary component. The two objectives therefore exert competing pressures on the same trainable realization.

Projecting the remaining objectives onto $v_{\perp,i}$, as described in the final paragraph of \textit{\nameref*{subsec:initialFormulation}}, alleviates but does not eliminate this conflict. Since the determining operator is linear in the generator,
\begin{equation}
    \label{eqn:projDeterminingResidual}
    \mathcal{D}_{\Delta}\sqBrTerm{v_{\perp,i}} = \mathcal{D}_{\Delta}\sqBrTerm{v_i} - \sum_{j=1}^{i-1} c_i^j\mathcal{D}_{\Delta}\sqBrTerm{v_j}.
\end{equation}
Hence, if the previously discovered generators satisfy the infinitesimal symmetry condition only approximately, their residual errors are inherited by the determining residual of $v_{\perp,i}$. The same observation applies after finite sampling to $\widehat{\mathcal{D}}_{\Delta}$. Therefore, inaccuracies accumulated during earlier discovery stages propagate into the optimization of subsequent generators rather than remaining confined to the generators in which they originated.

Moreover, the coefficients $\mathbf{C}_i$ defining $v_{\parallel,i}$ are themselves recomputed from the current realization of $v_i$ through the global least-squares projection over $\mathbf{S}^{\mathrm{off}}_{\mathrm{anc}}$. The complementary component presented to the remaining objectives is therefore not fixed during training, but changes together with the candidate realization. This additional coupling further complicates the simultaneous balancing of symmetry and novelty.

As a result, sequential discovery under the multi-objective formulation becomes progressively more sensitive to residual errors and optimization history as the number of recovered generators increases. A candidate must simultaneously approach the symmetry sector while being repelled from the portion of that sector already represented by the previously discovered generators, with neither requirement imposed as an exact subspace restriction.

A third difficulty concerns the solution-dependence objective $\mathcal{L}_\mathrm{dep}$. As established formally in \textit{\nameref*{subsec:intrAdmConstraints}}, a linear DE system possesses an additive solution sector $\mathcal{A}_\mathrm{sol}\subseteq\mathfrak{g}$, which may be infinite dimensional. This sector extends the solution-dependent behavior discussed in \textit{\nameref*{subsec:intrAndSltnDepSymSectors}} and provides a potentially large family of directions that satisfy the infinitesimal symmetry condition exactly. Consequently, $\mathcal{L}_\mathrm{sym}$ alone has no mechanism for distinguishing these directions from the finite-dimensional intrinsic sector sought by the framework, described in \textit{\nameref*{subsec:intrAndSltnDepSymSectors}}.

The gated penalty $\mathcal{L}_\mathrm{dep}$ attempts to repel the optimizer from this prohibited sector only when both the verticality and $\mathbf{u}$-independence detectors in \eqref{eqn:verticalGate} and \eqref{eqn:uIndGate} are sufficiently active. Whenever either gate becomes small, the solution-dependence penalty is correspondingly suppressed, while $\mathcal{L}_\mathrm{sym}$ continues to favor any direction that reduces the determining residual. For a large, and particularly an infinite-dimensional, additive solution sector, this creates numerous admissible symmetry directions that cannot be reliably excluded through a finite collection of smooth gates and scalar penalties.

The resulting optimization may therefore alternate between approaching directions associated with $\mathcal{A}_\mathrm{sol}$ under the symmetry objective and being repelled from them once the solution-dependence gates become active. This \textit{attraction--repulsion behavior} makes $\mathcal{L}_\mathrm{dep}$ highly sensitive to gate thresholds, sharpness parameters, relative loss weights, and the current optimization trajectory. More fundamentally, it attempts to exclude an entire symmetry sector through local penalty shaping rather than removing that sector from the admissible search space itself.

The nontriviality objective introduces a further numerical complication. Apart from the regularization term $\epsilon_\mathrm{ntr}$, the relative response $\mathcal{R}_\mathrm{ntr}$ in \eqref{eqn:relativeNonTrivResponse} is scale invariant because both $\mathbf{E}_v^{(s-1)}$ and $\mathrm{pr}^{(s)}(v)$ depend linearly on $v$. Consequently, reducing the magnitude of a candidate does not necessarily reduce its relative nontriviality response as long as the candidate remains above the numerical regularization floor. Thus, $\mathcal{R}_\mathrm{ntr}$ alone cannot ensure that the complementary component $v_{\perp,i}$ remains numerically visible, which necessitates the independent minimum-magnitude constraint $c_\mathrm{min}$.

Near the zero-response regime, however, the normalized ratio becomes increasingly sensitive to the numerical regularizer and small fluctuations in the numerator and denominator. In practice, this required a smooth activation of the nontriviality constraint so that $c_\mathrm{ntr}$ was not enforced aggressively before the candidate acquired sufficient magnitude. Hence, even the sampled nontriviality requirement could not be represented by a single stable objective, but instead required coupled magnitude and activation mechanisms whose behavior depended on additional numerical thresholds.

Finally, the determining equations themselves contain structural information that is largely discarded when they are represented only through the scalar symmetry loss $\mathcal{L}_\mathrm{sym}$. A generic nonlinear optimizer is not informed of restrictions on the admissible functional form of the infinitesimal coefficients, nor of possible decompositions of the represented symmetry space. It therefore tends to converge first toward symmetry directions that are comparatively easy to represent and optimize. Once such directions have been recovered and excluded by the novelty objective, optimization of the remaining directions may become progressively more difficult or stall altogether. For example, in the benchmark problem of Section~\ref{sec:experiments}, our preliminary optimization-based experiments repeatedly recovered four generators whose real span coincided with the empirically \textit{easy} symmetry subspace
\begin{equation}
    \label{eqn:easySymSubspace}
    \mathfrak{g}_\mathrm{easy} \coloneqq \operatorname{span}_{\mathbb{R}} \curBarTerm{
        \partial_x,
        \partial_y,
        \partial_t,
        u\partial_u
    }.
\end{equation}
Subsequent optimization attempts failed to produce a fifth globally independent direction: the learned candidates remained within, or converged back toward, $\mathfrak{g}_\mathrm{easy}$ despite the action of $\mathcal{L}_\mathrm{ind}$. Here, the qualifier \textit{easy} refers only to the empirical accessibility of this subspace under the original optimization dynamics and does not indicate an intrinsic algebraic classification.

This behavior further motivates treating the determining system as an operator carrying exploitable algebraic structure, rather than merely as a residual to be minimized. Structural restrictions implied by particular classes of DEs are discussed later in the context of applicability and scope in Section~\ref{sec:discApplAndLimit}.

\subsection{Operator-Theoretic Symmetry Representation}
\label{subsec:operatorTheoreticSymRep}
Starting with the present subsection, we develop the new operator-theoretic approach used throughout the remainder of this section. The first objective is to replace the moving local nullspace encountered in \textit{\nameref*{subsec:limitOfOptimFormulation}} by a fixed represented nullspace of the determining operator. This nullspace constitutes the first admissible restriction on the search space from which $\mathfrak{g}_\mathrm{int}$ will subsequently be recovered. Its dimension, however, remains dependent on the expressive space supplied by the chosen finite representation and therefore should not be identified a priori with the dimension of the full Lie symmetry algebra.

Let $\Delta^{(s)}=\mathbf{0}$ be an order-$s$ DE system and let $v\in\mathfrak{X}(E)$ be a candidate Lie point generator. At a given jet $z^{(s)}\in J^{(s)}(\pi)$, the prolongation operator $\mathrm{pr}^{(s)}:\mathfrak{X}(E)\to\mathfrak{X}\paraBrTerm{J^{(s)}(\pi)}$ in \eqref{eqn:prolongationOfGens} maps $v$ to its order-$s$ prolongation. Meanwhile,
\begin{equation}
    \mathrm{J}_{\Delta^{(s)}}\paraBrTerm{z^{(s)}}: 
    T_{z^{(s)}}J^{(s)}\paraBrTerm{\pi} \to T_{\mathbf{y}}\mathbb{R}^m
\end{equation}
denotes the Jacobian matrix of the DE system at that jet with $\mathbf{y} = \Delta^{(s)}\paraBrTerm{z^{(s)}}$. We define the corresponding \textit{pointwise determining operator} by
\begin{equation}
    \label{eqn:detResOp}
    \left.\mathcal{D}_{\Delta}\right\vert_{z^{(s)}}\sqBrTerm{v}
    \coloneqq
    \mathrm{J}_{\Delta^{(s)}}\paraBrTerm{z^{(s)}}
    \,
    \left. \mathrm{pr}^{(s)}\paraBrTerm{v} \right\vert_{z^{(s)}}.
\end{equation}
Accordingly, the infinitesimal symmetry condition in \eqref{eqn:symCondOverInfiniActOnEqnMan} requires $\left.\mathcal{D}_{\Delta}\right\vert_{z^{(s)}}\sqBrTerm{v}=\mathbf{0}$ for every $z^{(s)}\in\Sigma^{(s)}$. Since prolongation is linear in $v$ and $\mathrm{J}_{\Delta^{(s)}}\paraBrTerm{z^{(s)}}$ acts linearly on the resulting tangent vector, $\left.\mathcal{D}_{\Delta}\right\vert_{z^{(s)}}$ is itself a linear operator in the generator argument.

The difficulty identified in \textit{\nameref*{subsec:limitOfOptimFormulation}} therefore originates not from the determining operator itself, but from the nonlinear parameterization of its argument. Let $\Psi$ in \eqref{eqn:paramEvalMap} denote the realization map associated with a standard neural-network parameterization. Although $\left.\mathcal{D}_{\Delta}\right\vert_{z^{(s)}}$ is linear in $v$, the composition $\left.\mathcal{D}_{\Delta}\right\vert_{z^{(s)}}\circ\Psi$ is generally nonlinear in $\boldsymbol{\Theta}$. For convenience, define the corresponding \textit{parameterized residual map}
\begin{equation}
    \label{eqn:paramDetResidualMap}
    \mathbf{r}_{z^{(s)}}\paraBrTerm{\boldsymbol{\Theta}} \coloneqq 
    \left. \mathcal{D}_{\Delta} \right\vert_{z^{(s)}} \sqBrTerm{v_{\boldsymbol{\Theta}}}.
\end{equation}
Its zero set in $\mathcal{P}$ is therefore not, in general, a fixed linear subspace. Instead, recovering feasibility requires successive local corrections in parameter space. For example, a damped Gauss--Newton step may be written as
\begin{equation}
    \label{eqn:dgnForParams}
    \begin{aligned}
    \delta\boldsymbol{\Theta}_{\mathrm{DGN}}
    &\coloneqq \underset{\delta\boldsymbol{\Theta}}{\operatorname{argmin}} \\
    &\frac{1}{2}
    \Big\Vert
        \mathbf{r}_{z^{(s)}}\paraBrTerm{\boldsymbol{\Theta}}
        + \mathrm{J}_{\mathbf{r}_{z^{(s)}}}\paraBrTerm{\boldsymbol{\Theta}}
          \,\delta\boldsymbol{\Theta}
    \Big\Vert_2^2 + \\
    &\frac{\lambda}{2}
    \left\Vert\delta\boldsymbol{\Theta}\right\Vert_2^2 .
    \end{aligned}
\end{equation}
where $\lambda>0$ is a damping parameter. Such a correction is determined by the Jacobian of the residual map at the current parameter iterate and therefore restores only local feasibility. As $\boldsymbol{\Theta}$ changes, the corresponding local row and null spaces change as well.

The operator-theoretic formulation instead adopts a fixed functional representation that is exactly linear in its trainable coefficients. For this purpose, we are inspired by the spline-based functional representation employed by Kolmogorov--Arnold Networks (KANs)~\cite{liu2025kan}. A standard multilayer KAN, however, remains nonlinear in its trainable parameters and therefore does not provide the fixed linear structure required here. We consequently retain only its spline-based representational principle and construct a fixed spline dictionary: the basis functions and their underlying grid are frozen, while only the expansion coefficients remain trainable. Hence, the represented infinitesimal coefficients may remain highly nonlinear functions of their inputs while depending linearly on the finite trainable parameter vector.

Let
\begin{equation}
    \label{eqn:linearGenRepMap}
    \Psi^{(k)}: \mathbb{R}^P \to \mathfrak{X}(E),
    \quad
    \boldsymbol{\Theta} \longmapsto v_{\boldsymbol{\Theta}}
\end{equation}
denote the resulting realization map associated with the fixed dictionary $D_{(k)}$ of maximum \textit{interaction order} $k$, where $P$ is the total number of trainable parameters. By construction, $\Psi^{(k)}$ is linear in $\boldsymbol{\Theta}$. Since the determining operator is also linear in its generator argument, the composition
\begin{equation}
    \label{eqn:linearPointwiseDetOp}
    \left. \mathcal{D}_{\Delta} \right\vert_{z^{(s)}} \circ \Psi^{(k)}:
    \mathbb{R}^P \to \mathbb{R}^{m}
\end{equation}
is a fixed linear operator for every fixed $z^{(s)}$. Its kernel is therefore a fixed linear subspace of the parameter space $\mathbb{R}^{P}$, rather than a parameter-dependent local tangent approximation. The precise construction of $D_{(k)}$, including its spline basis, interaction structure, grid specification, and the representational consequences of these choices, is deferred to Section~\ref{sec:implAndCompConsid}.

To convert the pointwise operator in \eqref{eqn:detResOp} into a finite computational object, we evaluate it over a fixed on-shell anchor dataset
$\mathbf{S}^{\mathrm{on}}_{\mathrm{anc}}\subset\mathbf{S}^{\mathrm{on}}$, with $A_{\mathrm{on}}\coloneqq\left\vert\mathbf{S}^{\mathrm{on}}_{\mathrm{anc}}\right\vert$. We define the \textit{sampled determining operator} by stacking the pointwise determining responses over all anchor jets,
\begin{equation}
    \label{eqn:sampledDeterminingOperator}
    \begin{split}
        &\left. \widehat{\mathcal{D}}_{\Delta} \right\vert_{\mathbf{S}^{\mathrm{on}}_{\mathrm{anc}}}\sqBrTerm{v} \coloneqq \\
            &\qquad
            \operatorname{vec} \paraBrTerm{
                \paraBrTerm{
                \left.
                \mathcal{D}_{\Delta}
                \right\vert_{z_a^{(s)}}
                \sqBrTerm{v}
                }_{z_a^{(s)} \in \mathbf{S}^{\mathrm{on}}_{\mathrm{anc}}}
            } \in \mathbb{R}^{A_{\mathrm{on}} m},
    \end{split}
\end{equation}
where $m$ is the number of independent equations constituting $\Delta^{(s)}=\mathbf{0}$. Equivalently, each block of \eqref{eqn:sampledDeterminingOperator} is obtained from $\mathrm{J}_{\Delta^{(s)}}\paraBrTerm{z_a^{(s)}}$ acting on
$\left.\mathrm{pr}^{(s)}\paraBrTerm{v}\right\vert_{z_a^{(s)}}$. Since every pointwise determining operator is linear in $v$, their stacked realization
$\left.\widehat{\mathcal{D}}_{\Delta}\right\vert_{\mathbf{S}^{\mathrm{on}}_{\mathrm{anc}}}$ is also linear.

Composing the sampled determining operator with the fixed linear realization map $\Psi^{(k)}$ in \eqref{eqn:linearGenRepMap} yields the \textit{represented sampled determining operator}
\begin{equation}
    \label{eqn:representedSampledDetOp}
    \left. \widehat{\mathcal{D}}_{\Delta}^{(k)} \right\vert_{\mathbf{S}^{\mathrm{on}}_{\mathrm{anc}}} \coloneqq
    \left. \widehat{\mathcal{D}}_{\Delta} \right\vert_{\mathbf{S}^{\mathrm{on}}_{\mathrm{anc}}} \circ \Psi^{(k)}:
    \mathbb{R}^{P} \to \mathbb{R}^{A_{\mathrm{on}}m}.
\end{equation}
The corresponding \textit{sampled determining residual response} is therefore
\begin{equation}
    \label{eqn:sampledDeterminingResidualResp}
    \widehat{\mathbf{r}}_{\mathbf{S}^{\mathrm{on}}_{\mathrm{anc}}}\paraBrTerm{\boldsymbol{\Theta}}\coloneqq
    \left. \widehat{\mathcal{D}}_{\Delta}^{(k)} \right\vert_{\mathbf{S}^{\mathrm{on}}_{\mathrm{anc}}}\sqBrTerm{\boldsymbol{\Theta}}.
\end{equation}
Because both constituent maps in \eqref{eqn:representedSampledDetOp} are linear, the residual response is linear in $\boldsymbol{\Theta}$. Hence, its Jacobian
\begin{equation}
    \label{eqn:sampledDetResDerivWRTParams}
    \widehat{M}_{\Delta}^{(k)} \coloneqq
    \frac{\partial \widehat{\mathbf{r}}_{\mathbf{S}^{\mathrm{on}}_{\mathrm{anc}}}}{\partial\boldsymbol{\Theta}}
    \in
    \mathbb{R}^{A_{\mathrm{on}}m\times P}
\end{equation}
is independent of $\boldsymbol{\Theta}$. In particular,
\begin{equation}
    \label{eqn:sampledDetResidualMatrixForm}
    \widehat{\mathbf{r}}_{\mathbf{S}^{\mathrm{on}}_{\mathrm{anc}}} \paraBrTerm{\boldsymbol{\Theta}} =
    \widehat{M}_{\Delta}^{(k)}\boldsymbol{\Theta},
\end{equation}
so the sampled determining equations reduce to the fixed homogeneous linear system
\begin{equation}
    \label{eqn:sampledDetSystem}
    \widehat{M}_{\Delta}^{(k)}\boldsymbol{\Theta} = \mathbf{0}.
\end{equation}
Thus, unlike the parameter-dependent local nullspace encountered under the nonlinear realization map, the present formulation exposes a fixed nullspace directly in the parameter space.

The first admissible restriction is consequently defined as
\begin{equation}
    \label{eqn:symAdmissibleRestriction}
    \boldsymbol{\Theta} \in \widehat{N}_{\Delta}^{(k)} \coloneqq
    \ker \paraBrTerm{\widehat{M}_{\Delta}^{(k)}}\subseteq \mathbb{R}^{P}.
\end{equation}
Accordingly, the infinitesimal symmetry condition is no longer imposed through the loss $\mathcal{L}_{\mathrm{sym}}$ in \eqref{eqn:nrmInfSymCost}. Instead, all subsequent discovery operations are restricted directly to the represented \textit{sampled nullspace} $\widehat{N}_{\Delta}^{(k)}$.

The relationship between this parameter-space nullspace and the corresponding represented vector fields should be stated carefully. In general,
\begin{equation}
    \small
    \ker \paraBrTerm{\left. \widehat{\mathcal{D}}_{\Delta} \right\vert_{\mathbf{S}^{\mathrm{on}}_{\mathrm{anc}}} \circ \Psi^{(k)}} =
        \paraBrTerm{\Psi^{(k)}}^{-1} \paraBrTerm{
            \ker \left. \widehat{\mathcal{D}}_{\Delta} \right\vert_{\mathbf{S}^{\mathrm{on}}_{\mathrm{anc}}}
        }.
\end{equation}
If $\Psi^{(k)}$ is injective, its image satisfies
\begin{equation}
    \label{eqn:representedSampledKernel}
    \Psi^{(k)} \paraBrTerm{\widehat{N}_{\Delta}^{(k)}} =
    \operatorname{im} \paraBrTerm{\Psi^{(k)}} \cap \ker \left. \widehat{\mathcal{D}}_{\Delta} \right\vert_{\mathbf{S}^{\mathrm{on}}_{\mathrm{anc}}}.
\end{equation}
Hence, the recovered symmetry sector is necessarily restricted by the expressive richness of the chosen dictionary $D_{(k)}$.

Moreover, because \eqref{eqn:representedSampledKernel} is constructed from finitely many on-shell anchors, $\widehat{N}_{\Delta}^{(k)}$ should be interpreted as a represented sampled symmetry nullspace, rather than being identified automatically with the exact determining kernel over the entire equation manifold. Insufficient sampling or degeneracy of the representation may introduce additional null directions that do not correspond to genuine symmetries on $\Sigma^{(s)}$.

\begin{remark}
    \label{rem:degeneracyOfRealizationMap}
    Since $\Psi^{(k)}$ is linear in $\boldsymbol{\Theta}$, its derivative with respect to $\boldsymbol{\Theta}$ is independent of any evaluation point $\boldsymbol{\Theta}_0$. It is completely determined by the fixed dictionary $D_{(k)}$. Consequently, rank deficiency or redundancy in the dictionary may introduce non-identifiable parameter directions and artificial contributions to $\widehat{N}_{\Delta}^{(k)}$. The construction of $D_{(k)}$, the distribution of its spline basis functions over the prescribed grid, and the numerical rank safeguards used to control these effects are discussed in Section~\ref{sec:implAndCompConsid}.
\end{remark}

\subsection{Intrinsic Admissibility Constraints}
\label{subsec:intrAdmConstraints}
Definition~\ref{defn:intrSymAlg} requires every nonzero $v\in\mathfrak{g}_\mathrm{int}$ to be both nontrivial and solution-independent. Consequently, the represented sampled symmetry nullspace $\widehat{N}_{\Delta}^{(k)}$ in \eqref{eqn:symAdmissibleRestriction} is generally broader than the target intrinsic sector, as it may also contain directions that are trivial or carry prohibited solution-dependent information. For linear DE systems, an important part of the latter sector can be identified and removed through an additional operator-theoretic restriction. For nonlinear DE systems, however, solution dependence does not generally admit the same linear decomposition. Moreover, as shown later in this subsection, nontriviality itself cannot be imposed simply by restricting to another vector subspace and therefore remains enforced through $\mathcal{L}_\mathrm{ntr}$ in \eqref{eqn:nonTrivCost}.

For both homogeneous and inhomogeneous linear DE systems, the associated homogeneous operator generates an additive symmetry sector. Its algebraic structure is summarized below.

\begin{proposition}[Additive solution subalgebra and ideal]
    \label{pro:addSltnIdeal}
    Let
    \begin{equation}
        \Delta^{(s)} \coloneqq \mathrm{L}^{(s)}\sqBrTerm{\mathbf{u}} - \mathbf{g} = \mathbf{0}
    \end{equation}
    be a linear DE system, and define
    \begin{equation}
        \label{eqn:addSolAlg}
        \begin{aligned}
        \mathcal{A}_{\mathrm{sol}} \coloneqq \Big\{\,
            &v_{\boldsymbol{\varphi}}
            = \sum_{\alpha=1}^{q}
              \varphi^\alpha\paraBrTerm{\mathbf{x}}\partial_{u^\alpha}
            \;\Bigm|\\[-0.2em]
            &\mathrm{L}^{(s)}
              \sqBrTerm{\sqBrTerm{j^s i_{\boldsymbol{\varphi}}}
              \paraBrTerm{\mathbf{x}}}=\mathbf{0},\;
              \forall\,\mathbf{x}\in M\subset\mathbf{X}\,\Big\}.
        \end{aligned}
    \end{equation}
    Then $\mathcal{A}_{\mathrm{sol}}\leq\mathfrak{g}$ and
    \begin{equation}
        \sqBrTerm{ v_{\boldsymbol{\varphi}_1}, v_{\boldsymbol{\varphi}_2}} = 0
    \end{equation}
    for all $v_{\boldsymbol{\varphi}_1}, v_{\boldsymbol{\varphi}_2} \in\mathcal{A}_{\mathrm{sol}}$.
    Hence, $\mathcal{A}_{\mathrm{sol}}$ is an Abelian Lie subalgebra of $\mathfrak{g}$.

    If, in addition, the Lie point symmetries under consideration admit the \textit{projectible fiber-affine form}
    $v=\boldsymbol{\xi}\paraBrTerm{\mathbf{x}}\cdot\partial_{\mathbf{x}} + \paraBrTerm{A\paraBrTerm{\mathbf{x}}\mathbf{u} + \mathbf{b}\paraBrTerm{\mathbf{x}}}\cdot\partial_{\mathbf{u}}$,
    then
    \begin{equation}
        \sqBrTerm{\mathfrak{g}, \mathcal{A}_{\mathrm{sol}}} \subseteq \mathcal{A}_{\mathrm{sol}},
    \end{equation}
    and therefore $\mathcal{A}_{\mathrm{sol}}\triangleleft\mathfrak{g}$ is an Abelian ideal.
\end{proposition}

A proof of Proposition~\ref{pro:addSltnIdeal} is provided in \ref{subapp:additiveSolutionSubalgebraAndIdeal}. The first statement follows directly from linearity: if $\mathbf{u}$ satisfies the original system and $\boldsymbol{\varphi}$ belongs to the kernel of its associated homogeneous operator, then
\begin{equation}
    \mathrm{L}^{(s)} \sqBrTerm{\mathbf{u} + \varepsilon\boldsymbol{\varphi}} - \mathbf{g} =
    \mathrm{L}^{(s)}\sqBrTerm{\mathbf{u}} - \mathbf{g}
\end{equation}
for every sufficiently small $\varepsilon$. Thus, $v_{\boldsymbol{\varphi}}$ generates additive motion within the solution family. For a homogeneous linear system, $\boldsymbol{\varphi}$ is itself a solution of the original DE system and the construction coincides directly with the solution-dependent vertical class of Definition~\ref{defn:sltnDepGen}. For an inhomogeneous linear system, $\boldsymbol{\varphi}$ instead solves the associated homogeneous system; equivalently, it represents the difference between two solutions of the affine inhomogeneous problem.

Proposition~\ref{pro:addSltnIdeal} further shows that $\mathcal{A}_{\mathrm{sol}}$ inherits the functional freedom of $\ker\mathrm{L}^{(s)}$ and may therefore be infinite dimensional. This is one of the reasons, in addition to those discussed in \textit{\nameref*{subsec:intrAndSltnDepSymSectors}}, for excluding the additive solution sector from the target intrinsic algebra. Nevertheless, removal of $\mathcal{A}_{\mathrm{sol}}$ does not in general imply that the remaining symmetry algebra is finite dimensional, since other forms of arbitrary functional freedom may persist, as discussed in \ref{app:geomViewPntAndFormalInteg}.

For nonlinear DE systems, there is generally no analogous additive solution subspace generated by superposition. In particular, the purely vertical solution-dependent generators of Definition~\ref{defn:sltnDepGen} need not be closed under addition, since two admissible local solutions $\boldsymbol{\varphi}_i$ and $\boldsymbol{\varphi}_j$ do not generally imply that
$\boldsymbol{\varphi}_i+\boldsymbol{\varphi}_j$ is also a solution. Hence, the corresponding solution-dependent sector does not generally admit the same linear operator-theoretic exclusion employed for $\mathcal{A}_{\mathrm{sol}}$. The applicability of the proposed framework to nonlinear DE systems, together with the structural conditions associated with projectable fiber-affine generators, is discussed in Section~\ref{sec:discApplAndLimit}.

To identify the representable additive solution sector of a linear DE system, we introduce another operator that is linear in the candidate generator. Let $v_{\boldsymbol{\Theta}}\in\mathfrak{X}(E)$ be specified by the coefficient pair $\paraBrTerm{\boldsymbol{\xi}_{\boldsymbol{\Theta}}, \boldsymbol{\varphi}_{\boldsymbol{\Theta}}}$. Define
\begin{equation}
    \label{eqn:naddOperator}
    \mathcal{T}_{\mathrm{nadd}}\sqBrTerm{v_{\boldsymbol{\Theta}}}\coloneqq
    \begin{bmatrix}
        \boldsymbol{\xi}_{\boldsymbol{\Theta}}
        \\
        \operatorname{vec}
        \paraBrTerm{
            \partial_{\mathbf{u}}
            \boldsymbol{\varphi}_{\boldsymbol{\Theta}}
        }
    \end{bmatrix},
\end{equation}
where $\partial_{\mathbf{u}}\boldsymbol{\varphi}_{\boldsymbol{\Theta}}$ denotes the Jacobian of the vertical coefficients with respect to the dependent coordinates $\mathbf{u}$. Both constituent operations are linear in $v_{\boldsymbol{\Theta}}$ and therefore $\mathcal{T}_{\mathrm{nadd}}$ is a linear operator. Its kernel consists of represented generators satisfying $\boldsymbol{\xi}_{\boldsymbol{\Theta}}=\mathbf{0}$ and $\partial_{\mathbf{u}}\boldsymbol{\varphi}_{\boldsymbol{\Theta}}=\mathbf{0}$, i.e., purely vertical generators whose coefficients depend only on $\mathbf{x}$.

We evaluate \eqref{eqn:naddOperator} over the fixed on-shell anchor dataset and compose it with the linear realization map $\Psi^{(k)}$. This yields the sampled represented operator
\begin{equation}
    \label{eqn:sampledNaddOperator}
    \left. \widehat{\mathcal{T}}_{\mathrm{nadd}}^{(k)} \right\vert_{\mathbf{S}^{\mathrm{on}}_{\mathrm{anc}}} :
    \mathbb{R}^{P}\to\mathbb{R}^{A_{\mathrm{on}}\paraBrTerm{p+q^2}},
\end{equation}
whose action is obtained by stacking $\boldsymbol{\xi}_{\boldsymbol{\Theta}}$ and $\partial_{\mathbf{u}}\boldsymbol{\varphi}_{\boldsymbol{\Theta}}$ over all samples in $\mathbf{S}^{\mathrm{on}}_{\mathrm{anc}}$. Since $\Psi^{(k)}$ and $\mathcal{T}_{\mathrm{nadd}}$ are both linear in $\boldsymbol{\Theta}$, the operator in \eqref{eqn:sampledNaddOperator} is represented by a fixed matrix.

Within the represented sampled symmetry nullspace $\widehat{N}_{\Delta}^{(k)}$, simultaneous vanishing of these three responses identifies the representable purely vertical, $\mathbf{u}$-independent symmetry sector. For a linear DE system, Proposition~\ref{pro:addSltnIdeal} then identifies this sector with the additive solutions of the associated homogeneous operator. Accordingly, we define the \textit{sampled represented additive solution sector}
\begin{equation}
    \label{eqn:sampledAddSltnSector}
    \widehat{\mathcal{A}}_{\mathrm{sol}}^{(k)}
    \coloneqq
    \widehat{N}_{\Delta}^{(k)}
    \cap
    \ker
    \paraBrTerm{
        \left.
        \widehat{\mathcal{T}}_{\mathrm{nadd}}^{(k)}
        \right\vert_{\mathbf{S}^{\mathrm{on}}_{\mathrm{anc}}}
    }.
\end{equation}
As with the sampled determining nullspace, the qualifier \textit{sampled represented} is essential: \eqref{eqn:sampledAddSltnSector} identifies the portion of the additive sector visible within the finite dictionary $D_{(k)}$ and over the chosen anchor set.

Since $\widehat{\mathcal{A}}_{\mathrm{sol}}^{(k)}$ is a vector subspace of $\widehat{N}_{\Delta}^{(k)}$, we remove it directly from the admissible parameter space rather than attempting to repel the optimizer through $\mathcal{L}_{\mathrm{dep}}$. Using the Euclidean inner product of the parameter space $\mathbb{R}^{P}$, define
\begin{equation}
    \label{eqn:sltnIndAdmissibleRestriction}
    \widehat{N}_{\mathrm{nadd}}^{(k)} \coloneqq
    \widehat{N}_{\Delta}^{(k)} \cap
    \paraBrTerm{\widehat{\mathcal{A}}_{\mathrm{sol}}^{(k)}}^{\perp}.
\end{equation}
Because $\widehat{\mathcal{A}}_{\mathrm{sol}}^{(k)} \subseteq \widehat{N}_{\Delta}^{(k)}$, this construction yields the direct-sum decomposition
\begin{equation}
    \label{eqn:additiveDirectSumDecomp}
    \widehat{N}_{\Delta}^{(k)} =
    \widehat{\mathcal{A}}_{\mathrm{sol}}^{(k)} \oplus \widehat{N}_{\mathrm{nadd}}^{(k)}.
\end{equation}
Hence, $\widehat{N}_{\mathrm{nadd}}^{(k)}$ retains only representatives whose coefficient vectors contain no component along the sampled represented additive solution sector.

Importantly, \eqref{eqn:sltnIndAdmissibleRestriction} removes not only generators lying entirely in $\widehat{\mathcal{A}}_{\mathrm{sol}}^{(k)}$, but also additive components occurring inside mixed representatives. This is consistent with Remark~\ref{rem:mixedSltnDepGen}, which reserves the term \textit{solution-dependent generator} for the purely vertical class and does not classify a mixed generator with a nonvanishing horizontal component as solution-dependent merely because it contains a solution-dependent vertical contribution. The present linear-DE restriction is deliberately stronger: components lying in the sampled represented additive solution sector are removed from every admissible representative before sequential discovery proceeds. The orthogonal complement in \eqref{eqn:sltnIndAdmissibleRestriction} is a computational choice induced by the finite coefficient representation and should not be interpreted as a canonical orthogonal decomposition of the abstract Lie symmetry algebra.

To examine whether nontriviality can likewise be imposed through a linear admissible restriction, we first exploit the linearity of the prolonged characteristic information with respect to the generator. For the fixed linear realization $v_{\boldsymbol{\Theta}}=\Psi^{(k)}\paraBrTerm{\boldsymbol{\Theta}}$, define the \textit{sampled prolonged-characteristic operator}
\begin{equation}
    \label{eqn:sampledTrivOperator}
    \left. \widehat{\mathcal{T}}_{\mathrm{triv}}^{(k)} \right\vert_{\mathbf{S}^{\mathrm{on}}_{\mathrm{anc}}}\sqBrTerm{\boldsymbol{\Theta}}
    \coloneqq
    \operatorname{vec}\paraBrTerm{
        \paraBrTerm{
            \left. \mathbf{E}_{v_{\boldsymbol{\Theta}}}^{(s-1)} \right\vert_{z_a^{(s)}}
        }_{z_a^{(s)} \in \mathbf{S}^{\mathrm{on}}_{\mathrm{anc}}}
    }.
\end{equation}
Because $\Psi^{(k)}$ is linear in $\boldsymbol{\Theta}$, $\mathbf{Q}_v$ is linear in $v$, and the total derivative operators entering $\mathbf{E}_v^{(s-1)}$ are linear, the operator in \eqref{eqn:sampledTrivOperator} is linear in $\boldsymbol{\Theta}$.

\begin{proposition}[Sampled triviality and geometric quotient]
    \label{pro:sampledTrivQuotient}
    Let $\widehat{N}_{\mathrm{nadd}}^{(k)}\subseteq\mathbb{R}^{P}$ be the represented admissible coefficient subspace defined in \eqref{eqn:sltnIndAdmissibleRestriction}, and let $\left. \widehat{\mathcal{T}}_{\mathrm{triv}}^{(k)} \right\vert_{\mathbf{S}^{\mathrm{on}}_{\mathrm{anc}}}$ be the \textit{sampled prolonged-characteristic operator} in \eqref{eqn:sampledTrivOperator}. Then the \textit{sampled-trivial sector}
    \begin{equation}
        \label{eqn:trivVecSubsp}
        \widehat{K}_{\mathrm{triv}}^{(k)} \coloneqq
            \widehat{N}_{\mathrm{nadd}}^{(k)} \cap
            \ker \paraBrTerm{
                \left. \widehat{\mathcal{T}}_{\mathrm{triv}}^{(k)} \right\vert_{\mathbf{S}^{\mathrm{on}}_{\mathrm{anc}}}
            }
    \end{equation}
    is a real vector subspace. Moreover, the quotient
    \begin{equation}
        \label{eqn:geomNonTrivSubsp}
        \widehat{N}_{\mathrm{geom}}^{(k)} \coloneqq
        \frac{\widehat{N}_{\mathrm{nadd}}^{(k)}}{\widehat{K}_{\mathrm{triv}}^{(k)}}
    \end{equation}
    is naturally isomorphic to the image of the sampled prolonged-characteristic operator restricted to $\widehat{N}_{\mathrm{nadd}}^{(k)}$,
    \begin{equation}
        \label{eqn:geomNonTrivIso}
        \widehat{N}_{\mathrm{geom}}^{(k)}
        \cong
        \operatorname{im}
        \paraBrTerm{
            \left. \widehat{\mathcal{T}}_{\mathrm{triv}}^{(k)}
            \right\vert_{\mathbf{S}^{\mathrm{on}}_{\mathrm{anc}}, \widehat{N}_{\mathrm{nadd}}^{(k)}}
        }.
    \end{equation}
    In contrast, the set of coefficient vectors having nonzero sampled prolonged-characteristic response,
    \begin{equation}
        \label{eqn:sampledNonTrivSet}
        \widehat{\mathcal{V}}_{\mathrm{ntr}}^{(k)} \coloneqq
            \widehat{N}_{\mathrm{nadd}}^{(k)}\setminus\widehat{K}_{\mathrm{triv}}^{(k)},
    \end{equation}
    is not a vector subspace.
\end{proposition}

A proof for Proposition~\ref{pro:sampledTrivQuotient} is also included in \ref{subapp:sampledTrivAndGeomQuotient}. Proposition~\ref{pro:sampledTrivQuotient} follows directly from linearity and the first isomorphism theorem. In particular, the directions whose prolonged characteristic information vanishes over the entire anchor dataset form a kernel and can therefore be identified as a linear subspace $\widehat{K}_{\mathrm{triv}}^{(k)}$. The set obtained by removing this kernel, however, is not closed under either scalar multiplication by zero or vector addition and hence cannot serve as another admissible coefficient subspace.

One could instead operate on the quotient $\widehat{N}_{\mathrm{geom}}^{(k)}$, but its elements are equivalence classes modulo sampled-trivial directions rather than explicit generator representatives. Selecting a particular linear complement of $\widehat{K}_{\mathrm{triv}}^{(k)}$ would provide representatives, but such a complement is not canonical and would introduce an additional representation-dependent choice. Furthermore, removal of the exact sampled kernel would not prevent discovery along near-kernel directions whose prolonged-characteristic response is nonzero but numerically negligible.

For these reasons, nontriviality remains enforced through the relative response and minimum-magnitude constraints constituting $\mathcal{L}_{\mathrm{ntr}}$ in \eqref{eqn:nonTrivCost}, now restricted to candidates already lying in $\widehat{N}_{\mathrm{nadd}}^{(k)}$. Thus, unlike the infinitesimal symmetry condition and the additive solution sector, nontriviality is not converted into a further linear admissible restriction.

\subsection{Sequential Discovery and Novelty Restrictions}
\label{subsec:seqDiscoveryAndNoveltyConstraints}
The admissible restrictions introduced in \textit{\nameref*{subsec:operatorTheoreticSymRep}} and \textit{\nameref*{subsec:intrAdmConstraints}} determine which represented candidate generators are eligible for the recovery of $\mathfrak{g}_\mathrm{int}$. Novelty, however, is not an intrinsic property of a candidate in isolation, but is defined relative to the generators already discovered. In particular, every successive generator must satisfy
\begin{equation}
    v_i \notin \operatorname{span}_{\mathbb{R}}\curBarTerm{v_1,\dotsc,v_{i-1}},
    \quad\forall\, i>1.
\end{equation}
In the initial optimization formulation of \textit{\nameref*{subsec:initialFormulation}}, this requirement was imposed by decomposing $v_i$ into the complementary component $v_{\perp,i}$ and the component $v_{\parallel,i}$ represented by the previously discovered span. The operator-theoretic formulation developed below instead constructs a novelty restriction that excludes this previously represented subspace directly from the space available to each successive discovery step.

The need for a global test of linear independence nevertheless remains unchanged. In particular, linear independence is an ambient vector-field property over real constants, and finite evaluation over an inadequately sampled domain may still introduce accidental rank loss or poor conditioning. We therefore retain the off-shell anchor dataset $\mathbf{S}^{\mathrm{off}}_{\mathrm{anc}}$ introduced in \eqref{eqn:anchorDataset} when constructing the operator-theoretic novelty restriction.

We now construct the operator-theoretic novelty restriction using the same off-shell ambient vector-field responses introduced in the initial optimization formulation. For each $z_a^{(s)}\in\mathbf{S}^{\mathrm{off}}_{\mathrm{anc}}$, let
\begin{equation}
    \label{eqn:perAnchorAmbientResp}
    \mathbf{y}_{E,a}\sqBrTerm{v_{\boldsymbol{\Theta}}} \coloneqq
    \left. v_{\boldsymbol{\Theta}} \right\vert_{z_a} =
    \begin{bmatrix}
        \boldsymbol{\xi}_{\boldsymbol{\Theta}}\paraBrTerm{z_a} \\
        \boldsymbol{\varphi}_{\boldsymbol{\Theta}}\paraBrTerm{z_a}
    \end{bmatrix}
    \in \mathbb{R}^{p+q},
\end{equation}
where $z_a \coloneqq \pi_0^s\paraBrTerm{z_a^{(s)}}$. Thus, although the anchors belong to $J^{(s)}(\pi)$, novelty is evaluated through the zeroth-order responses of the ambient vector fields in $T_{z_a}E$, rather than through their prolongations.

Stacking the responses over the fixed off-shell anchor dataset gives
\begin{equation}
    \label{eqn:stackedAmbientResp}
    \begin{split}
        &Y_E\sqBrTerm{v_{\boldsymbol{\Theta}}} \coloneqq \\
            &\qquad
            \operatorname{vec}\paraBrTerm{
                \paraBrTerm{\mathbf{y}_{E,a}\sqBrTerm{v_{\boldsymbol{\Theta}}}}_{z_a^{(s)}\in\mathbf{S}^{\mathrm{off}}_{\mathrm{anc}}}
            }
            \in\mathbb{R}^{A_{\mathrm{off}}\paraBrTerm{p+q}}.
    \end{split}
\end{equation}

As in the initial formulation in \textit{\nameref*{subsec:initialFormulation}}, the horizontal and vertical ambient responses are measured using a \textit{common blockwise metric}. Let $d_{\boldsymbol{\xi}}>0$ and $d_{\boldsymbol{\varphi}}>0$ denote the fixed horizontal and vertical scaling factors, respectively, obtained from the normalization statistics associated with the off-shell anchor set. Define the \textit{single-anchor scaling matrix}
\begin{equation}
    \label{eqn:ambientRespMetricSingle}
    W_{E,0} \coloneqq
    \begin{bmatrix}
        d_{\boldsymbol{\xi}}^{-1}I_p & 0 \\
        0 & d_{\boldsymbol{\varphi}}^{-1}I_q
    \end{bmatrix},
\end{equation}
and its corresponding stacked metric
\begin{equation}
    \label{eqn:ambientRespMetric}
    W_E \coloneqq
    I_{A_{\mathrm{off}}}\otimes W_{E,0}.
\end{equation}
The same metric is therefore applied to every off-shell anchor and, importantly, to both the previously discovered generators and the candidate generator. We define the metric-normalized stacked response by
\begin{equation}
    \label{eqn:normStackedAmbientResp}
    \widetilde{Y}_E\sqBrTerm{v_{\boldsymbol{\Theta}}}\coloneqq
    W_E Y_E\sqBrTerm{v_{\boldsymbol{\Theta}}}.
\end{equation}
The normalization statistics entering $W_E$ are fixed during construction of the novelty restriction and are independent of the current coefficient vector $\boldsymbol{\Theta}$. Hence, $\widetilde{Y}_E\sqBrTerm{v_{\boldsymbol{\Theta}}}$ remains linear in $\boldsymbol{\Theta}$. Details of the normalization statistics and their numerical evaluation are deferred to Section~\ref{sec:implAndCompConsid}.

For the $i^\mathrm{th}$ discovery step, $i>1$, the previously discovered generators define the normalized old-response matrix
\begin{equation}
    \label{eqn:normOldRespMatrix}
    \widetilde{R}_{E,i}\coloneqq
    \begin{bmatrix}
        \,|\, &        & \,|\, \\
        \widetilde{Y}_E\sqBrTerm{v_1} & \cdots & \widetilde{Y}_E\sqBrTerm{v_{i-1}} \\
        \,|\, &        & \,|\,
    \end{bmatrix}.
\end{equation}
Equations~\eqref{eqn:normStackedAmbientResp} and~\eqref{eqn:normOldRespMatrix} are the metric-normalized counterparts of $Y_i$ in \eqref{eqn:newAnchorRespVec} and $R_i$ in \eqref{eqn:oldRespMatrix}, respectively.

Let $\widetilde{Q}_{E,i}$ be an orthonormal basis for the column space of $\widetilde{R}_{E,i}$ obtained through a rank-revealing QR factorization, such that
\begin{equation}
    \label{eqn:oldRespQR}
    \operatorname{col}\paraBrTerm{\widetilde{Q}_{E,i}} =
    \operatorname{col}\paraBrTerm{\widetilde{R}_{E,i}},
    \quad
    \widetilde{Q}_{E,i}^{\top}\widetilde{Q}_{E,i} = I.
\end{equation}
Since the previously discovered generators and the metric $W_E$ are fixed during the $i^\mathrm{th}$ discovery step, $\widetilde{Q}_{E,i}$ is also fixed. We therefore define the \textit{sampled represented novelty operator}
\begin{equation}
    \label{eqn:noveltyOp}
    \left. \widehat{\operatorname{B}}_{E,i}^{(k)} \right\vert_{\mathbf{S}^{\mathrm{off}}_{\mathrm{anc}}}\sqBrTerm{\boldsymbol{\Theta}}
    \coloneqq
    \widetilde{Q}_{E,i}^{\top}\widetilde{Y}_E\sqBrTerm{v_{\boldsymbol{\Theta}}}.
\end{equation}
Because the realization $v_{\boldsymbol{\Theta}} = \Psi^{(k)}\paraBrTerm{\boldsymbol{\Theta}}$ is linear in $\boldsymbol{\Theta}$, while $W_E$ and $\widetilde{Q}_{E,i}$ are fixed, the operator in \eqref{eqn:noveltyOp} is linear in $\boldsymbol{\Theta}$. Its kernel consists of coefficient directions whose metric-normalized, globally stacked off-shell ambient responses are orthogonal to the response space generated by the previously discovered generators.

Accordingly, for every $i>1$, the parameter space available to the next discovery step is restricted to
\begin{equation}
    \label{eqn:noveltyRestriction}
    \widehat{N}_{\mathrm{nov},i}^{(k)}\coloneqq\widehat{N}_{\mathrm{nadd}}^{(k)}
    \cap\ker
    \paraBrTerm{\left. \widehat{\operatorname{B}}_{E,i}^{(k)} \right\vert_{\mathbf{S}^{\mathrm{off}}_{\mathrm{anc}}}}.
\end{equation}
For the first generator, no previously discovered direction exists, and hence
\begin{equation}
    \label{eqn:firstNoveltyRestriction}
    \widehat{N}_{\mathrm{nov},1}^{(k)}\coloneqq\widehat{N}_{\mathrm{nadd}}^{(k)}.
\end{equation}
Thus, each successive discovery is carried out directly in a coefficient subspace whose metric-normalized off-shell ambient response is orthogonal to the response space already recovered. In contrast to the initial optimization formulation, no parallel response component $v_{\parallel, i}$ needs to be learned and subsequently suppressed through a novelty loss.

\section{Implementation and Computational Considerations}
\label{sec:implAndCompConsid}
Section~\ref{sec:framework} established the theoretical framework underlying the proposed discovery procedure, but deliberately deferred implementation details that are central to its computational realization and did not describe the sequential discovery algorithm in operational terms. The present section addresses these aspects without attempting to document every low-level implementation detail. For further implementation-specific information, we refer the reader to the accompanying Code Ocean capsule, which provides the source code with sufficient documentation and a modular package structure for detailed inspection. We begin with the fixed-dictionary realization of candidate generators, which preserves the linear dependence on the coefficient vector required by the operator formulation in \eqref{eqn:linearPointwiseDetOp}. We then introduce the RSP test used to assess whether the represented determining system stabilizes under successive enrichment of the model. This is followed by the numerical enforcement of nontriviality and the metric normalization used in constructing the novelty restriction. Finally, we describe the sequential recovery procedure, including the criteria used to accept individual generators, determine whether further discovery steps should be attempted, and terminate the process once the available represented symmetry directions have been exhausted. A schematic flowchart summarizing the resulting training and discovery procedure is provided in Fig.~\ref{fig:sequentialRecoveryFlowchart}.

\subsection{Fixed Dictionary and Generator Realization}
\label{subsec:fixedDictAndGenRealization}
As established in \textit{\nameref*{subsec:operatorTheoreticSymRep}}, the composition of the determining operator in \eqref{eqn:detResOp} with the realization map in \eqref{eqn:linearGenRepMap} must remain linear in the model parameters $\boldsymbol{\Theta}$. This property permits the candidate coefficients to be restricted directly to $\ker\paraBrTerm{\widehat{M}_{\Delta}^{(k)}}$ through \eqref{eqn:symAdmissibleRestriction}, thereby excluding symmetry-violating directions before optimization. A conventional neural-network realization does not, in general, preserve this linear dependence on its trainable parameters. We therefore replace our initial neural-network model by a fixed spline-based dictionary that retains sufficient representational richness while remaining exactly linear in $\boldsymbol{\Theta}$. The construction is inspired by the spline parametrization employed in Kolmogorov--Arnold Networks (KANs)~\cite{liu2025kan}, as previously noted in \textit{\nameref*{subsec:operatorTheoreticSymRep}}; unlike a general trainable KAN architecture, however, the grid and basis functions are fixed and only their expansion coefficients are optimized.

Let $\boldsymbol{\chi}=\paraBrTerm{\chi^1,\dotsc,\chi^d}$ denote the input to the generator realization, and let $\Omega=\prod_{i=1}^{d}\sqBrTerm{a_i,b_i}\subset\mathbb{R}^{d}$ be a domain containing both the training and validation datasets. We intentionally do not identify $d$ with $p+q$, since, as discussed in Section~\ref{sec:discApplAndLimit}, the input dimension of the generator realization may vary with the class and order of the underlying DE system. For each coordinate $\chi^i$, let $\big\{B_{\nu_i}^{(r)}(\chi^i)\big\}_{\nu_i=1}^{K_o}$ denote a fixed degree-$r$ \textit{B-spline basis} constructed over $\sqBrTerm{a_i,b_i}$ on a grid of resolution $o$ using the Cox--de Boor recursion~\cite{deBoor2001PracticalGuide}, where, under the adopted convention that $o$ denotes the number of uniform core-grid intervals, $K_o=o+r$.

To incorporate multivariate dependence while preserving linearity in the expansion coefficients, we construct tensor-product interactions among the univariate spline bases. For a nonempty subset $S\subseteq\curBarTerm{1,\dotsc,d}$ and an associated spline-index tuple $\boldsymbol{\nu}_S=\paraBrTerm{\nu_i}_{i\in S}$, define
\begin{equation}
    \label{eqn:tensorSplineFeature}
    B_{S,\boldsymbol{\nu}_S}^{(r)}\paraBrTerm{\boldsymbol{\chi}}
    \coloneqq
    \prod_{i\in S} B_{\nu_i}^{(r)}\paraBrTerm{\chi^i}.
\end{equation}
We define the \textit{interaction order} $k$ as the maximum number of distinct input coordinates entering any such tensor product. Accordingly, the fixed dictionary is
\begin{equation}
    \label{eqn:fixedSplineDict}
    \begin{aligned}
    D_{(k)} \coloneqq \Big\{\,
        &B_{S,\boldsymbol{\nu}_S}^{(r)} \;\Bigm|\;
        \emptyset\neq S\subseteq\curBarTerm{1,\dotsc,d},\\[-0.2em]
        &\vert S\vert\leq k,\quad
        1\leq\nu_i\leq K_o,\; i\in S
    \,\Big\}.
    \end{aligned}
\end{equation}
Thus, $k=1$ yields an additive dictionary consisting only of univariate spline features, whereas $k=2$ augments these features by all pairwise tensor products associated with distinct input coordinates. Larger values of $k$ analogously introduce higher-order coordinate interactions without altering the linear dependence of the realized generator on its coefficient vector $\boldsymbol{\Theta}$.
\begin{remark}
    \label{rem:interOrderConfusion}
    Throughout the remainder of this section, $k$ denotes the \textit{maximum interaction order} admitted by the nested dictionary $D_{(k)}$; the same index appearing as the superscript $(k)$ on $\Psi^{(k)}$ and the associated sampled operator and coefficient-space objects in Section~\ref{sec:framework} indicates that they are constructed from this dictionary level.
\end{remark}

The total number of scalar features listed in $D_{(k)}$ is
\begin{equation}
    \label{eqn:totalNumberOfSplines}
    n_{(k)}^{D} \coloneqq \left\vert D_{(k)}\right\vert =
    \sum_{\ell=1}^{k} \binom{d}{\ell} K_o^{\ell},
\end{equation}
where $K_o^{\ell}$ is the number of $\ell$-way tensor-product spline features associated with one fixed subset of $\ell$ input coordinates. We emphasize that $n_{(k)}^{D}$ counts dictionary entries rather than the dimension of their linear span. Although the univariate collection associated with each individual coordinate forms a B-spline basis, the combined dictionary $D_{(k)}$ may contain exact linear dependencies across different interaction orders, as discussed below.

For illustration, consider $d=3$ and $K_o=4$. The first-order feature block is
\begin{equation}
    \small
    \label{eqn:sampleBSplineBasis}
    \mathbf{B}_1^{(r)} \coloneqq
    \Big\{
        B_{\nu_i}^{(r)}\paraBrTerm{\chi^i}
        \Bigm |
        i\in\curBarTerm{1,2,3}, 
        \nu_i\in\curBarTerm{1,2,3,4}
    \Big\},
\end{equation}
and therefore $\big\vert\mathbf{B}_1^{(r)}\big\vert=3\times4=12$. For $\ell=2$, each feature has the form $B_{\nu_i}^{(r)}\paraBrTerm{\chi^i}
B_{\nu_j}^{(r)}\paraBrTerm{\chi^j}$ with $i<j$. A fixed coordinate pair $\paraBrTerm{i,j}$ admits $4\times4=16$ choices of the spline-index pair $\paraBrTerm{\nu_i,\nu_j}$, while $\binom{3}{2}=3$ distinct coordinate pairs are available. Hence, $\left\vert\mathbf{B}_2^{(r)}\right\vert = 3\times4^2=48$, and
\begin{equation}
    \label{eqn:numOfBSplinesSampleCase}
    n_{(2)}^{D} = \left\vert\mathbf{B}_1^{(r)}\right\vert + \left\vert\mathbf{B}_2^{(r)}\right\vert = 60.
\end{equation}
More generally, the interaction-order hierarchy induces the nested enrichment
\begin{equation}
    \small
    \label{eqn:nestedSplineDict}
    \begin{split}
        D_{(k+1)} &= D_{(k)} \cup
        \Bigg\{
            \prod_{m=1}^{k+1}
            B_{\nu_{i_m}}^{(r)}
            \paraBrTerm{\chi^{i_m}}
            \Bigm| \\
            \quad
            & 1\leq i_1<\dotsc<i_{k+1}\leq d,\; 1\leq\nu_{i_m}\leq K_o
        \Bigg\}.
    \end{split}
\end{equation}

Although the realization remains linear in the trainable coefficients $\boldsymbol{\Theta}$, its dependence on the dictionary inputs need not be linear. At the univariate level, each $B_{\nu_i}^{(r)}\paraBrTerm{\chi^i}$ is a localized degree-$r$ piecewise-polynomial function, so linear combinations of these basis functions permit nonlinear dependence of the generator coefficients, including the components $\xi^i$ and $\varphi^\alpha$, on their admissible inputs. Increasing the grid resolution $o$ enriches this univariate dependence by introducing additional localized degrees of freedom, while increasing the spline degree $r$ enlarges the local polynomial complexity. At the multivariate level, increasing the interaction order $k$ introduces tensor-product features coupling progressively larger subsets of distinct input coordinates. Hence, the representational richness of $D_{(k)}$ arises jointly from the local spline resolution and degree at the B-spline level and from cross-coordinate interactions at the dictionary level, while the realization remains exactly linear in $\boldsymbol{\Theta}$.

We emphasize, however, that this increasing representational richness should not be interpreted as an established completeness property of the finite dictionary. At present, we have not established a universal-approximation /density theorem or a quantitative approximation bound for the truncated KAN-inspired dictionary $D_{(k)}$. Consequently, failure to represent a particular symmetry direction may reflect limitations of the chosen finite values of $o$, $r$, and $k$, rather than absence of that symmetry from the underlying DE system. The computational role of successive dictionary enrichment and the corresponding rank-stabilization criterion are discussed in \textit{\nameref*{subsec:rankStabPreflightTest}}.

The nested enrichment in \eqref{eqn:nestedSplineDict} does not, in general, imply a corresponding increase in the dimension of the represented function space. Exact cross-order redundancies induced by the partition-of-unity property of B-splines, together with additional rank deficiencies that may arise from insufficient sampling of their local supports, are discussed in \ref{app:splineDictRedAndSamplingIndRankDef}.

More explicitly, any scalar coefficient function represented through the fixed dictionary can be written as
\begin{equation}
    \label{eqn:fixedDictScalarExpansion}
    f_{\boldsymbol{\theta}}\paraBrTerm{\boldsymbol{\chi}} =
    \sum_{\ell=1}^{n_{(k)}^{D}} \theta^{\ell} \psi_{\ell}\paraBrTerm{\boldsymbol{\chi}},
    \quad
    \psi_{\ell}\in D_{(k)},
\end{equation}
where $\boldsymbol{\theta}$ denotes the corresponding coefficient block of the global parameter vector $\boldsymbol{\Theta}$. Thus, the potentially nonlinear dependence on $\boldsymbol{\chi}$ is carried entirely by the fixed dictionary functions $\psi_{\ell}$, whereas the dependence on the trainable coefficients remains linear. The coefficient blocks associated with the admissible components of $\boldsymbol{\xi}$ and $\boldsymbol{\varphi}$ are then assembled into $\boldsymbol{\Theta}$ to define the realization $v_{\boldsymbol{\Theta}}=\Psi^{(k)}\paraBrTerm{\boldsymbol{\Theta}}$.

The fixed dictionary $D_{(k)}$ provides the functional building blocks of the generator realization, but the admissible dependence of the generator components on the coordinates is selected according to the class, linearity, and differential order of the underlying DE system. Accordingly, the implementation instantiates either a structurally restricted \textit{projectible fiber-affine} realization, when such a form is appropriate, or the more general point-field realization in which the coefficient functions are expanded directly over $D_{(k)}$. This choice determines both the effective dictionary input dimension $d$ introduced above and the total coefficient dimension $P$ entering the realization map $\Psi^{(k)}$ in \eqref{eqn:linearGenRepMap}.

In either case, the basis functions and spline grids remain fixed, and only their expansion coefficients are trainable. Hence, the resulting vector field $v_{\boldsymbol{\Theta}}$ remains linear in $\boldsymbol{\Theta}$, preserving the coefficient-space operator formulation developed in \textit{\nameref*{subsec:operatorTheoreticSymRep}}. The mathematical conditions motivating the realization selected for each DE class, together with the corresponding applicability and completeness restrictions, are discussed separately in Section~\ref{sec:discApplAndLimit}.

Although the framework operates at the infinitesimal level, a discovered generator can subsequently be integrated to recover its associated local one-parameter transformation. For a discovered generator $v_i$ and an initial jet $z_0^{(s)}\in J^{(s)}\paraBrTerm{\pi}$, we numerically solve
\begin{equation}
    \label{eqn:numericalFiniteFlow}
    \frac{\mathrm{d}z^{(s)}}{\mathrm{d}\epsilon} =
        \left. \mathrm{pr}^{(s)}\paraBrTerm{v_i} \right\vert_{z^{(s)}\paraBrTerm{\epsilon}},
    \quad
    z^{(s)}\paraBrTerm{0} = z_0^{(s)}.
\end{equation}
The resulting trajectory approximates the local prolonged flow defined in \eqref{eqn:algFlowOnE} and induced by the one-parameter subgroup $\exp\paraBrTerm{\epsilon v_i}$. In practice, \eqref{eqn:numericalFiniteFlow} is integrated using a differentiable ODE solver of the type employed in neural-ODE frameworks~\cite{chen2018neural}; hence, reconstruction of the finite transformation does not require a closed-form evaluation of the abstract Lie-group exponential in \eqref{eqn:expMap}.

The reconstruction also connects directly with the group-level representation in \eqref{eqn:anyGUsingExp}. Once a basis $\curBarTerm{v_1,\dotsc,v_r}$ of the recovered Lie algebra has been obtained, finite products of the numerically reconstructed one-parameter flows realize transformations in the identity-connected component $G_0$. In particular, sufficiently near the identity, an ordered product
\begin{equation}
    \label{eqn:secondKindCoordinates}
    g\paraBrTerm{\boldsymbol{\epsilon}} =
    \exp\paraBrTerm{\epsilon^1v_1} \cdots \exp\paraBrTerm{\epsilon^rv_r}
\end{equation}
provides \textit{canonical coordinates of the second kind}, with the ordering becoming relevant when the recovered algebra is non-Abelian. Although such coordinates are local, a connected transformation can in principle be constructed patchwise by composing a sequence of sufficiently small second-kind transformations, each remaining within a neighborhood where the local parametrization is valid. The numerical results in Section~\ref{sec:experiments} provide evidence for the practical need for such a patchwise strategy when transporting beyond the region over which a single learned local realization remains accurate; a systematic patchwise-transport construction is left for future work.

\subsection{Rank-Stabilization Preflight (RSP) Test}
\label{subsec:rankStabPreflightTest}
Although the applicability of the proposed framework is discussed in detail in Section~\ref{sec:discApplAndLimit}, we note here that certain classes of DE systems may admit an infinite-dimensional $\mathfrak{g}_{\mathrm{int}}$ through arbitrary functional freedom in their determining systems. In contrast, any finite dictionary $D_{(k)}$ necessarily yields only a finite-dimensional represented coefficient space and, consequently, a finite-dimensional represented geometric symmetry space. Symbolic and formal approaches address such functional freedom by analyzing the determining system after compatibility completion and reduction to an involutive or standard form~\cite{reid1991finding,Seiler2010,reid1991algorithms}. For background on the geometric meaning of functional freedom and involutivity, we refer the reader to \ref{app:geomViewPntAndFormalInteg}, together with~\cite[Chapters~14--15]{olver1995equivalence} and~\cite[Chapters~4--5]{ivey2003cartan}.

At present, we have not developed a numerical analogue capable of simultaneously diagnosing arbitrary functional freedom and carrying out the compatibility and involutivity analysis associated with Cartan--Kuranishi completion and the Cartan--K\"{a}hler framework~\cite[Chapter~7]{Seiler2010}. Developing such a procedure is beyond the scope of the present work, although Section~\ref{sec:conclusion} outlines a possible trajectory toward it within the operator-theoretic framework introduced here. Instead, we employ a computational RSP test: the represented determining system is recomputed under successive enrichments of $D_{(k)}$, and the evolution of its numerically resolved rank and nullity is monitored to determine whether the represented symmetry dimension stabilizes.

Choose a sequence of nested knot refinements for the KAN dictionaries introduced in \textit{\nameref*{subsec:fixedDictAndGenRealization}}, while initially fixing the interaction order $k$ and spline degree $r$. For grid resolutions $o_1<o_2<\cdots$, define the corresponding dictionary-generated spaces by
\begin{equation}
    \small
    \label{eqn:nestedRefinementSeq}
    \mathcal{V}_{i}^{(k,r)} \coloneqq \operatorname{span} \paraBrTerm{D_{(k),o_i}^{(r)}},
    \quad
    \mathcal{V}_{1}^{(k,r)} \subseteq \mathcal{V}_{2}^{(k,r)} \subseteq \cdots .
\end{equation}
Here, each knot grid is chosen as a refinement of its predecessor, so that the coarser spline space is exactly representable on the finer grid through standard knot refinement. Accordingly, $D_{(k),o_{i+1}}^{(r)}$ provides a richer coefficient representation than $D_{(k),o_i}^{(r)}$, although the individual B-spline basis functions themselves need not form nested sets.

For the uniform spline grids used here, this nesting can be enforced by choosing successive resolutions such that $o_{i+1}=m_i o_i$ with $m_i\in\mathbb{N}$ and $m_i>1$. Over a fixed core interval, every knot of the coarser grid is then retained by the finer grid, and, at fixed spline degree $r$, standard knot refinement yields the required inclusion of the corresponding spline spaces. Thus, the refinement increases only the representational resolution while preserving the underlying physical domain.

Let $\Psi_i^{(k)}:\mathbb{R}^{P_i}\to\mathfrak{X}(E)$ denote the realization induced by $D_{(k),o_i}^{(r)}$, and let
\begin{equation}
    \label{eqn:refinedSampledDetKernel}
    \widehat{N}_{\Delta,i}^{(k)} \coloneqq
    \ker\paraBrTerm{\widehat{M}_{\Delta,i}^{(k)}}
\end{equation}
be the corresponding represented sampled determining nullspace in \eqref{eqn:symAdmissibleRestriction}. Since the coefficient dimensions $P_i$ vary under refinement, the spaces $\widehat{N}_{\Delta,i}^{(k)}$ are not compared by literal coefficient-space inclusion. Instead, after identifying a coarse realization with its representation on the refined spline grid, the nested function spaces in \eqref{eqn:nestedRefinementSeq} imply the corresponding inclusion of represented \textit{sampled symmetry fields},
\begin{equation}
\label{eqn:refinementOfKernOfRepSmplDetOp}
    \Psi_i^{(k)}\paraBrTerm{\widehat{N}_{\Delta,i}^{(k)}} \subseteq
    \Psi_{i+1}^{(k)}\paraBrTerm{\widehat{N}_{\Delta,i+1}^{(k)}},
\end{equation}
up to the numerical rank tolerance used in constructing the sampled kernels.

\textit{\nameref*{subsec:intrAdmConstraints}} identified one source of arbitrary functional freedom that is known a priori for linear DE systems, namely the additive solution sector $\mathcal{A}_{\mathrm{sol}}$. Since this sector can be isolated directly within the represented determining kernel, we define the admissible space used by the preflight test as
\begin{equation}
    \label{eqn:refinementOfKernLinVSNonlinSys}
    \widehat{N}_{\mathrm{adm},i}^{(k)}\coloneqq
    \begin{cases}
        \widehat{N}_{\mathrm{nadd},i}^{(k)},
        & \text{linear systems}, \\[1mm]
        \widehat{N}_{\Delta,i}^{(k)},
        & \text{nonlinear systems}.
    \end{cases}
\end{equation}
The first branch removes the represented additive solution sector before the stabilization test is applied. As emphasized in \textit{\nameref*{subsec:intrAdmConstraints}} and \ref{app:geomViewPntAndFormalInteg}, however, this does not preclude other forms of arbitrary functional freedom in the exact determining system. At every finite refinement level, $\widehat{N}_{\mathrm{adm},i}^{(k)}$ remains finite-dimensional; if unresolved functional freedom is present in the underlying determining equations, its numerical signature may instead appear as continued growth of the represented admissible dimension under refinement.

Raw determining nullity alone is not sufficient for this purpose. It may increase because of additive directions, coefficient nonidentifiability, or sampled-trivial directions that do not represent additional intrinsic geometric symmetries. We therefore also evaluate the sampled prolonged-characteristic operator associated with each refinement level, $\widehat{\mathcal{T}}_{\mathrm{triv},i}^{(k)}$, restricted to $\widehat{N}_{\mathrm{adm},i}^{(k)}$, and define
\begin{equation}
    \label{eqn:geometricRankOfDetSys}
    \begin{split}
        d_{o_i} &\coloneqq
        \operatorname{rank}
        \paraBrTerm{
            \left. \widehat{\mathcal{T}}_{\mathrm{triv},i}^{(k)}\right\vert_{\widehat{N}_{\mathrm{adm},i}^{(k)}}
        } \\
        &=
        \dim
        \paraBrTerm{\widehat{N}_{\mathrm{adm},i}^{(k)} / \widehat{K}_{\mathrm{triv},i}^{(k)}}
        =
        \dim \paraBrTerm{\widehat{N}_{\mathrm{geom},i}^{(k)}},
    \end{split}
\end{equation}
where
$\widehat{K}_{\mathrm{triv},i}^{(k)} = \ker\paraBrTerm{\left. \widehat{\mathcal{T}}_{\mathrm{triv},i}^{(k)} \right\vert_{\widehat{N}_{\mathrm{adm},i}^{(k)}}}$ and $\widehat{N}_{\mathrm{geom},i}^{(k)}$ is the sampled geometric quotient defined in \textit{\nameref*{subsec:intrAdmConstraints}}. We refer to $d_{o_i}$ as the \textit{represented geometric rank} at refinement level $o_i$.

All ranks and nullspaces entering the preflight test are numerical. For a sampled matrix $A$ with singular values $\sigma_1\geq\sigma_2\geq\cdots$, we define
\begin{equation}
    \label{eqn:numericalRankPreflight}
    \operatorname{rank}_{\tau} \paraBrTerm{A}
    \coloneqq
    \#\curBarTerm{j \mid \sigma_j > \tau_{\mathrm{rank}}\sigma_1},
\end{equation}
using the same relative threshold $\tau_{\mathrm{rank}}$ throughout a refinement sequence. In practice, the stability of the inferred rank is additionally checked under a modest variation of this threshold so that a plateau is not determined by a single marginal singular value.

The refinement sequence is regarded as numerically stabilized only when the represented geometric rank remains unchanged over a prescribed number of successive refinements,
\begin{equation}
    \label{eqn:geometricRankStabilization}
    d_{o_j} = d_{o_{j+1}} = \cdots = d_{o_{j+L_{\mathrm{stab}}}},
\end{equation}
for some prescribed $L_{\mathrm{stab}}\geq 1$, while the associated numerical ranks remain robust to the chosen SVD tolerance. Such a plateau is interpreted operationally as stabilization of the represented intrinsic symmetry dimension along the selected dictionary-refinement path; it is not, by itself, a formal proof that the exact determining system possesses no additional functional freedom. If the represented geometric rank continues to increase, or if the inferred plateau is numerically unstable, the sequential discovery stage is not initiated and the framework makes no completeness claim for $\mathfrak{g}_{\mathrm{int}}$.

Although the construction above uses successive grid refinements at fixed $k$ and $r$ as the primary enrichment path, the same stabilization principle may be applied to any genuinely nested sequence of represented function spaces. In particular, interaction order or spline degree may also be enriched when warranted by the realization under consideration, provided that the resulting spaces admit a consistent nested comparison.

\subsection{Numerical Enforcement of Nontriviality}
\label{subsec:numericEnfOfNontriv}
Proposition~\ref{pro:sampledTrivQuotient} in \textit{\nameref*{subsec:intrAdmConstraints}} showed that sampled triviality is characterized by the kernel $\widehat{K}_{\mathrm{triv}}^{(k)}$ of the prolonged-characteristic operator, while the corresponding geometric information is represented by the quotient $\widehat{N}_{\mathrm{geom}}^{(k)}$. Unlike the determining and novelty conditions, however, nontriviality cannot be imposed by restricting the candidate to another linear admissible subspace. Indeed, the nontrivial feasible set $\widehat{\mathcal{V}}_{\mathrm{ntr}}^{(k)}$ defined in \eqref{eqn:sampledNonTrivSet} excludes the sampled-trivial kernel from the admissible coefficient space and therefore does not itself form a vector subspace.

Consequently, even after the linear determining, solution-independence, and novelty restrictions have been imposed, optimization must still prevent a candidate generator from collapsing toward $\widehat{K}_{\mathrm{triv}}^{(k)}$, including the zero vector field. For this reason, we retain the nonlinear loss $\mathcal{L}_{\mathrm{ntr}}$ introduced in \eqref{eqn:nonTrivCost}. The remainder of this subsection describes its numerical realization and the continuation strategy used to enforce the corresponding minimum-response condition during training.

The analytical form of $\mathcal{L}_{\mathrm{ntr}}$, including the relative prolonged-characteristic response, the minimum-magnitude requirement, and the quadratic exterior penalty $\mathcal{P}$, was introduced in \textit{\nameref*{subsec:initialFormulation}}. We therefore do not repeat its derivation here. Instead, we describe the numerical mechanisms used to evaluate and stabilize these quantities during optimization, including the lower-tail response statistic, smooth activation of the nontriviality penalty, regularization of small denominators, auxiliary speed-tail control, and continuation of the penalty parameter $\gamma$ in \eqref{eqn:quadExtPenalty}.

Let $\mathcal{R}_{\mathrm{ntr},l}\paraBrTerm{v}$ in \eqref{eqn:relativeNonTrivResponse} denote the relative prolonged-characteristic response of $v$ at $z_l^{(s)}\in\mathbf{S}^{\mathrm{on}}$. If only the mean response
\begin{equation}
    \label{eqn:meanRelProCharResp}
    \overline{\mathcal{R}}_{\mathrm{ntr}}
    \paraBrTerm{v}
    \coloneqq
    \frac{1}{N_{\mathrm{on}}}
    \sum_{l=1}^{N_{\mathrm{on}}}
    \mathcal{R}_{\mathrm{ntr},l}
    \paraBrTerm{v}
\end{equation}
is monitored, then $v$ may exhibit a strong prolonged-characteristic response over most samples while becoming nearly trivial on a non-negligible portion of the on-shell dataset, without the average necessarily revealing this behavior. We therefore define
\begin{equation}
    \label{eqn:lowerTailStatistic}
    \mathcal{R}_{\mathrm{tail}}^{\rho}\paraBrTerm{v}\coloneqq
    q_{\rho}\paraBrTerm{
        \curBarTerm{
        \mathcal{R}_{\mathrm{ntr},l}
        \paraBrTerm{v}
        }_l
    },
\end{equation}
where $q_{\rho}$ denotes the empirical $\rho$-quantile. For small $\rho$, typically $\rho\leq 0.1$, we refer to $\mathcal{R}_{\mathrm{tail}}^{\rho}$ as the \textit{lower-tail statistic}.

Rather than enforcing the nontriviality penalty uniformly throughout training, we use this statistic to construct the smooth activation gate
\begin{equation}
    \label{eqn:lowerTailStatisticGate}
    g_{\mathrm{ntr}}^{\rho}\paraBrTerm{v}\coloneqq
    \sigma
    \sqBrTerm{
        \kappa_{\mathrm{gntr}}
        \paraBrTerm{
            \tau_{\mathrm{gntr}}
            -
            \mathcal{R}_{\mathrm{tail}}^{\rho}\paraBrTerm{v}
        }
    },
\end{equation}
where $\sigma$ is the sigmoid function, while $\kappa_{\mathrm{gntr}}>0$ and $\tau_{\mathrm{gntr}}>0$ denote the gate-sharpening parameter and gate-activation threshold, respectively. Thus, $g_{\mathrm{ntr}}^{\rho}$ approaches unity when the lower tail of the relative response becomes sufficiently small and approaches zero when the sampled response remains safely away from the trivial sector.

The nontriviality loss in \eqref{eqn:nonTrivCost} is consequently evaluated in gated form as
\begin{equation}
    \label{eqn:gatedNonTrivCost}
    \mathcal{L}_{\mathrm{tail}}\coloneqq \mathcal{P}\paraBrTerm{c_{\mathrm{min}}} +
    \frac{g_{\mathrm{ntr}}^{\rho}\paraBrTerm{v}}{B}
    \sum_{z_l^{(s)}\in\mathbf{S}^{\mathrm{on}}_{B}} \mathcal{P}\paraBrTerm{c_{\mathrm{ntr},l}},
\end{equation}
where
$c_{\mathrm{ntr},l} = \tau_{\mathrm{ntr}} - \mathcal{R}_{\mathrm{ntr},l}\paraBrTerm{v}\leq0$ is the samplewise nontriviality inequality constraint introduced previously, and $c_{\mathrm{min}}$ is the minimum-magnitude inequality constraint in \eqref{eqn:minMagConstraint}. The lower-tail statistic therefore does not replace the samplewise nontriviality condition; rather, it controls when that condition is actively emphasized during optimization.

A separate numerical safeguard is used to discourage pathological spatial concentration of the recovered vector field. For $z_l^{(s)}\in\mathbf{S}^{\mathrm{on}}$, let $z_l=\pi_0^s\paraBrTerm{z_l^{(s)}}\in E$ and define 
\begin{equation}
    \label{eqn:pointwiseSpeed}
    s_l \coloneqq \left\Vert \left.v\right\vert_{z_l} \right\Vert_2
\end{equation} 
as the \textit{pointwise speed} of the candidate generator. During optimization, we use a smoother \textit{upper-tail statistic}. Let $I_{\eta}$ contain the indices of the largest fraction $\eta$ of the sampled speeds, with
\begin{equation}
    \label{eqn:upperSpeedTailMean}
    \overline{s}_{\eta}\coloneqq \frac{1}{\left\vert I_{\eta} \right\vert}
    \sum_{l\in I_{\eta}} s_l,
    \quad
    \left\vert I_{\eta}\right\vert = \left\lceil\eta B\right\rceil .
\end{equation}
The corresponding \textit{speed-concentration penalty} is
\begin{equation}
    \label{eqn:speedTailRegularizer}
    \mathcal{L}_{\mathrm{spk}} \coloneqq
        \max\paraBrTerm{
            0, \;
            \log \paraBrTerm{ \frac{\overline{s}_{\eta}} {\texttt{sg}\paraBrTerm{q_{50}}} } - \log\rho_{\mathrm{spk}}
        }^2,
\end{equation}
where $\texttt{sg}$ denotes a \textit{stop-gradient operation} and $\rho_{\mathrm{spk}}>1$ is the tolerated upper-tail ratio. Detaching the median prevents the optimizer from reducing the upper-tail ratio by increasing the typical speed over the sampled domain; consequently, the gradient of this regularizer acts only to suppress the localized high-speed tail. The nontriviality objective is therefore augmented as
\begin{equation}
    \label{eqn:nonTrivWithSpeedTail}
    \mathcal{L}_{\mathrm{ntr}}^{\mathrm{reg}} \coloneqq
    \mathcal{L}_{\mathrm{tail}} + \lambda_{\mathrm{spk}} \mathcal{L}_{\mathrm{spk}},
\end{equation}
with $\lambda_{\mathrm{spk}}>0$ controlling the strength of this auxiliary safeguard. In the present implementation, $\eta=0.01$, so that the regularizer acts on the mean of the largest $1\%$ of sampled speeds.

\begin{remark}
    \label{rem:supprLocalAmplSpikes}
    Although the speed-concentration penalty \eqref{eqn:speedTailRegularizer} does not directly constrain derivatives of the recovered vector field, it suppresses localized amplitude spikes that, in our experiments, were associated with strong point-to-point variations and unstable numerical integration of corresponding finite flow $\operatorname{Fl}_v^z(\epsilon)$.
\end{remark}

Finally, the quadratic exterior penalty $\mathcal{P}$ introduced in \eqref{eqn:quadExtPenalty} is enforced through a continuation strategy in its penalty parameter $\gamma$. Rather than imposing the final penalty strength from the beginning of training, $\gamma$ is initialized at a smaller value and increased progressively toward its prescribed final value as optimization proceeds. This allows the candidate generator to first move toward a favorable region of parameter space before the nontriviality inequalities are enforced more sharply, thereby reducing numerical stiffness during the early stages of training. The precise continuation schedule and associated implementation parameters are provided in the accompanying source code.

\subsection{Metric-Normalized Novelty Restriction}
\label{subsec:metricNormNoveltyRestr}
The metric-normalized novelty construction, including $W_E$, the normalized stacked responses in \eqref{eqn:ambientRespMetric}, and the sampled novelty operator $\widehat{\operatorname{B}}_{E,i}^{(k)}$, was introduced in \textit{\nameref*{subsec:seqDiscoveryAndNoveltyConstraints}}. We therefore restrict the present discussion to the scale factors $d_{\boldsymbol{\xi}}>0$ and $d_{\boldsymbol{\varphi}}>0$ deferred from that section. The horizontal and vertical zeroth-order ambient responses over $\mathbf{S}^{\mathrm{off}}_{\mathrm{anc}}$ are assigned the RMS scales $s_{\boldsymbol{\xi}}$ and $s_{\boldsymbol{\varphi}}$ according to \eqref{eqn:blockRMSScale}--\eqref{eqn:blockRMSScaleEMA}, using the pooled responses of the current candidate and previously discovered generators. We then set
\begin{equation}
    \label{eqn:ambientMetricScales}
    d_{\boldsymbol{\xi}}\coloneqq\paraBrTerm{s_{\boldsymbol{\xi}}+\epsilon}\sqrt{p},
    \quad
    d_{\boldsymbol{\varphi}}\coloneqq\paraBrTerm{s_{\boldsymbol{\varphi}}+\epsilon}\sqrt{q},
\end{equation}
where the dimensional factors compensate for the natural growth of the Euclidean block norm with the number of components.

The scale estimates are updated periodically through the EMA introduced in \eqref{eqn:blockRMSScaleEMA}. At each refresh, the updated values are detached and used to construct a common metric $W_E$, after which the metric-dependent novelty restriction in \eqref{eqn:noveltyRestriction} is rebuilt. Both $W_E$ and the resulting restriction are then held fixed until the next scheduled refresh. The same frozen $W_E$ is applied to the previously recovered responses and to the current candidate-response operator, ensuring that both are compared within the same normalized ambient geometry.

\subsection{Sequential Recovery and Stopping Criteria}
\label{subsec:seqRecoveryAndStopping}
\textit{\nameref*{subsec:seqDiscoveryAndNoveltyConstraints}} established the construction of the \textit{search subspace} $\widehat{N}_{\mathrm{nov},i}^{(k)}$, but did not specify the sequential algorithm by which the generators of $\mathfrak{g}_{\mathrm{int}}$ are recovered one at a time or the conditions under which this process terminates. The present subsection addresses both aspects from an algorithmic perspective. The complete sequential discovery lifecycle, including initialization, optimization, validation, acceptance, novelty-space reconstruction, and termination branches, is summarized schematically in \ref{app:flowChartsForAlgorithms}.

We assume throughout that $\Delta^{(s)}=\mathbf{0}$ lies within the applicability scope specified in Section~\ref{sec:discApplAndLimit} and, whenever required, has passed the RSP test in \textit{\nameref*{subsec:rankStabPreflightTest}}. For the selected dictionary level, $\widehat{N}_{\Delta}^{(k)}$ and subsequently $\widehat{N}_{\mathrm{nadd}}^{(k)}$ are constructed from $\mathbf{S}_{\mathrm{anc}}^{\mathrm{on}}$ before sequential optimization begins, as described in \textit{\nameref*{subsec:operatorTheoreticSymRep}} and \textit{\nameref*{subsec:intrAdmConstraints}}. For nonlinear systems, where no additive solution sector is removed, we set $\widehat{N}_{\mathrm{nadd}}^{(k)}=\widehat{N}_{\Delta}^{(k)}$; the DE classes for which this realization is applicable are discussed in Section~\ref{sec:discApplAndLimit}. At discovery step $i=1,2,\dotsc$, the initial linear search subspace is then
\begin{equation}
    \label{eqn:seqInitialSearchSubspace}
    \widehat{N}_{\mathrm{nov},i}^{(k)}
    =
    \begin{cases}
        \widehat{N}_{\mathrm{nadd}}^{(k)}, & i=1, \\[1mm] 
        \widehat{N}_{\mathrm{nov},i}^{(k)}\; \text{in \eqref{eqn:noveltyRestriction}}, & i>1.
    \end{cases}
\end{equation}
Thus, only the novelty restriction evolves from one accepted generator to the next, while the determining and nonadditive admissible restrictions remain fixed at the chosen dictionary level throughout the entire discovery of $\mathfrak{g}_\mathrm{int}$.

No additional solvability-based Lie-bracket restriction is imposed on the successively recovered generators. As shown in \ref{app:whyNotImposingSolvability}, enforcing a normal-subalgebra chain during sequential recovery can make admissibility depend on the numerically encountered recovery path and can exclude otherwise legitimate directions that remain representable by $D_{(k)}$.

Before optimization, we also evaluate the spatial concentration of a candidate initialization. Let $q_p\paraBrTerm{v_{\boldsymbol{\Theta}}}$ denote the empirical $p^\mathrm{th}$ percentile of the sampled pointwise speeds defined in \eqref{eqn:pointwiseSpeed}. We define the \textit{speed-tail ratio}
\begin{equation}
    \label{eqn:speedTailRatio}
    \rho_{99}\paraBrTerm{v_{\boldsymbol{\Theta}}}\coloneqq
        \frac{
            q_{99}\paraBrTerm{v_{\boldsymbol{\Theta}}}
        }{
            q_{50}\paraBrTerm{v_{\boldsymbol{\Theta}}}
        }.
\end{equation}
Large values of $\rho_{99}$ indicate that the vector-field magnitude is concentrated over a comparatively small portion of the sampled region. Although $\rho_{99}$ does not directly constrain derivatives of the vector field, such concentration was associated in our numerical experiments with strong point-to-point variations and poor numerical integration of the corresponding finite flow (see~\ref{rem:supprLocalAmplSpikes}). We therefore also use it as an initialization safeguard.

To initialize the $i^\mathrm{th}$ candidate, let $U_i$ denote an orthonormal coefficient-space basis of $\widehat{N}_{\mathrm{nov},i}^{(k)}$ and draw
\begin{equation}
    \label{eqn:seqCandidateInitialization}
    \mathbf{a}_i \sim \mathcal{N}\paraBrTerm{\mathbf{0}, I_{\dim\paraBrTerm{\widehat{N}_{\mathrm{nov},i}^{(k)}}}}.
\end{equation}
The initial coefficient vector is then
\begin{equation}
    \label{eqn:initialCoeffVec}
    \boldsymbol{\Theta}_i \coloneqq U_i \frac{\mathbf{a}_i}{\left\Vert \mathbf{a}_i\right\Vert_2},
\end{equation}
so that $\boldsymbol{\Theta}_i$ lies directly in the current linear search subspace. As discussed in \ref{app:splineDictRedAndSamplingIndRankDef}, the realization $\Psi^{(k)}$ need not be injective: distinct coefficient vectors may represent the same vector field because of exact or sampling-induced dependencies in $D_{(k)}$. This pathological situation is called \textit{dictionary nonidentifiability}. Consequently, the magnitude of $\boldsymbol{\Theta}_i$ alone is not a reliable measure of the visibility of its realized vector field. 

An initialization is admitted only if 
\begin{enumerate*}[label=\textit{(\roman*)}]
    \item its ambient response is visible and 
    \item its speed distribution is not excessively concentrated.
\end{enumerate*}
We therefore define the \textit{off-shell ambient-response magnitude}
\begin{equation}
    \label{eqn:initAmbientVisibility}
    \begin{split}
        &\operatorname{RMS}_{E}\paraBrTerm{v_{\boldsymbol{\Theta}_i}}
        \coloneqq \\
        &\qquad
        \sqBrTerm{
            \frac{1}{A_{\mathrm{off}}} 
            \sum_{z_a^{(s)}\in\mathbf{S}^{\mathrm{off}}_{\mathrm{anc}}}
            \left\Vert \left. v_{\boldsymbol{\Theta}_i}\right\vert_{\pi_0^s\paraBrTerm{z_a^{(s)}}}\right\Vert_2^2
        }^{1/2}.
    \end{split}
\end{equation}
The initialization is accepted when
\begin{equation}
    \label{eqn:initCandidateAcceptance}
    \operatorname{RMS}_{E}\paraBrTerm{v_{\boldsymbol{\Theta}_i}} > \tau_{\mathrm{vis}},
    \quad
    \rho_{99}\paraBrTerm{v_{\boldsymbol{\Theta}_i}} < \tau_{\mathrm{spd}},
\end{equation}
where $\tau_{\mathrm{vis}}>0$ and $\tau_{\mathrm{spd}}>0$ denote the \textit{visibility} and \textit{maximum speed-tail} thresholds, respectively. Random initialization is repeated until these conditions are satisfied or the prescribed initialization budget is exhausted; the latter case is incorporated into the stopping logic discussed below.

During optimization, the metric-dependent novelty restriction is periodically refreshed every $t_{\mathrm{nov}}$ training steps using the most recent normalization statistics entering $W_E$, following the construction in \textit{\nameref*{subsec:metricNormNoveltyRestr}}. Since this refresh generally changes the orthonormal basis $U_i$ of $\widehat{N}_{\mathrm{nov},i}^{(k)}$, the current coefficient vector $\boldsymbol{\Theta}_i$ is reprojected onto the updated search subspace. Between successive refreshes, let
\begin{equation}
    \label{eqn:noveltySubspaceProjector}
    P_i \coloneqq U_iU_i^{\mathsf T}
\end{equation}
denote the orthogonal projector onto $\widehat{N}_{\mathrm{nov},i}^{(k)}$. At every training step $t$, the optimizer update $\delta\boldsymbol{\Theta}_i^t$ is replaced by
\begin{equation}
    \label{eqn:projectedOptimizerUpdate}
    \delta\boldsymbol{\Theta}_i^t \longmapsto P_i \delta\boldsymbol{\Theta}_i^t,
\end{equation}
so that optimization evolves only along directions permitted by the current determining, solution-independence, and novelty restrictions.

The regularized nontriviality objective $\mathcal{L}_{\mathrm{ntr}}^{\mathrm{reg}}$ in \eqref{eqn:nonTrivWithSpeedTail} is then optimized until the candidate $v_{\boldsymbol{\Theta}_i}$ satisfies the \textit{principal acceptance criteria}
\begin{equation}
    \label{eqn:cndMetByCandGen}
    \begin{split}
        \operatorname{RMS}_{E}\paraBrTerm{v_{\boldsymbol{\Theta}_i}} &>\tau_{\mathrm{vis}}, \\
        \rho_{99}\paraBrTerm{v_{\boldsymbol{\Theta}_i}} &< \tau_{\mathrm{spd}}, \\
        \mathcal{L}_{\mathrm{sym}}^{\mathrm{raw}} &< \tau_{\mathrm{sym}}, \\
        \mathcal{R}_{\mathrm{tail}}^{\rho}\paraBrTerm{v_{\boldsymbol{\Theta}_i}} &> \tau_{\mathrm{ntr}},
    \end{split}
\end{equation}
on the corresponding validation samples. Here, the ambient-response visibility criterion is evaluated on an off-shell validation set, whereas the speed-tail, determining-residual, and prolonged-characteristic criteria are evaluated on on-shell validation samples. The first two conditions ensure that the candidate remains both numerically visible and sufficiently well resolved across the sampled region; the latter two verify approximate symmetry admissibility and nontriviality independently of the training samples. Once all conditions in \eqref{eqn:cndMetByCandGen} are satisfied, $v_{\boldsymbol{\Theta}_i}$ is accepted as the $i^\mathrm{th}$ recovered generator and incorporated into the construction of $\widehat{N}_{\mathrm{nov},i+1}^{(k)}$ for the subsequent discovery step.

Finally, we specify the stopping criteria for the sequential discovery procedure. Two qualitatively distinct termination modes are distinguished:
\begin{enumerate*}[label=\textit{(\roman*)}]
    \item \textit{clean exhaustion} of the represented novel directions and
    \item \textit{recovery failure} while a represented novel direction remains.
\end{enumerate*}
This distinction is essential because only the first mode supports the conclusion that the available symmetry directions have been exhausted within the chosen stabilized representation.

The clean-exhaustion criterion is met immediately when $\dim\paraBrTerm{\widehat{N}_{\mathrm{nov},i}^{(k)}} = 0$. In this case, all represented admissible directions have already been accounted for by the previously accepted generators, up to the numerical rank tolerances entering the novelty construction, and no further initialization is attempted. Clean exhaustion may also be declared when
$\dim\paraBrTerm{\widehat{N}_{\mathrm{nov},i}^{(k)}}>0$ in coefficient space, but repeated sampling within this subspace, over a prescribed initialization budget, fails to produce any candidate satisfying $\operatorname{RMS}_{E}\paraBrTerm{v_{\boldsymbol{\Theta}_i}} > \tau_{\mathrm{vis}}$. Such a situation indicates that the remaining coefficient directions are numerically negligible under the realized ambient-response map, for example because of residual dictionary nonidentifiability or sampling-induced rank deficiency. The sequential procedure is then terminated and the already accepted generators are taken to exhaust the numerically resolvable symmetry directions available within the chosen stabilized representation.

Failure of the second initialization condition in \eqref{eqn:initCandidateAcceptance} has a different interpretation. If candidates with $\operatorname{RMS}_{E}\paraBrTerm{v_{\boldsymbol{\Theta}_i}}>\tau_{\mathrm{vis}}$ can be produced, but repeated initialization yields only fields satisfying $\rho_{99}\paraBrTerm{v_{\boldsymbol{\Theta}_i}}\geq\tau_{\mathrm{spd}}$, then represented novelty is still present, but no initialization with the required numerical regularity has been found. This case is therefore classified as a recovery failure rather than as exhaustion of the represented space.

The same recovery-failure status is assigned when an initialization satisfying both conditions in \eqref{eqn:initCandidateAcceptance} is admitted to optimization, but no trained candidate satisfies all acceptance criteria in \eqref{eqn:cndMetByCandGen} within the prescribed optimization budget. Such failure may arise, for example, because the determining residual remains above $\tau_{\mathrm{sym}}$, the prolonged-characteristic response remains below the required nontriviality threshold, or the speed-tail criterion becomes unacceptable during optimization. In these cases, the existence of a visible direction in $\widehat{N}_{\mathrm{nov},i}^{(k)}$ prevents the algorithm from interpreting unsuccessful recovery as structural exhaustion.

Accordingly, only clean exhaustion permits the sequential algorithm to conclude that the represented intrinsic symmetry directions have been exhausted at the selected dictionary level and numerical tolerances. Termination by recovery failure makes no such statement: the generators accepted before the failed step remain valid recovered directions, but no completeness claim is made for the remaining represented symmetry space.

\section{Numerical Experiments}
\label{sec:experiments}

This section evaluates the proposed framework on the two-dimensional spatial wave equation, using the final manuscript-aligned implementation described in Section~\ref{sec:implAndCompConsid}. The numerical study is designed to examine not only the sequential recovery of the represented intrinsic symmetry directions, but also the behavior of the progressively restricted search space, the algebraic consistency of the recovered generators, and the locality of the fixed dictionary representation. We first introduce the benchmark and experimental configuration, then present the sequential discovery results, and finally examine the reconstruction and extrapolation behavior of a known analytic symmetry generator.

\subsection{Benchmark and Experimental Configuration}
\label{subsec:benchAndExpConfig}
This subsection introduces the DE system used as the principal numerical benchmark and explains why it provides a sufficiently rich test of the structural components of the proposed framework while retaining a well-understood analytic symmetry structure for validation. We then summarize the common numerical configuration used throughout the experiments.

We consider the \textit{wave equation in two spatial dimensions}~\cite[Chapter~10]{haberman1983elementary},
\begin{equation}
    \label{eqn:waveEqn2D}
    \Delta^{(2)}\paraBrTerm{x,y,t;u_{(2)}} = u_{tt} - c^2\paraBrTerm{u_{xx}+u_{yy}} = 0,
\end{equation}
with the normalized wave speed $c=1$ in all experiments. Equation~\eqref{eqn:waveEqn2D} is a regular linear PDE of genuine order $s=2$ and satisfies the hypotheses of Proposition~\ref{pro:projFibAffSym}. Hence, every admitted Lie point-symmetry generator has the PFA form in~\eqref{eqn:projFibAffSym}. Moreover, under the dictionary-refinement sequence specified below, the represented geometric rank $d_{o_i}$ defined in~\eqref{eqn:geometricRankOfDetSys} stabilizes, so the benchmark passes the RSP criterion described in \textit{\nameref*{subsec:rankStabPreflightTest}}. It therefore lies within the admissible scope summarized in \textit{\nameref*{subsec:applAndAdmissibleGenRealizations}}.

The Lie point-symmetry structure of~\eqref{eqn:waveEqn2D} is also known analytically~\cite[Example~2.43]{olver2000applications}. After separating the additive solution sector associated with linear superposition, its finite-dimensional intrinsic symmetry algebra $\mathfrak{g}_{\mathrm{int}}$ has dimension $11$. At the same time, linearity introduces the nontrivial additive solution sector $\mathcal{A}_{\mathrm{sol}}$ described in \textit{\nameref*{subsec:intrAdmConstraints}}, which must be removed from the represented determining nullspace before the RSP and intrinsic sequential discovery are applied. The benchmark therefore exercises, within a single DE system, the sampled determining restriction, additive-sector exclusion, represented-rank stabilization, nontriviality criterion, global novelty restriction, and graceful stopping mechanism. Its known finite-dimensional symmetry structure additionally provides an independent reference for validating the recovered dimension, Lie-algebraic relations, and individual reconstructed symmetry directions.

For all experiments, we generated samples directly on the equation shell of~\eqref{eqn:waveEqn2D}, rather than from a particular solution graph. The independent coordinates $\mathbf{x}$ were sampled on the uniform patch
\begin{equation}
    \label{eqn:waveEqnExpDomain}
    \Omega_{\mathbf{X}} = \sqBrTerm{0.30,0.36} \times \sqBrTerm{0.30,0.36} \times \sqBrTerm{0.25,0.256},
\end{equation}
using $101\times101\times11$ points in $\paraBrTerm{x,y,t}$, corresponding to a common spacing of $6\times10^{-4}$ and $112{,}211$ jet-space samples. At each base point, the remaining jet-fiber coordinates were drawn independently from a standard normal distribution and subsequently projected onto the equation shell. For~\eqref{eqn:waveEqn2D} with $c=1$, this projection is obtained directly by imposing $u_{tt}=u_{xx}+u_{yy}$.

Each on-shell sample was paired with an off-shell sample obtained by perturbing the corresponding jet-space point along a randomly sampled unit direction with \textit{perturbation magnitude} $10^{-2}$. The resulting paired dataset was randomly divided into $80\%$ training and $20\%$ validation subsets, yielding $89{,}769$ and $22{,}442$ samples, respectively. The on-shell $\mathbf{S}^{\mathrm{on}}$ and paired off-shell $\mathbf{S}^{\mathrm{off}}$ components were subsequently supplied to the corresponding sampled operators and constraints described in Sections~\ref{sec:framework} and~\ref{sec:implAndCompConsid}, without restricting the samples to a single solution manifold.

For the fixed generator realization, we used the dictionary $D_{(2)}$ introduced in \textit{\nameref*{subsec:fixedDictAndGenRealization}}, with the independent coordinates $\boldsymbol{\chi}=\paraBrTerm{x,y,t}$ as inputs, maximum interaction order $k=2$, cubic B-splines with degree $r=3$, and grid resolution $o=8$. The fixed core-grid intervals were $\sqBrTerm{-0.66,1.34}$ for both $x$ and $y$ and $\sqBrTerm{-1.05,0.95}$ for $t$. Thus, $K_o=o+r=11$ univariate B-spline functions were associated with each input coordinate, and the univariate and pairwise-interaction blocks together yielded $n_{(2)}^{D}=396$ scalar dictionary features. Under the PFA realization appropriate to~\eqref{eqn:waveEqn2D}, the three horizontal coefficients together with the multiplicative and additive vertical coefficients therefore resulted in a total coefficient dimension of $P=1980$.

Before sequential discovery, we applied the RSP test of \textit{\nameref*{subsec:rankStabPreflightTest}} over the nested grid-refinement sequence $o_1=4<o_2=8<o_3=16$, while keeping the spline degree, interaction order, and physical core-grid intervals fixed. Numerical ranks were evaluated using a relative SVD threshold of $10^{-10}$; a tolerance sweep showed that this tighter threshold was required to resolve the determining-operator rank at the coarsest refinement level, while the resulting rank was unchanged upon further reduction to $10^{-12}$. The represented geometric rank remained constant across all three refinements,
\begin{equation}
    d_{4}=d_{8}=d_{16}=11,
\end{equation}
thereby satisfying the stabilization criterion used to proceed with the nominal $o=8$ realization in the subsequent experiments.

\subsection{Sequential Discovery of Intrinsic Symmetries}
\label{subsec:seqDiscoveryOfIntrSymmetries}
Having established stabilization of the represented geometric rank, we fix the nominal resolution $o=8$ and examine the sequential recovery of the corresponding intrinsic symmetry directions. Before any generator optimization is performed, the sampled determining restriction $\widehat{M}_{\Delta}^{(2)}$ has rank $231$ in the $P=1980$-dimensional coefficient representation, leaving a determining nullspace $\widehat{N}_{\Delta}^{(2)}$ of dimension $1749$. The large coefficient-space nullity reflects the substantial redundancy of the spline representation discussed in \ref{app:splineDictRedAndSamplingIndRankDef} and should therefore not be interpreted directly as the number of geometrically distinct symmetries. After removal of the represented additive solution sector, the resulting additive-free coefficient representation has dimension $11$, in agreement with both the RSP result of the preceding subsection and the known intrinsic symmetry dimension of~\eqref{eqn:waveEqn2D}.

This reduction can also be verified at the response level. The sampled characteristic responses of the full determining nullspace, the represented additive sector, and the resulting intrinsic sector have numerical ranks $29$, $18$, and $11$, respectively. In particular, the intrinsic response rank remains equal to $11$ over relative SVD thresholds from $10^{-4}$ to $10^{-12}$, indicating that the finite-dimensional target supplied to the sequential discovery stage is numerically well separated from the redundant coefficient directions. We therefore begin the first optimization step with an $11$-dimensional novelty-restricted admissible space and remove one globally novel direction after each accepted generator.

\begin{table*}[!t]
    \centering
    \caption{Sequential recovery of the represented intrinsic symmetry directions for the two-dimensional wave equation at the nominal grid resolution $o=8$. Here, $\dim\paraBrTerm{T_i}$ denotes the dimension of the novelty-restricted coefficient space immediately before optimization of the $i^{\mathrm{th}}$ generator. The validation symmetry RMS and speed-tail ratio are evaluated for the accepted generator on the held-out validation set. Each accepted generator removes one globally novel direction; after the eleventh generator, the novelty-restricted space collapses to zero and the procedure terminates by clean represented exhaustion.}
    \label{tab:waveEqnSeqDiscovery}
    \setlength{\tabcolsep}{5.5pt}
    \renewcommand{\arraystretch}{1.12}
    \begin{tabular}{cccc}
        \toprule
        \textbf{Generator} &
        $\boldsymbol{\dim\paraBrTerm{T_i}}$ &
        \textbf{Validation symmetry RMS} &
        \textbf{Speed-tail ratio} \\
        \midrule
        $v_1$  & $11$ & $3.33\times10^{-15}$ & $1.037$ \\
        $v_2$  & $10$ & $5.68\times10^{-15}$ & $1.011$ \\
        $v_3$  & $9$  & $2.77\times10^{-15}$ & $1.003$ \\
        $v_4$  & $8$  & $5.24\times10^{-15}$ & $1.002$ \\
        $v_5$  & $7$  & $4.92\times10^{-14}$ & $1.648$ \\
        $v_6$  & $6$  & $7.86\times10^{-14}$ & $1.661$ \\
        $v_7$  & $5$  & $8.76\times10^{-14}$ & $1.654$ \\
        $v_8$  & $4$  & $7.32\times10^{-14}$ & $1.658$ \\
        $v_9$  & $3$  & $1.10\times10^{-13}$ & $1.651$ \\
        $v_{10}$ & $2$ & $5.85\times10^{-14}$ & $1.993$ \\
        $v_{11}$ & $1$ & $1.06\times10^{-13}$ & $1.969$ \\
        \bottomrule
    \end{tabular}
\end{table*}

As shown in Table~\ref{tab:waveEqnSeqDiscovery}, all eleven accepted generators satisfy the sampled determining condition to near machine precision on the independent validation set. The validation symmetry RMS remains between approximately $2.8\times10^{-15}$ and $1.1\times10^{-13}$ throughout the complete discovery sequence, several orders of magnitude below the prescribed acceptance threshold $\tau_{\mathrm{sym}}=2\times10^{-8}$. At the same time, each accepted generator removes exactly one globally novel represented direction: $\dim\paraBrTerm{T_i}$ decreases successively from $11$ to $1$ as $v_1,\ldots,v_{11}$ are recovered. Thus, the progressive novelty restriction remains consistent with the represented geometric rank $d_8=11$ identified before optimization, rather than producing an early loss or duplication of an admissible direction.

The speed-tail ratio $\rho_{99}$ introduced in \textit{\nameref*{subsec:seqRecoveryAndStopping}} provides an additional view of how the recovered directions change as this novelty space is progressively exhausted. For the first four generators, $\rho_{99}$ remains close to unity, ranging from approximately $1.002$ to $1.037$, indicating an almost uniform distribution of pointwise generator speeds over the sampled domain. This transition has a natural interpretation from the known analytical symmetry structure of~\eqref{eqn:waveEqn2D}: the intrinsic algebra contains the particularly simple four-dimensional subspace
\begin{equation}
    \operatorname{span}
    \curBarTerm{
        \partial_x,\,
        \partial_y,\,
        \partial_t,\,
        u\partial_u
    }.
\end{equation}
Once four independent directions have been recovered, the speed-tail ratio increases abruptly to approximately $1.65$ for $v_5,\ldots,v_9$, and further to approximately $1.97$--$1.99$ for $v_{10}$ and $v_{11}$. Since the sequential procedure does not constrain the recovered numerical basis to coincide with any particular analytical basis, we do not identify $v_1,\ldots,v_4$ individually with the four canonical generators above. Nevertheless, the coincidence between the dimension of this simple analytical sector and the observed transition after the fourth recovered direction is consistent with the procedure first resolving comparatively simple symmetry combinations before moving into the more spatially structured directions required to complete the represented algebra.

Nontriviality remains satisfied throughout the same sequence. The validation lower-tail nontriviality statistic exceeds the prescribed threshold $\tau_{\mathrm{ntr}}=2.0\times10^{-3}$ for every accepted generator, with values ranging from $2.002\times10^{-3}$ to $3.379\times10^{-3}$. The earlier directions generally lie more comfortably above this threshold, whereas several of the later generators approach it more closely; for example, the seventh generator yields approximately $2.002\times10^{-3}$. We do not interpret these values as defining a canonical ordering of symmetry complexity, since both the recovered basis and its sequential ordering are nonunique. Rather, together with the increase in $\rho_{99}$, they indicate that the remaining admissible directions become progressively more constrained by the simultaneous requirements of global novelty, sampled symmetry, and nontriviality.

After acceptance of $v_{11}$, the novelty-restricted represented space collapses to zero before initialization of a twelfth candidate. The run therefore terminates by clean represented exhaustion, rather than by a failure of initialization or optimization. In conjunction with the stabilized RSP value $d_4=d_8=d_{16}=11$, this provides numerical evidence that all \textit{resolvable intrinsic directions} within the chosen stabilized representation and sampling configuration have been recovered. Representative integral curves of $v_4$, $v_7$, and $v_{10}$, selected from the three speed-tail regimes identified above, are shown in Fig.~\ref{fig:waveEqnRecoveredGenIntegralCurves} of \ref{app:suppNumVisualizations}. These visualizations complement the speed-tail diagnostic by illustrating the increasingly nonuniform local flow geometry of the recovered directions, but they are not used in the numerical acceptance or stopping criteria. The remaining question is then algebraic rather than dimensional: whether the span of the eleven recovered generators is numerically closed under Lie brackets. We address this next through post-recovery bracket validation.

\begin{table*}[!t]
    \centering
    \caption{Post-recovery Lie-bracket closure of the recovered intrinsic symmetry generators for the two-dimensional wave equation.}
    \label{tab:waveEqnBracketClosure}
    \setlength{\tabcolsep}{6pt}
    \renewcommand{\arraystretch}{1.12}
    \begin{tabular}{lc}
        \toprule
        \textbf{Quantity} & \textbf{Value} \\
        \midrule
        Recovered generators & $11$ \\
        Unordered generator pairs & $55$ \\
        Fitting samples & $89\,769$ \\
        Validation samples & $22\,442$ \\
        Response-design rank & $11$ \\
        Response-design condition number & $1.55\times10^{1}$ \\
        Median held-out relative closure error & $1.64\times10^{-10}$ \\
        Maximum held-out relative closure error & $2.69\times10^{-8}$ \\
        Numerically zero brackets & $0$ \\
        \bottomrule
    \end{tabular}
\end{table*}

To assess whether the recovered span is also consistent with the Lie-algebraic structure of the intrinsic symmetry space, we performed a post-recovery bracket-closure test on off-shell samples. For each unordered pair $\paraBrTerm{v_i,v_j}$, with $i<j$, a single vector of constant coefficients $c_{ij}^{k}\in\mathbb{R}$ was fitted over the off-shell fitting set according to
\begin{equation}
    \sqBrTerm{v_i,v_j}
    \approx
    \sum_{k=1}^{11} c_{ij}^{k}v_k .
\end{equation}
The fitted coefficients were then held fixed and evaluated on the disjoint off-shell validation set. Thus, the test does not permit point-dependent coefficients to reproduce the bracket locally; rather, it directly tests whether each recovered bracket can be represented by one global element of the finite-dimensional recovered span. For each nonzero bracket pair, the \textit{held-out relative closure error} is defined by
\begin{equation}
    \label{eqn:relativeClosureError}
    \varepsilon_{ij}^{\mathrm{br}}
    \coloneqq
    \frac{
        \operatorname{RMS}
        \paraBrTerm{
            \sqBrTerm{v_i,v_j}
            -
            \sum_{k=1}^{11}c_{ij}^{k}v_k
        }
    }{
        \operatorname{RMS}
        \paraBrTerm{\sqBrTerm{v_i,v_j}}
    }.
\end{equation}

The first numerical requirement for interpreting such a fit is that the sampled responses of the recovered generators remain linearly identifiable. To this end, we form a response-design matrix whose $i^\mathrm{th}$ column contains the sampled components of the recovered generator $v_i$ over the fitting anchors. As reported in Table~\ref{tab:waveEqnBracketClosure}, this matrix has numerical rank $11$, equal to the number of recovered generators, with condition number approximately $1.55\times10^{1}$. Hence, the fitted structure coefficients are not obtained from a rank-deficient representation of the recovered basis. This is important because an underdetermined response design could otherwise produce nonunique coefficient vectors even when a small bracket residual is observed.

All $55$ unordered generator pairs were classified as numerically nonzero. A bracket was regarded as numerically zero when its RMS did not exceed $10^{-10}$ times the overall recovered-generator RMS, which for this experiment gives a zero-bracket threshold of approximately $1.00\times10^{-11}$. The smallest observed bracket RMS was approximately $5.62\times10^{-3}$ and therefore remained well above this threshold. The absence of numerically zero pairwise brackets in the recovered basis should not be interpreted as a disagreement with an analytical symmetry basis containing commuting generators. The pattern of vanishing brackets, as well as the individual structure constants, is basis dependent, and the sequential discovery procedure does not constrain the recovered generators to coincide with a canonical analytical basis. Generic linear combinations of such generators may therefore yield nonzero brackets for every recovered basis pair.

The held-out closure accuracy is summarized by the relative validation residuals. Across all $55$ brackets, the median relative error is $1.64\times10^{-10}$, while the largest value is $2.69\times10^{-8}$, attained for $\sqBrTerm{v_7,v_9}$. Importantly, these errors are evaluated using the same constant structure coefficients determined solely from the fitting samples; no refitting is performed on the validation set. The small held-out residuals therefore indicate that the fitted bracket relations persist across disjoint off-shell samples rather than merely reproducing the points used to determine the coefficients.

Taken together with the full recovered response rank and the preceding clean represented exhaustion, these results provide numerical evidence that the recovered $11$-dimensional span is closed under Lie brackets within the sampled domain and to the numerical accuracy supported by the chosen finite representation. Accordingly, the recovered generators behave numerically as a finite-dimensional Lie algebra, rather than merely as eleven independently admissible symmetry directions. This conclusion is a sampled numerical closure statement and is not intended as an exact symbolic verification of the structure constants.

\subsection{Approximation of a Known Symmetry Generator and Locality of the Representation}
\label{subsec:approxOfKnownSymGenAndLocalityOfRep}
The preceding experiments establish that the sequential procedure recovers the full represented intrinsic dimension and that the resulting span is numerically closed under Lie brackets. We finally examine two complementary properties that are not determined by these tests alone: whether the recovered numerical basis can accurately reconstruct a specific known analytical symmetry direction, and how far such a reconstruction can be continued away from the region in which the fixed dictionary was sampled and fitted. This provides both an independent check on the content of the recovered span and a direct illustration of the local character of the finite dictionary representation.

As a reference direction, we select the analytical inversion-in-$x$ symmetry of~\eqref{eqn:waveEqn2D}~\cite[(2.65)]{olver2000applications},
\begin{equation}
    \label{eqn:waveEqnInvXGen}
    v_{\mathrm{inv},x} =
    \paraBrTerm{x^2-y^2+t^2}\partial_x + 2xy\,\partial_y + 2xt\,\partial_t - xu\,\partial_u .
\end{equation}
Because the sequential discovery procedure does not constrain the recovered generators to coincide with a prescribed analytical basis, the relevant question is not whether one recovered generator reproduces~\eqref{eqn:waveEqnInvXGen} individually, but whether this direction lies in their global constant-coefficient span. We therefore construct
\begin{equation}
    \label{eqn:waveEqnInvXSpanApprox}
    \widetilde{v}_{\mathrm{inv},x} \coloneqq \sum_{i=1}^{11} a_i v_i,
    \quad
    a_i\in\mathbb{R}.
\end{equation}
and determine the coefficients $a_i$ by a global least-squares fit while keeping all recovered generators fixed. The fit uses both the original on-shell samples and their paired off-shell perturbations, so that the resulting coefficient vector represents a single global combination of the recovered basis rather than a point-dependent local reconstruction.

We first quantify the reconstruction accuracy on the fitting and held-out validation data and examine the pointwise agreement of $v_{\mathrm{inv},x}$ and $\widetilde{v}_{\mathrm{inv},x}$ in the vicinity of the sampled domain. We then compare their finite integral curves as the flow is continued beyond this region. The latter comparison is intended to probe the locality of the fixed spline representation rather than the membership of the analytical generator in the recovered span, and it provides the numerical motivation for the patchwise transport strategy discussed below.

\begin{table*}[!t]
    \centering
    \caption{Global constant-span reconstruction of the analytical inversion-in-$x$ generator from the eleven recovered generators.}
    \label{tab:invXSpanFit}
    \setlength{\tabcolsep}{7pt}
    \renewcommand{\arraystretch}{1.12}
    \begin{tabular}{lccc}
        \toprule
         &
        \textbf{Relative RMS} &
        \textbf{Absolute RMS} &
        \textbf{Cosine similarity} \\
        \midrule
        Fitting &
        $4.12\times10^{-10}$ &
        $9.82\times10^{-10}$ &
        $1.000$ \\
        Validation &
        $4.09\times10^{-10}$ &
        $9.89\times10^{-10}$ &
        $1.000$ \\
        \bottomrule
    \end{tabular}
\end{table*}

Table~\ref{tab:invXSpanFit} summarizes the resulting constant-span reconstruction. The fitting problem is strongly overdetermined, with $1.30\times10^{6}$ sampled scalar equations for the $11$ coefficients in~\eqref{eqn:waveEqnInvXSpanApprox}. The corresponding response matrix has full column rank $11$ at the relative SVD threshold $10^{-10}$ and condition number approximately $3.05\times10^{3}$; no ridge regularization is applied. The relative RMS reported in the table is the absolute RMS reconstruction residual normalized by the RMS magnitude of the analytical target, while the cosine similarity measures the directional alignment of the stacked analytical and reconstructed vector-field responses.

The reconstruction error is essentially unchanged when the fitted coefficients are transferred to the held-out samples. On the fitting data, the relative and absolute RMS errors are $4.12\times10^{-10}$ and $9.82\times10^{-10}$, respectively, while the corresponding validation values are $4.09\times10^{-10}$ and $9.89\times10^{-10}$. The cosine similarity is $1.000$ to the reported precision for both sets. Importantly, the coefficients $a_i$ are determined only from the fitting data and remain fixed during validation. Thus, a single global linear combination of the recovered generators reproduces the known inversion-in-$x$ symmetry to approximately $10^{-10}$ relative accuracy on independent samples in the neighborhood represented by the fixed dictionary.

This agreement is also evident at the level of the induced finite flows. Fig.~\ref{fig:waveEqn2D_custSymGenSpanFit_InversionX_integCurves2D} in \ref{app:suppNumVisualizations} compares the integral curves generated by $v_{\mathrm{inv},x}$ and $\widetilde{v}_{\mathrm{inv},x}$. The initial-point grid spans $x,y\in\sqBrTerm{0.2,0.4}$, $t\in\sqBrTerm{0.2,0.3}$, and $u\in\sqBrTerm{-0.6,0.6}$. In particular, the independent-coordinate ranges already extend substantially beyond the base-coordinate intervals used to generate the fitting and validation data in \eqref{eqn:waveEqnExpDomain}. The trajectories are integrated up to $\varepsilon_{\max}=0.1$ using $100$ steps. Across all six displayed coordinate-plane projections, the reconstructed and analytical trajectories are visually indistinguishable at the plotted resolution, consistent with the pointwise reconstruction errors reported in Table~\ref{tab:invXSpanFit}.

This provides an independent numerical check that the recovered span contains an accurate representation of this analytical intrinsic symmetry direction, while leaving open the separate question of how that representation behaves when the flow carries points beyond the region in which the fixed dictionary is reliably constrained.

To examine the extent and eventual limitations of this extrapolation, we next repeat the finite-flow comparison after shifting only the initialization interval associated with the $x$ coordinate. Relative to the preceding visualization, the $y$, $t$, and $u$ ranges are left unchanged, while the initial $x$ values are moved to $\sqBrTerm{0.8,0.86}$. Fig.~\ref{fig:waveEqn2D_custSymGenSpanFit_InversionX_integCurves2D_shiftedX} in \ref{app:suppNumVisualizations} compares the analytical and reconstructed flows over the shorter interval $\varepsilon_{\max}=0.04$. Despite this reduced integration horizon, the departure between the two representations is now readily visible. In the $x$--$y$ and $x$--$t$ projections, both the directions and the extents of the reconstructed trajectories deviate from their analytical counterparts, whereas the remaining projections primarily exhibit a mismatch in trajectory scale. The contrast with Fig.~\ref{fig:waveEqn2D_custSymGenSpanFit_InversionX_integCurves2D} therefore shows that, although the recovered representation remains remarkably accurate well beyond the sampled $x$ interval, this extrapolation should not be interpreted as a uniformly accurate global representation of the analytical vector field.

This behavior motivates a patchwise treatment of the recovered generators on larger domains. Rather than requiring a single fixed dictionary realization to extrapolate accurately throughout the entire domain, one may construct local representations on overlapping coordinate patches $\Omega_X^{(a)}$ and transport the already recovered generator directions across their overlaps. In such a continuation, the novelty, linear-independence, and nontriviality conditions used to identify the generators need not be reimposed independently on each patch; their role is to determine the symmetry directions during the initial recovery. Subsequent patches instead provide local representations of those same directions, with overlap consistency supplying the information required to relate \textit{neighboring realizations}. The present extrapolation experiment thus provides a direct numerical motivation for the patchwise generator-transport extension identified as future work in Section~\ref{sec:conclusion}.

\section{Discussion, Applicability, and Limitations}
\label{sec:discApplAndLimit}
Sections~\ref{sec:framework} and~\ref{sec:implAndCompConsid} establish the theoretical construction and computational machinery used to recover represented intrinsic symmetry generators. Let $\mathfrak{g}_{\mathrm{rec}}\coloneqq\operatorname{span}\curBarTerm{v_1,\dotsc,v_r}$
denote the span of the generators accepted by the proposed framework. Because the determining, nontriviality, solution-independence, and novelty conditions are enforced through finite-dimensional representations and sampled operators, $\mathfrak{g}_{\mathrm{rec}}$ should be interpreted as a numerical approximation to a subspace of the target intrinsic algebra $\mathfrak{g}_{\mathrm{int}}$, rather than as an exact algebraic identification. The preceding sections, together with \ref{app:splineDictRedAndSamplingIndRankDef}, already describe the principal representational and sampling mechanisms through which the equality $\mathfrak{g}_{\mathrm{rec}} = \mathfrak{g}_{\mathrm{int}}$ may fail. They do not, however, fully characterize the scope of the resulting completeness claim, the limitations of the RSP, or the distinction between sampled represented completeness and exact geometric completeness.

A separate issue concerns the class of DE systems to which the framework may be applied. The sequential-recovery flowchart in Fig.~\ref{fig:sequentialRecoveryFlowchart} begins by asking whether the given system $\Delta^{(s)}=\mathbf{0}$ is admissible and halts before discovery when it lies outside the intended scope. The mathematical criteria underlying that decision, including the admissible generator realization and the assumptions under which it is justified, have so far been deferred.

Accordingly, this section clarifies
\begin{enumerate*}[label=\textit{(\roman*)}]
    \item the admissible classes of DE systems, together with the structural assumptions and model restrictions required for each class;
    \item the conditions under which recovery may be regarded as geometrically complete within the represented and sampled setting, together with the limitations of such a claim; and
    \item the geometric symmetry structure described by $\mathfrak{g}_{\mathrm{rec}}$, including the distinction between infinitesimal, identity-connected, and genuinely global symmetry information.
\end{enumerate*}
Together, these questions delineate the \emph{applicability}, the scope of \emph{completeness claims}, and the \emph{geometric scope} of the proposed framework.

\subsection{Applicability and Admissible Generator Realizations}
\label{subsec:applAndAdmissibleGenRealizations}
In this subsection, we do not seek a \textit{canonical classification} of DEs, but rather several classifications for understanding the scope of our current and future framework. We distinguish an ODE from a PDE, linear DE from a nonlinear DE, and finally, the order also matters in determining the scope of this framework. We also discuss when the projectable fiber-affine (PFA) restriction, that is briefly touched in Proposition~\ref{pro:addSltnIdeal}, is mathematically legitimate and when the RSP is required. Finally, we conclude the completeness claim in accordance with all the evidence provided in this subsection.

Unlike linear DEs, nonlinear DEs do not possess the generic additive solution-symmetry sector discussed in \textit{\nameref*{subsec:intrAdmConstraints}}. In particular, the superposition principle no longer applies: linear combinations of solutions of a nonlinear DE system need not be solutions, and consequently there is no generic linear space analogous to $\mathcal{A}_{\mathrm{sol}}$ that can be projected out before sequential discovery. This does not preclude individual nonlinear DEs from admitting solution-dependent point symmetries; rather, such symmetries must be identified through the solution-independence criterion introduced in \textit{\nameref*{subsec:intrAdmConstraints}}, instead of through a universal linear nullspace projection. Moreover, as discussed in \ref{app:geomViewPntAndFormalInteg}, the corresponding determining system may retain other forms of arbitrary functional freedom.

For determined ODE systems of genuine order $s>1$, however, the unrestricted Lie point-symmetry algebra is finite-dimensional; regardless of whether the underlying ODE system is linear or nonlinear.
\begin{proposition}[Finite Dimensionality of Determined ODE Symmetries]
    \label{pro:finiteDimDetODESyms}
    Let $\Delta^{(s)}=\mathbf{0}$ be a regular, determined system of ODEs of genuine order $s\geq2$, locally solvable~\cite[Section~2.6]{olver2000applications} for its highest-order derivatives as $\mathbf{u}_{s} = \mathbf{F}\paraBrTerm{x,\mathbf{u}_{(s-1)}}$, where $\mathbf{u}\in\mathbb{R}^{q}$ and $q\geq1$. Then its local Lie point-symmetry algebra $\mathfrak{g}_{\mathrm{pt}}$ is finite-dimensional, with
    \begin{equation}
        \label{eqn:finDimOfDetODEs}
        \dim\paraBrTerm{\mathfrak{g}_{\mathrm{pt}}}
        \leq
        \begin{cases}
            q^2+4q+3, & s=2, \\[1mm]
            q^2+sq+3, & s\geq3.
        \end{cases}
    \end{equation}
\end{proposition}
\begin{remark}[Determined System]
    \label{rem:determinedSystem}
    A system is called \textquote{determined} in here when the number of equations equals the number of dependent variables and the highest-order derivative Jacobian has full rank, so that the system is locally solvable for all highest-order derivatives. See also the maximal-rank assumption in \textit{\nameref*{subsec:deAndjetSpaceGeom}}.
\end{remark}

The bounds in Proposition~\ref{pro:finiteDimDetODESyms} are established for general determined ODE systems of the stated form and therefore apply, in particular, to the ODE systems considered here~\cite{ndogmo2016algebraic}. Finite dimensionality of the exact point-symmetry algebra, however, does not by itself imply that a chosen finite dictionary contains a sufficiently rich realization of all of its generators. We therefore retain the RSP for regular, determined ODE systems with $s>1$. Stabilization of the represented geometric rank provides operational evidence that further enrichment along the chosen nested representation does not reveal additional resolvable intrinsic directions, whereas failure to stabilize prevents a completeness claim within the present framework.

As stated previously, there is no generic additive solution sector that can be removed from the determining nullspace $\ker \widehat{\mathcal{D}}_{\Delta}$ before the discovery begins. Candidate generators must instead satisfy the solution-independence requirement individually upon discovery. If a candidate is identified as a solution-dependent vertical generator, it is excluded from $\mathfrak{g}_{\mathrm{rec}}$ as an approximation to $\mathfrak{g}_{\mathrm{int}}$, but its realized direction is retained in the novelty restriction to prevent subsequent rediscovery of the same generator.

The preceding conclusion changes fundamentally at $s=1$. Regular first-order ODEs, irrespective of linearity or nonlinearity, possess an unrestricted local point-symmetry structure with arbitrary functional freedom.
\begin{proposition}[Residual Functional Freedom for First-Order ODEs]
    \label{pro:firstOrdODEResidFuncFreedom}
    Let a regular determined first-order ODE system with $q\geq1$ dependent variables be locally written as
    \begin{equation}
        \label{eqn:firstOrdODESolved}
        \frac{d\mathbf{u}}{dx} = \mathbf{F}\paraBrTerm{x,\mathbf{u}}.
    \end{equation}
    Then, in a neighborhood of every regular point, there exist local point coordinates
    $\paraBrTerm{s,\mathbf{r}}
    $
    in which the system takes the \textit{rectified form}
    \begin{equation}
        \label{eqn:firstOrdODERectified}
        \frac{d\mathbf{r}}{ds} = \mathbf{0}.
    \end{equation}
    In these coordinates, every local Lie point-symmetry generator has the form
    \begin{equation}
        \label{eqn:firstOrdODEGenForm}
        v = A\paraBrTerm{s,\mathbf{r}}\partial_s + \sum_{\alpha=1}^{q}B^\alpha\paraBrTerm{\mathbf{r}}\partial_{r^\alpha},
    \end{equation}
    where $A$ is arbitrary and the functions $B^\alpha\paraBrTerm{\mathbf{r}}$ are arbitrary smooth functions of the local \textit{solution-leaf coordinates}.
    
    Moreover, the characteristic satisfies
    \begin{equation}
        \label{eqn:firstOrdODECharRect}
        \left. Q_v^\alpha\right\vert_{\Sigma^{(1)}} = B^\alpha\paraBrTerm{\mathbf{r}}.
    \end{equation}
    Hence, the arbitrary sector $A\paraBrTerm{s,\mathbf{r}}\partial_s$ is precisely tangent to the solution foliation and is removed by the characteristic-based nontriviality criterion $\mathcal{L}_\mathrm{tail}$ in \eqref{eqn:gatedNonTrivCost}, whereas the arbitrary functional family
    \begin{equation}
        \label{eqn:firstOrdODENontrivFuncFam}
        \sum_{\alpha=1}^{q} B^\alpha\paraBrTerm{\mathbf{r}}\partial_{r^\alpha}
    \end{equation}
    remains nontrivial. Consequently, quotienting by the characteristic-zero tangent sector does not render the unrestricted local point-symmetry structure finite-dimensional.
\end{proposition}

The proof of Proposition~\ref{pro:firstOrdODEResidFuncFreedom} is given in \ref{subapp:firstOrdODEResidFuncFreedom}. The proposition clarifies that first-order ODEs are not excluded because the characteristic-based nontriviality test becomes ineffective at $s=1$. Rather, the characteristic correctly removes the arbitrary solution-tangent sector, while an independent arbitrary functional freedom remains in transformations of the local solution-leaf coordinates. Consequently, quotienting by the sampled-trivial sector does not reduce the unrestricted exact local point-symmetry structure to a finite-dimensional algebra. We therefore exclude all first-order ODEs outright from the scope of the current framework.

This result also identifies a fundamental limitation of the RSP for first-order ODEs. As the represented function space is enriched, additional directions associated with the arbitrary functions $B^\alpha\paraBrTerm{\mathbf{r}}$ may become resolvable, so failure of the represented geometric rank in~\eqref{eqn:geometricRankOfDetSys} to stabilize appropriately prevents a completeness claim. Conversely, an apparent finite plateau cannot override the exact local structural result of Proposition~\ref{pro:firstOrdODEResidFuncFreedom}: it signifies only stabilization within the chosen finite dictionary, sampled realization, and numerical rank tolerance, rather than exhaustion of the unrestricted point-symmetry structure.

The situation is less definitive for nonlinear PDEs. Unlike regular, determined ODE systems with $s>1$, there is no corresponding general finite-dimensionality result for their unrestricted local Lie point-symmetry algebras. Depending on the structure of the governing equations, the determining system may close to a finite-dimensional algebra or may retain arbitrary functional degrees of freedom, as discussed from the formal-integrability viewpoint in \ref{app:geomViewPntAndFormalInteg}. In particular, nonlinear PDEs may admit infinite-dimensional Lie point-symmetry algebras parameterized by arbitrary functions; the \textit{Liouville equation} provides a classical example~\cite{gubbiotti2016partial,levi2015structure}. Consequently, finite dimensionality of the target algebra cannot be inferred solely from nonlinearity, differential order, or regularity.

For linear systems of sufficiently high differential order, the determining equations impose a stronger admissible restriction: admitted point-symmetry generators acquire a PFA form. We examine this structural simplification next.
\begin{proposition}[Projectable Fiber-Affine Form]
    \label{pro:projFibAffSym}
    Let $\Delta^{(s)} \coloneqq L^{(s)}[\mathbf{u}] - \mathbf{f}=\mathbf{0}$ be a regular linear DE of genuine order $s$, with nondegenerate highest-order part in the standard sense of~\cite[Section~2.6]{olver2000applications}, every admitted Lie point-symmetry generator has the form
    \begin{equation}
        \label{eqn:projFibAffSym}
        v = \sum_{i=1}^{p} \xi^i(\mathbf{x})\,\partial_{x^i} + 
            \sum_{\alpha=1}^{q} \sqBrTerm{A^\alpha_{\;\beta}(\mathbf{x})u^\beta +b^\alpha(\mathbf{x})}\partial_{u^\alpha}, 
    \end{equation}
    provided that either $\Delta^{(s)} = \mathbf{0}$ is a PDE with $s\geq2$ or is an ODE with $s\geq3$, where summation over index $\beta$ is assumed.
    
    Hence, the admitted point generators are projectable in the independent variables and fiber-affine in the dependent variables.
\end{proposition}

The proof of Proposition~\ref{pro:projFibAffSym} is provided in \ref{subapp:projectableFiberAffineForm}. For PDE systems, the preceding structural restrictions established in Section~\ref{sec:framework} and this section do not by themselves imply finite dimensionality of the residual intrinsic symmetry algebra. In the linear case, removal of the additive solution sector $\mathcal{A}_{\mathrm{sol}}$ eliminates the universal functional freedom induced by superposition, while the PFA form restricts the admissible dependence of the remaining generator coefficients and thereby permits a substantially simpler realization of the represented generators. Neither operation, however, guarantees that the corresponding determining system contains no additional arbitrary functional freedom.

Proposition~\ref{pro:projFibAffSym} does not extend the unrestricted PFA characterization to linear PDEs of first order. Nevertheless, for a scalar first-order linear PDE, the symmetry conditions can be characterized explicitly once the admissible generator is restricted to the PFA class.

\begin{proposition}[PFA Symmetries of a First-Order Linear PDE]
    \label{pro:firstOrderLinPFA}
    Let
    $\Delta^{(1)}[u]\coloneqq L[u]-f = a^i\paraBrTerm{\mathbf{x}}u_i + c\paraBrTerm{\mathbf{x}}u - f\paraBrTerm{\mathbf{x}} = 0$
    be a regular scalar first-order linear PDE, and let $v = \xi^i\paraBrTerm{\mathbf{x}}\partial_{x^i} + \sqBrTerm{A\paraBrTerm{\mathbf{x}}u + b\paraBrTerm{\mathbf{x}}}\partial_u$ be a PFA vector field. Then $v$ is an admitted Lie point-symmetry generator if and only if there exists a scalar smooth function $\mu\paraBrTerm{\mathbf{x}}$ such that
    \begin{equation}
        \label{eqn:firstOrdPFACond}
        \begin{aligned}
            \sqBrTerm{\boldsymbol{\xi}, \boldsymbol{a}} &= \mu\, \boldsymbol{a}, \\[1mm]
            \boldsymbol{a}\cdot\nabla A + \boldsymbol{\xi}\cdot\nabla c &= \mu\, c, \\[1mm]
            \boldsymbol{\xi}\cdot\nabla f -L[b] - Af &= \mu f.
        \end{aligned}
    \end{equation}
\end{proposition}

The same condition admits a compact operator-theoretic formulation.

\begin{proposition}[Operator Criterion for First-Order Linear PDEs]
    \label{pro:firstOrderLinPFAOpTh}
    Under the assumptions of Proposition~\ref{pro:firstOrderLinPFA}, let $L = \boldsymbol{a}\cdot\nabla+c$ and $R = A-\boldsymbol{\xi}\cdot\nabla$.
    Then the PFA field $v=\boldsymbol{\xi}\cdot\partial_{\mathbf{x}} + \paraBrTerm{Au+b}\partial_u$ is a Lie point symmetry of $L[u]=f$ if and only if there exists $\mu\in C^\infty$ such that
    \begin{equation}
        \label{eqn:firstOrdPFACondOpTh}
        [L,R] = \mu L, \quad L[b] + \paraBrTerm{R+\mu}f = 0.
    \end{equation}
\end{proposition}

The proofs of Propositions~\ref{pro:firstOrderLinPFA} and~\ref{pro:firstOrderLinPFAOpTh} are provided in \ref{subapp:pfaSymOfFirstOrderLinPDEs}. These propositions characterize the admitted symmetries within the PFA realization; they do not establish that every unrestricted Lie point symmetry of a first-order linear PDE must possess this form. Accordingly, first-order linear PDEs may be treated under a prescribed PFA model class, but completeness of the resulting recovery is understood relative to that restricted realization unless an independent structural argument establishes its exhaustiveness for the DE under consideration. No analogous PFA reduction is required for the ODE classes considered here: first-order ODEs are excluded from the present scope, whereas regular determined ODEs with $s=2$ are treated through an unrestricted realization whose target point-symmetry algebra is already known to be finite-dimensional by Proposition~\ref{pro:finiteDimDetODESyms}.

The preceding results establish the applicability limits adopted by the present framework. Regular, determined ODE systems of genuine order $s>1$ possess finite-dimensional local Lie point-symmetry algebras and are therefore admissible, subject to the RSP used to assess whether the chosen represented function space is sufficiently rich. First-order ODEs are excluded outright because arbitrary nontrivial functional freedom remains even after quotienting by the characteristic-zero solution-tangent sector. For linear PDEs with $s\geq2$, Proposition~\ref{pro:projFibAffSym} justifies the projectable fiber-affine realization after removal of the additive solution sector. Scalar first-order linear PDEs may also be treated within a prescribed PFA realization according to Propositions~\ref{pro:firstOrderLinPFA} and~\ref{pro:firstOrderLinPFAOpTh}, although completeness in that case remains relative to the restricted realization unless its exhaustiveness is established independently. Nonlinear PDEs may likewise be analyzed within the represented framework, but no unrestricted exact completeness claim is made without additional structural information excluding unresolved arbitrary functional freedom.

Accordingly, the RSP should be interpreted as a test of represented geometric completeness, not as a proof of exact formal closure. For admissible ODE systems with $s>1$, exact finite dimensionality is already guaranteed and the preflight therefore addresses representability of the finite target algebra. For PDE systems, by contrast, stabilization alone cannot establish the absence of additional functional degrees of freedom in the exact determining system. Failure to stabilize prevents a completeness claim in either case, but does not by itself imply that the exact symmetry algebra is infinite-dimensional.

For convenience, Fig.~\ref{fig:deAdmissibility_FlowChart} summarizes these distinctions through a decision flowchart indicating whether a given DE class lies within the present scope, which admissible generator realization is used, and whether the RSP is required before sequential recovery.

\subsection{Geometric Scope of the Recovered Symmetries}
\label{subsec:geomScopeOfRecSyms}
The proposed framework recovers infinitesimal Lie point-symmetry generators and therefore characterizes the local Lie-algebraic structure associated with transformations connected continuously to the identity. As discussed in \textit{\nameref*{subsec:fixedDictAndGenRealization}}, finite transformations generated by the recovered vector fields may subsequently be reconstructed through numerical integration of their prolonged flows. This reconstruction, however, should not be interpreted as recovery of the full global symmetry group.

In particular, the Lie algebra determines only the local structure of the identity-connected component. Disconnected components and discrete symmetries are not encoded by infinitesimal generators and therefore lie outside the direct scope of the present discovery procedure. Moreover, the Lie algebra does not determine the global topology of the associated Lie group: distinct connected Lie groups may possess isomorphic Lie algebras, being related through different global quotients of the corresponding simply connected covering group~\cite[Sections~1.3 and~5.8]{hall2015lie}. Consequently, even exact recovery of $\mathfrak{g}_{\mathrm{int}}$---and hence of algebraic constructions determined solely by it, such as its universal enveloping algebra---would not by itself identify the full global symmetry group.

Accordingly, $\mathfrak{g}_{\mathrm{rec}}$ should be interpreted as a numerical approximation to the local intrinsic symmetry algebra $\mathfrak{g}_\mathrm{int}$ within the represented domain. The extension of the corresponding generators and finite transformations across larger or overlapping regions is a separate geometric problem; the patchwise strategy motivated in \textit{\nameref*{subsec:fixedDictAndGenRealization}} is therefore left for future development.

\section{Conclusion}
\label{sec:conclusion}
In this work, we developed a data-driven framework for recovering intrinsic Lie point symmetries of DE systems from sampled jet-space data without prescribing the symmetry generators a priori. Starting from the conventional infinitesimal invariance condition, we first showed that imposing symmetry admissibility, nontriviality, solution independence, and novelty simultaneously leads to a difficult multi-objective optimization problem. We therefore reformulated the discovery task in an operator-theoretic, feasible-space-first form. At a fixed finite representation level, the sampled determining equations define a linear admissible space that is further restricted by removal of the additive solution sector when applicable and by global novelty with respect to previously recovered generators, while geometric nontriviality is enforced through the prolonged characteristic information. A fixed spline-based dictionary, nonlinear in its inputs but linear in its trainable coefficients, then makes these restrictions computationally explicit and permits the intrinsic generators to be recovered sequentially. The resulting procedure distinguishes clean exhaustion of the numerically resolvable represented directions from recovery failure, thereby separating a meaningful represented-completeness statement from unsuccessful optimization.

The framework was developed together with an explicit analysis of the structural assumptions under which such a completeness statement is meaningful. In particular, the RSP provides an operational test of represented geometric completeness under successive dictionary refinement, rather than a substitute for formal involutivity or a proof that the exact determining system contains no unresolved functional freedom. Within these limits, the two-dimensional spatial wave equation provides a particularly informative benchmark: its symmetry structure is well understood, its linearity introduces an additive solution-dependent sector that must be removed, and its remaining represented intrinsic sector permits the complete discovery pipeline to be examined within a stabilized finite representation. The numerical study therefore probes not only sequential generator recovery, but also the progressive restriction of the represented geometric space, numerical regularity of increasingly constrained generators, reconstruction of known analytic symmetry directions, and the locality of the learned representation. Accordingly, the recovered algebra $\mathfrak{g}_{\mathrm{rec}}$ is interpreted as a numerical approximation to the local intrinsic symmetry structure $\mathfrak{g}_{\mathrm{int}}$ within the represented and sampled domain, rather than as recovery of the full global symmetry group.

Several natural extensions follow from the distinction between recovering an intrinsic infinitesimal structure and exploiting that structure geometrically. Under the applicability conditions established in Section~\ref{sec:discApplAndLimit}, an immediate continuation of the present work is the construction of a $G$-coframe adapted to $\mathfrak{g}_{\mathrm{rec}}$, interpreted as a numerical approximation to the relevant finite-dimensional intrinsic symmetry structure. In particular, we plan to employ the cross-section and moving-frame constructions of Fels and Olver~\cite{fels1998moving,fels1999moving} to obtain symmetry-adapted invariant coordinates and thereby reduce or re-express a given DE system $\Delta^{(s)}=\mathbf{0}$. For practical applications, however, the governing DE cannot generally be considered independently of its prescribed initial and boundary data. The symmetry reduction must therefore be restricted to those recovered infinitesimal directions whose associated transformations preserve not only the DE, but also the relevant domain, boundary manifolds, and imposed data of the corresponding initial-boundary value problem~\cite{jassim2019symmetry}. Assuming that the numerically recovered generators identify, to the required accuracy, a Lie subalgebra with this invariance property, a differentiable computational realization could combine infinitesimal invariance constraints with neural-ODE-based integration of the corresponding local flows~\cite{chen2018neural} to construct invariant coordinates adapted to the associated connected subgroup $\widetilde{G}_0\subseteq G_0$. The resulting reduced DE system could then be solved, for example using PINNs~\cite{raissi2019physics}, and mapped back to the original coordinates so that the accuracy and practical utility of the symmetry-adapted reconstruction can be quantified.

The current spline-based dictionary representation $D_{(k)}$ is inherently local over $\Omega_{\mathbf{X}}$, and its extrapolation accuracy eventually deteriorates away from the training support, as demonstrated in Section~\ref{sec:experiments}. Independently rediscovering $\mathfrak{g}_{\mathrm{rec}}$ over a collection of local domains $\Omega_{\mathbf{X}}^{\sqBrTerm{a}}$, $a=1,2,\dotsc$, would, however, run counter to the intended use of the proposed framework: the intrinsic algebra should be discovered once and subsequently reused across different domains and problems. Taking inspiration from the atlas construction of Floryan and Graham~\cite{floryan2022data}, we therefore envision a patchwise transport procedure over overlapping domains satisfying $\Omega_{\mathbf{X}}^{\sqBrTerm{a}}\cap\Omega_{\mathbf{X}}^{\sqBrTerm{a+1}}\neq\varnothing$. For each recovered generator, a \textit{patch-transfer operator}
\begin{equation}
    T_{a\to a+1}: \boldsymbol{\Theta}_{j}^{\sqBrTerm{a}} \longmapsto \boldsymbol{\Theta}_{j}^{\sqBrTerm{a+1}}
\end{equation}
could be constructed by requiring consistency of the corresponding local realizations over the overlap,
\begin{equation}
    \left. v_j^{\sqBrTerm{a+1}} \right\vert_{\mathbf{S}^{\sqBrTerm{a,a+1}}} \approx
    \left. v_j^{\sqBrTerm{a}} \right\vert_{\mathbf{S}^{\sqBrTerm{a,a+1}}},
\end{equation}
where $\boldsymbol{\Theta}_{j}^{\sqBrTerm{a}}$ denotes the local coefficient representation of the already recovered generator $v_j$, and $\mathbf{S}^{\sqBrTerm{a,a+1}}$ is sampled from the overlap $\Omega_{\mathbf{X}}^{\sqBrTerm{a,a+1}}\coloneqq\Omega_{\mathbf{X}}^{\sqBrTerm{a}}\cap\Omega_{\mathbf{X}}^{\sqBrTerm{a+1}}$. In this way, patchwise continuation would transfer local representations of already discovered intrinsic directions rather than restart the sequential discovery procedure on every domain, thereby providing an atlas-like extension of $\mathfrak{g}_{\mathrm{rec}}$ beyond the original training support.

Another application that we find particularly interesting is \textit{flow-based generative modeling}, in which an invertible transformation transports a simple reference density $\mu\paraBrTerm{\mathbf{z}}$ toward a target density through a continuous flow~\cite{kobyzev2020normalizing}. If $\boldsymbol{\phi}_t$ denotes the associated flow map generated by a velocity field $\mathbf{w}_t$, with
\begin{equation}
    \frac{d}{dt}\boldsymbol{\phi}_t\paraBrTerm{\mathbf{z}} = \mathbf{w}_t\paraBrTerm{\boldsymbol{\phi}_t\paraBrTerm{\mathbf{z}},t},
\end{equation}
then the transported density $\rho_t$ satisfies the \textit{Liouville (or continuity) equation}
\begin{equation}
    \label{eqn:LiouvilleEqn}
    \partial_t\rho_t + \nabla_{\mathbf{z}}\!\cdot\paraBrTerm{\rho_t\mathbf{w}_t} = 0.
\end{equation}
For a prescribed velocity field, \eqref{eqn:LiouvilleEqn} is linear in $\rho_t$. In density-driven transport constructions such as the dual-ascent formulation of Tabak and Vanden-Eijnden~\cite{tabak2010density}, however, $\mathbf{w}_t$ is itself determined from the evolving density and its spatial derivatives, so that substitution into \eqref{eqn:LiouvilleEqn} yields a nonlinear \textit{density-evolution PDE}. This provides a natural setting in which to apply the proposed framework: intrinsic symmetries of the resulting density dynamics could first be recovered, followed by the construction of symmetry-adapted group and invariant coordinates. The generative transport could then be parameterized primarily in the invariant coordinates rather than requiring the learned flow to repeatedly represent motion along symmetry-equivalent directions. If successful, such a reduction could decrease the effective complexity of the transport problem and improve its numerical conditioning and efficiency. We emphasize the density-evolution formulation here because it lies directly within the DE-based Lie point-symmetry architecture developed in this work. A corresponding treatment formulated directly at the level of likelihood-based optimization or an associated variational functional would require additional machinery, including variational symmetries where appropriate~\cite[Chapter~4]{olver2000applications}, and is therefore left outside the scope of the present framework.

The preceding three research directions represent concrete extensions of the present framework. A more foundational continuation is to strengthen the criterion by which closure of the represented determining structure is assessed. As emphasized throughout the manuscript, the current RSP is intentionally an operational represented test: stabilization of the sampled geometric rank defined in \eqref{eqn:geometricRankOfDetSys} does not establish formal integrability, involutivity, or the absence of unresolved arbitrary functional freedom, as discussed in \ref{app:geomViewPntAndFormalInteg}. A natural long-term objective is therefore to develop a numerical/operator-theoretic analogue of Cartan involutivity for the determining system itself. Such a procedure would examine the behavior of the determining equations under successive differential prolongation---not to be confused with the prolongation operators introduced in Section~\ref{sec:preliminaries}; see Remark~\ref{rem:prolongationConfusion}---and numerically monitor compatibility conditions, symbol growth, rank evolution across prolonged levels, and the appearance or disappearance of new differential consequences. The objective would be to determine whether the determining system reaches a finite-type closure, in which case a finite represented symmetry space would be supported by the prolonged differential structure itself, or instead retains persistent symbolic freedom indicative of an unresolved functional symmetry sector. Such a diagnostic would provide a substantially stronger foundation for completeness claims than stabilization of a finite sampled dictionary rank alone.

Taken together, the present work represents a step from data-driven approximation of individual infinitesimal symmetries toward a systematic operator-theoretic framework for recovering and exploiting the geometric structure of DE systems. The emphasis throughout has been not only on identifying admissible generators, but also on distinguishing what can be inferred reliably within a finite represented and sampled setting from what requires deeper geometric information. The future directions outlined above aim to extend this perspective from symmetry discovery toward symmetry-adapted reduction, transport, probabilistic modeling, and ultimately stronger numerical characterizations of formal geometric closure. In this sense, the broader objective is to bring data-driven computation and the differential-geometric structure of DEs into a common computational framework.

\section*{Acknowledgements}
I.G. was financially supported by the Scientific and Technological Research Council of T\"{u}rkiye (T\"{U}B\.{I}TAK) and the Council of Higher Education (Y\"{O}K) in T\"{u}rkiye through the 2211-C and 100/2000 scholarship programs, respectively. The authors also gratefully acknowledge T\"{U}B\.{I}TAK for providing access to the National Grid System of T\"{u}rkiye (TRUBA), which was instrumental in meeting the computational requirements associated with the development and testing of the accompanying software.

I.G. gratefully acknowledges the sustained support and guidance of Prof. Peter J. Olver over the past three years. His contributions included providing crucial theoretical insights, helping resolve numerous challenges arising from the advanced mathematical background required for this research, and offering guidance on state-of-the-art developments in continuous group symmetries of PDEs and $G$-coframe construction.

I.G. also gratefully acknowledges Pelin \c{C}ilo\u{g}lu for her extensive guidance and discussions on optimization theory, functional analysis, and real analysis. Her support during the earlier stages of I.G.'s mathematical development contributed substantially to the analytical foundation that later proved valuable in carrying out this research.

\section*{Data availability}
No external datasets were used in this study. All datasets analyzed in the numerical experiments were generated computationally using the procedures described in the manuscript. The initial jet-space dataset consists of samples on $\Sigma^{(s)}$, while the additional datasets used in the symmetry-recovery and generator-approximation experiments were generated from the recovered or prescribed symmetry generators. The data-generation routines and all information required to regenerate the datasets used in the reported numerical experiments will be made available together with the accompanying source code.

\section*{Code availability}
The source code implementing the data-driven symmetry-discovery framework and the complete computational workflow used for the numerical experiments in this study has been prepared in a Code Ocean capsule. The capsule includes the Python package source, the YAML configuration files defining the manuscript benchmark, scheduler-independent task wrappers for data generation, sequential symmetry discovery, custom-generator approximation, integral-curve generation, visualization, and Lie-bracket closure testing, together with curated software tests and documentation for installation and reproduction.

The software is distributed as a self-contained \texttt{Python 3.12+} project and can be installed using the package metadata supplied with the release. The manuscript workflow can be executed either task-by-task or through the supplied full-workflow driver. The accompanying README documents the reference software environment, the expected computational products, and reproducibility considerations associated with stochastic equation-shell sampling and accelerator-dependent floating-point arithmetic. The numerical data used for the reported results are described separately in the Data availability statement. At the time of this preprint, the capsule is being kept private while journal peer review is ongoing; a public persistent link will be added to a future version of the preprint once the capsule is released.

\bibliographystyle{unsrtnat}
\bibliography{bibliography}

\clearpage
\onecolumn
\makesupplementarytitle

\appendix
\numberwithin{equation}{section}
\numberwithin{figure}{section}
\numberwithin{table}{section}
\renewcommand{\theequation}{\Alph{section}.\arabic{equation}}
\renewcommand{\thefigure}{\Alph{section}.\arabic{figure}}
\renewcommand{\thetable}{\Alph{section}.\arabic{table}}

\section{Geometric Viewpoint and Formal Integrability}
\label{app:geomViewPntAndFormalInteg}
The material in this section is not required to follow the computational construction of the proposed framework. Nevertheless, it provides the geometric foundation for several statements made throughout the paper, including Remark~\ref{rem:nonTrivFormalInteg}, and clarifies the future trajectory of the present research. We therefore briefly review the relevant elements of Cartan's exterior differential-system formulation.

\begin{definition}[Contact forms and contact ideal]
    \label{defn:contactForm}
    Let $\Omega^{(s),\,l} \coloneqq \Omega^1\paraBrTerm{J^{(s)}(\pi)}$ denote the space of smooth differential $l$-forms on $J^{(s)}(\pi)$~\cite[Section~3.11]{renteln2013manifolds}. For example, $\boldsymbol{\omega} = \sum_i P_i(z^{(s)})\,dz^{(s),\,i} \in \Omega^{(s),\,1}$ is a one-form, where $P_i:J^{(s)}(\pi)\to\mathbb{R}$ are its coefficients and $dz^{(s),\,i}$ are the coordinate one-forms spanning $T^\ast J^{(s)}(\pi)$.

    A one-form $\theta\in\Omega^{(s),\,1}$ is called a
    \textit{contact form} if
    \begin{equation}
        \label{eqn:cntFormProp}
        \paraBrTerm{j^s i_f}^{\ast}\theta \equiv 0
    \end{equation}
    for every local section $i_f:M\to E$. Thus, the $p$-dimensional integral manifolds of the contact system satisfying the usual independence condition are locally the prolongations of sections of $\pi$~\cite[Section~1.9]{ivey2003cartan}.

    In the jet coordinates introduced previously, the basic contact forms may be written as
    \begin{equation}
        \label{eqn:basicContactForms}
        \theta_I^\alpha = du_I^\alpha - \sum_{i=1}^{p} u_{I,i}^\alpha\, dx^i,
    \end{equation}
    where $\alpha=1,\dotsc,q$, $0\leq\vert I\vert\leq s-1$, and $I=0$ gives $\theta^\alpha = du^\alpha-\sum_i u_i^\alpha dx^i$.
    Writing
    \begin{equation}
        \boldsymbol{\theta}_k \coloneqq \curBarTerm{
            \theta_I^\alpha:
            \vert I\vert=k,\;
            \alpha=1,\dotsc,q
        },
    \end{equation}
    the order-$s$ \textit{contact ideal} is the
    differential ideal~\cite[Appendix~B.4]{ivey2003cartan} generated by $\boldsymbol{\theta}_0,\dotsc,\boldsymbol{\theta}_{s-1}$,
    \begin{equation}
        \label{eqn:contactIdeal}
        \mathcal{C}^{(s)} \coloneqq 
        \left\langle
            \boldsymbol{\theta}_0,\dotsc,
            \boldsymbol{\theta}_{s-1}
        \right\rangle_{\mathrm{diff}}.
    \end{equation}
\end{definition}

For a local solution section $i_f$ of the DE system, its prolonged graph $\Gamma_f^{(s)}=j^s i_f(M)$ therefore satisfies
\begin{equation}
    \label{eqn:contactPullback}
    \paraBrTerm{j^s i_f}^{\ast}\paraBrTerm{\theta_I^\alpha} \equiv 0,
    \quad
    \alpha=1,\dotsc,q,
    \quad
    0\leq\vert I\vert<s.
\end{equation}
Since pull-back $\paraBrTerm{j^s i_f}^{\ast}$ commutes with exterior differentiation $d$ and wedge products, every differential form in $\mathcal{C}^{(s)}$ consequently vanishes when pulled back to $\Gamma_f^{(s)}$.

\begin{definition}[Exterior differential system]
    \label{defn:eds}
    Let $\Delta^{(s)}=\mathbf{0}$ be a regular DE system with equation manifold $\Sigma^{(s)}\subset J^{(s)}(\pi)$, and let
    $\iota_{\Sigma}:\Sigma^{(s)}\hookrightarrow J^{(s)}(\pi)$ denote the inclusion map. The differential ideal
    \begin{equation}
        \label{eqn:edsDiffIdeal}
        \mathcal{I}_{\mathrm{diff}}^{(s)} \coloneqq \iota_{\Sigma}^{\ast}\mathcal{C}^{(s)}
    \end{equation}
    together with the independence form
    \begin{equation}
        \label{eqn:edsIndForm}
        \Omega \coloneqq dx^1\wedge\cdots\wedge dx^p
    \end{equation}
    defines the EDS associated with the DE system, subject to the independence condition $\Omega\neq0$ on its admissible $p$-dimensional integral elements.

    At a point $z^{(s)}\in\Sigma^{(s)}$, a $p$-dimensional plane $\mathcal{E}_{z^{(s)}}\subset T_{z^{(s)}}\Sigma^{(s)}$ is called an
    \textit{integral element} of the EDS~\cite[Definition~1.9.3]{ivey2003cartan} if
    \begin{equation}
        \label{eqn:integralElement}
        \left. \omega \right\vert_{\mathcal{E}_{z^{(s)}}} = 0
        \quad
        \forall\, 
        \omega\in\mathcal{I}_{\mathrm{diff}}^{(s)},
        \quad
        \left. \Omega \right\vert_{\mathcal{E}_{z^{(s)}}} \neq 0.
    \end{equation}
\end{definition}

One may regard the independence condition $\left.\Omega\right\vert_{\mathcal{E}_{z^{(s)}}}\neq0$ as restricting attention to $p$-dimensional integral elements that project nondegenerately onto the base space and are therefore locally graph-like. On the other hand,
$\mathcal{I}_\mathrm{diff}^{(s)}$ imposes the differential constraints that an admissible integral element must satisfy in order to be compatible with the jet structure of a prolonged local section. Whether such pointwise integral elements
\begin{enumerate*}[label=\textit{(\roman*)}]
    \item exist with the required dimension,
    \item vary smoothly, and
    \item can be \textquote{fitted together} to form a $p$-dimensional integral manifold
\end{enumerate*}
is a separate integrability question~\cite[Chapters~14--15]{olver1995equivalence}, which motivates the compatibility and involutivity considerations discussed below.

Consider the regular first-order ODE introduced previously in \textit{\nameref*{subsec:deAndjetSpaceGeom}},
\begin{equation}
    \label{eqn:contactEDSExample}
    \Delta^{(1)} \coloneqq u_x-xu=0.
\end{equation}
Its first-order jet bundle has local coordinates $z^{(1)}=(x,u,u_x)$, while its equation manifold is
\begin{equation}
    \Sigma^{(1)} = \curBarTerm{
        (x,u,u_x)\in J^{(1)}(\pi):
        u_x=xu
    }.
\end{equation}
Since $p=q=s=1$, the contact ideal $\mathcal{I}_\mathrm{diff}$ is generated by the single contact form
\begin{equation}
    \theta=du-u_x\,dx.
\end{equation}
Restricting it to $\Sigma^{(1)}$ gives
\begin{equation}
    \left.\theta\right\vert_{\Sigma^{(1)}}
    =
    du-xu\,dx,
\end{equation}
while the independence form is simply $\Omega=dx$.

At a point $z^{(1)}=(x,u,xu)\in\Sigma^{(1)}$, let $\mathcal{E}_{z^{(1)}}\subset T_{z^{(1)}}\Sigma^{(1)}$ be a one-dimensional integral element.
The conditions
\begin{equation}
    \left.\theta\right\vert_{\mathcal{E}_{z^{(1)}}}=0,
    \quad
    \left.dx\right\vert_{\mathcal{E}_{z^{(1)}}}\neq0
\end{equation}
imply that its tangent direction may be represented by
\begin{equation}
    w = \partial_x + xu\,\partial_u + u\paraBrTerm{1+x^2}\partial_{u_x}, 
\end{equation}
where $\partial_u$ coefficient is given by the contact condition, while tangency to $\Delta^{(1)} = 0$ defines $\partial_{u_x}$ coefficient. Observe that $w$ has the characteristic $\left. Q_w \right\vert_{\Sigma^{(1)}} \equiv 0$ implying that $w$ is trivial with respect to Definition~\ref{defn:nonTrivGen}. Consequently, fitting these integral elements together produces integral curves whose projection onto $E$ satisfies $u_x=xu$. Their solution sections are $u=f(x)=C e^{x^2/2}$, and their prolonged graphs
\begin{equation}
    \Gamma_f^{(1)} = \curBarTerm{
        \paraBrTerm{
            x,
            C e^{x^2/2},
            x C e^{x^2/2}
        }
    }
\end{equation}
are one-dimensional integral manifolds of the EDS.

The previous example presents a case in which the admissible integral elements can be smoothly fitted together into genuine solution graphs. For a general DE system, however, the existence of pointwise integral elements does not guarantee such integrability. The obstructions to doing so lead to the notions of compatibility, prolongation, torsion, and involutivity considered next.

The regularity assumptions imposed on $\Sigma^{(s)}$ earlier ensure that, locally, the DE system defines a smooth differential submanifold and satisfies
the standard nondegeneracy conditions required for the geometric treatment of its solutions~\cite[Section~2.6]{olver2000applications}. Regularity alone, however, does not guarantee that admissible jets extend consistently to higher orders.

Formal compatibility is examined by successively prolonging the DE system through total derivatives defined \eqref{eqn:totalDerivative} and projecting the resulting higher-order equation manifolds back to lower jet orders. If, for some $l\geq0$,
\begin{equation}
    \label{eqn:hiddenIntegCndDetection}
    \pi_{s+l}^{s+l+1}
    \paraBrTerm{\Sigma^{(s+l+1)}}
    \subsetneq
    \Sigma^{(s+l)},
\end{equation}
the prolongation has revealed additional \textit{integrability (compatibility) conditions} that must be incorporated into the system. Once no new independent lower-order conditions arise under further prolongation, the system is \textit{formally integrable}~\cite[Section~2.3]{Seiler2010}.

In Cartan's EDS formulation, the same compatibility problem is encoded locally through the associated Pfaffian system, its tableau, and torsion. We briefly introduce these objects next. To pass from the EDS formulation above to Cartan's linear Pfaffian description, we retain the same restricted contact forms but distinguish between their degree-one span and the differential ideal they generate. Namely, let
\begin{equation}
    \mathcal{I}^{(s)} \coloneqq \operatorname{span}_{C^\infty(\Sigma^{(s)})} 
        \curBarTerm{
            \left.\theta_I^\alpha\right\vert_{\Sigma^{(s)}}:
            \alpha=1,\dotsc,q,\;
            0\leq\vert I\vert<s
        },
\end{equation}
so that
\begin{equation}
    \mathcal{I}_{\mathrm{diff}}^{(s)} = \left\langle \mathcal{I}^{(s)} \right\rangle_{\mathrm{diff}}.
\end{equation}
Thus, $\mathcal{I}^{(s)}$ and $\mathcal{I}_{\mathrm{diff}}^{(s)}$ are generated by the same contact one-forms; the former denotes their Pfaffian span in
$\Omega^1(\Sigma^{(s)})$, whereas the latter denotes the full differential ideal they generate~\cite[Section~5.1]{ivey2003cartan}.

Since the determining equations form an overdetermined linear PDE system for the infinitesimal coefficients, as described in \textit{\nameref*{subsec:infSymCndAndDetEqns}}, they may themselves be represented, after passage to the appropriate jet space and first-order formulation, as a linear Pfaffian system~\cite[Example~5.1.4]{ivey2003cartan}.

\begin{remark}
    \label{rem:determiningPfaffianCoordinates}
    Throughout this appendix, we apply Cartan's equivalence machinery to the original DE system. When the same construction is applied to its determining system, the collection $\curBarTerm{dx^1,\dotsc,dx^p; du^1,\dotsc,du^q}$ replaces $\curBarTerm{dx^1,\dotsc,dx^p}$ as the independence one-forms, since the determining equations are posed over the total space $E$ rather than the base space $\mathbf{X}$. Accordingly, $\paraBrTerm{\boldsymbol{\xi},\boldsymbol{\varphi}}$ become the dependent coordinates of the determining system.
\end{remark}

\begin{definition}[Linear Pfaffian system]
    \label{defn:linearPfaffianSystem}
    Let $\Delta^{(s)}=\mathbf{0}$ be a regular DE system with equation manifold $\Sigma^{(s)}$, and let $\mathcal{I}^{(s)}\subset\Omega^1\paraBrTerm{\Sigma^{(s)}}$ denote the Pfaffian system generated by the restricted contact one-forms
    introduced in Definition~\ref{defn:contactForm}. Its generated differential ideal is $\mathcal{I}_\mathrm{diff}^{(s)}$ in
    \eqref{eqn:edsDiffIdeal}.

    Let further $\omega^i\coloneqq dx^i$, $i=1,\dotsc,p$, satisfy the independence condition $\Omega\coloneqq\omega^1\wedge\cdots\wedge\omega^p\neq 0$, and define
    \begin{equation}
        \mathcal{J}^{(s)} \coloneqq \mathcal{I}^{(s)} \cup \operatorname{span}\curBarTerm{\omega^1,\dotsc,\omega^p}.
    \end{equation}
    The pair $\paraBrTerm{\mathcal{I}^{(s)},\mathcal{J}^{(s)}}$ is called a \textit{linear Pfaffian system} if
    \begin{equation}
        d\theta^a \equiv 0 \paraBrTerm{\operatorname{mod}\mathcal{J}^{(s)}}
    \end{equation}
    for every local generator $\theta^a$ of $\mathcal{I}^{(s)}$~\cite[Definition~5.1.1]{ivey2003cartan}.

    Let $\curBarTerm{\tau^\beta}$ be one-forms complementary to $\mathcal{J}^{(s)}$ in $T^\ast\Sigma^{(s)}$. Then the corresponding
    structure equations take the local form
    \begin{equation}
        \label{eqn:structureEqns}
        d\theta^a \equiv A_{\beta i}^{a}\, \tau^\beta\wedge\omega^i + T_{ij}^{a}\, \omega^i\wedge\omega^j
        \paraBrTerm{\operatorname{mod}\mathcal{I}^{(s)}},
    \end{equation}
    where $A_{\beta i}^{a}$ and $T_{ij}^{a}$ are the \textit{tableau} and \textit{torsion} coefficients, respectively~\cite[Section~5.1]{ivey2003cartan}.
\end{definition}

At a point $z^{(s)}\in\Sigma^{(s)}$, let $V^\ast\cong\mathcal{J}^{(s)}/\mathcal{I}^{(s)}$ be spanned by the independence forms $\omega^i$, and let $W^\ast\cong\mathcal{I}^{(s)}$ be spanned by the Pfaffian generators $\theta^a$. The coefficients $A_{\beta i}^{a}$ therefore define the \textit{tableau}
\begin{equation}
    \label{eqn:tableau}
    \mathbf{A} \coloneqq \operatorname{span} \curBarTerm{\mathbf{A}_\beta} \subseteq W\otimes V^\ast,
\end{equation}
which encodes the admissible highest-order differential freedom of the system~\cite[Section~4.1]{ivey2003cartan}. In the language of PDE
theory, this role is closely related to that of the \textit{principal symbol}, which captures the highest-order differential structure of a differential
operator or system~\cite[Section~9]{taylor2011pde1}.

After restricting the complementary forms to an integral element $\mathcal{E}_{z^{(s)}}$, the independence condition permits the local graph representation
\begin{equation}
    \label{eqn:linAlgGrphRepOfIntegElt}
    \left. \paraBrTerm{\tau^\beta-P_i^\beta\omega^i}\right\vert_{\mathcal{E}_{z^{(s)}}} = 0,
\end{equation}
where $P_i^\beta$ are its \textit{slope coefficients}. Substituting \eqref{eqn:linAlgGrphRepOfIntegElt} into \eqref{eqn:structureEqns} and imposing
$\left.d\theta^a\right\vert_{\mathcal{E}_{z^{(s)}}}=0$ yields algebraic conditions on $P_i^\beta$. Their homogeneous part defines the \textit{first prolongation} of the tableau,
\begin{equation}
    \label{eqn:firstTableauProlongation}
    \mathbf{A}^{(1)} \coloneqq \paraBrTerm{\mathbf{A}\otimes V^\ast} \cap \paraBrTerm{W\otimes S^2V^\ast}
    \subseteq \mathbf{A}\otimes V^\ast,
\end{equation}
which represents the admissible next-order differential freedom associated with $\mathbf{A}$~\cite[Section~4.1]{ivey2003cartan}.

\begin{remark}
    \label{rem:prolongationConfusion}
    Although $\mathbf{A}\longmapsto\mathbf{A}^{(1)}$, $\mathrm{pr}^{(s)}:\mathfrak{X}(E)\to \mathfrak{X}\paraBrTerm{J^{(s)}(\pi)}$ in \eqref{eqn:prolongationOfGens}, and $j^s i_f:M\to J^{(s)}(\pi)$ are all referred to as prolongations, they act on different geometric
    objects. The first prolongs a tableau in tensor space, the second prolongs a vector field, and the third prolongs a local section.
    Their common feature is the encoding of higher-order differential information, but their outputs serve distinct geometric purposes.
\end{remark}

\begin{definition}[Involutive tableau]
    \label{defn:involutivity}
    Let $\paraBrTerm{\mathcal{I}^{(s)},\mathcal{J}^{(s)}}$ be a linear Pfaffian system with tableau $\mathbf{A}$. The tableau $\mathbf{A}$ is called \textit{involutive} if it satisfies \textit{Cartan's test}, evaluated through its first prolongation $\mathbf{A}^{(1)}$, as specified in Definition~11.7 of~\cite{olver1995equivalence} or Definition~4.5.4 of~\cite{ivey2003cartan}.
\end{definition}

At the tableau level, involutivity means that the tableau prolongation has exactly the dimension predicted by Cartan's test, with subsequent higher-order freedom determined by the Cartan characters. If $\mathbf{A}$ is not involutive, the system is prolonged and $\mathbf{A}^{(1)}$ becomes the next tableau to be tested, now through $\mathbf{A}^{(2)}$; the same procedure is repeated at successive orders as needed~\cite[Sections~5.5--5.6]{ivey2003cartan}. Once an involutive tableau is reached, its Cartan characters quantify the remaining differential freedom. In particular, a non-Frobenius determining system need not possess arbitrary functional freedom, since its tableau may eventually become trivial; i.e. $\mathbf{A}^{(l)} = 0$ for some $l \geq 1$, after finitely many prolongations. Once an involutive nonzero tableau is reached, however, at least one Cartan character is positive, and the corresponding Cartan--K\"{a}hler data include arbitrary functions, thereby yielding an infinite-dimensional Lie point-symmetry algebra. See Definition~1.3.5 and Example~4.1.4 in~\cite{ivey2003cartan} for the definition of a \textit{Frobenius system} and the corresponding tableau, respectively.

The torsion coefficients in \eqref{eqn:structureEqns} are not all intrinsic obstructions to integrability, since the complementary forms
$\curBarTerm{\tau^\beta}$ are not uniquely determined. In particular, they may be replaced locally by
\begin{equation}
    \label{eqn:torsionAbsorption}
    \widetilde{\tau}^{\beta} = \tau^\beta + P_i^\beta\omega^i
    \quad
    \paraBrTerm{\operatorname{mod}\mathcal{I}^{(s)}}.
\end{equation}
Such a change leaves the tableau unchanged, but modifies the apparent torsion by
\begin{equation}
    \label{eqn:affTorsionEqn}
    \widetilde{\mathbf{T}} = \mathbf{T} + \delta(\mathbf{P}),
\end{equation}
where
\begin{equation}
    \delta: \mathbf{A}\otimes V^\ast \longrightarrow W\otimes\Lambda^2 V^\ast
\end{equation}
is the alternation map. Consequently, only the equivalence class
\begin{equation}
    \label{eqn:essentialTorsion}
    \sqBrTerm{\mathbf{T}} \in \frac{W\otimes\Lambda^2V^\ast}{\operatorname{im}\delta} = \operatorname{coker}\delta
\end{equation}
is independent of the choice of complementary forms and is called the \textit{essential torsion}~\cite[Section~5.5]{ivey2003cartan}.

Indeed, after representing an integral element by \eqref{eqn:linAlgGrphRepOfIntegElt}, the condition $\left.d\theta^a\right\vert_{\mathcal{E}_{z^{(s)}}}=0$ takes the affine form
\begin{equation}
    \label{eqn:homoTorsionEqn}
    \delta(\mathbf{P})+\mathbf{T}=\mathbf{0}.
\end{equation}
With the sign convention adopted above, this is precisely the condition $\widetilde{\mathbf{T}}=\mathbf{0}$ under the complementary-coframe transformation \eqref{eqn:torsionAbsorption}. Hence, an admissible integral element can exist at $z^{(s)}$ only if $\sqBrTerm{\mathbf{T}}=\mathbf{0}$. Whenever this condition holds and $\mathbf{P}_0$ is a particular solution to \eqref{eqn:homoTorsionEqn}, the apparent torsion can be absorbed by the corresponding change of complementary coframe, while the remaining homogeneous freedom is precisely $\ker\delta=\mathbf{A}^{(1)}$. Equivalently, the admissible integral elements form an affine space modeled on $\mathbf{A}^{(1)}$; i.e. $\mathbf{P}_0 + \ker\delta$.

If $\sqBrTerm{\mathbf{T}}=\mathbf{0}$ does not hold identically on $\Sigma^{(s)}$, its vanishing imposes additional compatibility conditions and the system must be restricted to the corresponding locus
\begin{equation}
    \label{eqn:torsionRestrSet}
    \widetilde{\Sigma}^{(s)}
    \coloneqq
    \curBarTerm{z^{(s)}\in\Sigma^{(s)}: \sqBrTerm{\mathbf{T}}=\mathbf{0}}.
\end{equation}
The linear Pfaffian system and its tableau are then recomputed on $\widetilde{\Sigma}^{(s)}$ before Cartan's involutivity test, introduced in Definition~\ref{defn:involutivity}, is applied. Thus, the restriction $\sqBrTerm{\mathbf{T}}=\mathbf{0}$ provides the EDS-side manifestation of the same compatibility phenomenon expressed by the strict projection \eqref{eqn:hiddenIntegCndDetection}: it removes jets satisfying the original finite-order equations that cannot be extended consistently to admissible higher-order integral elements. Further prolongation may, however, reveal additional integrability conditions not detected by the torsion at the current order.

\begin{figure}[h!]
    \centering
    \includegraphics[width=1.0\textwidth]{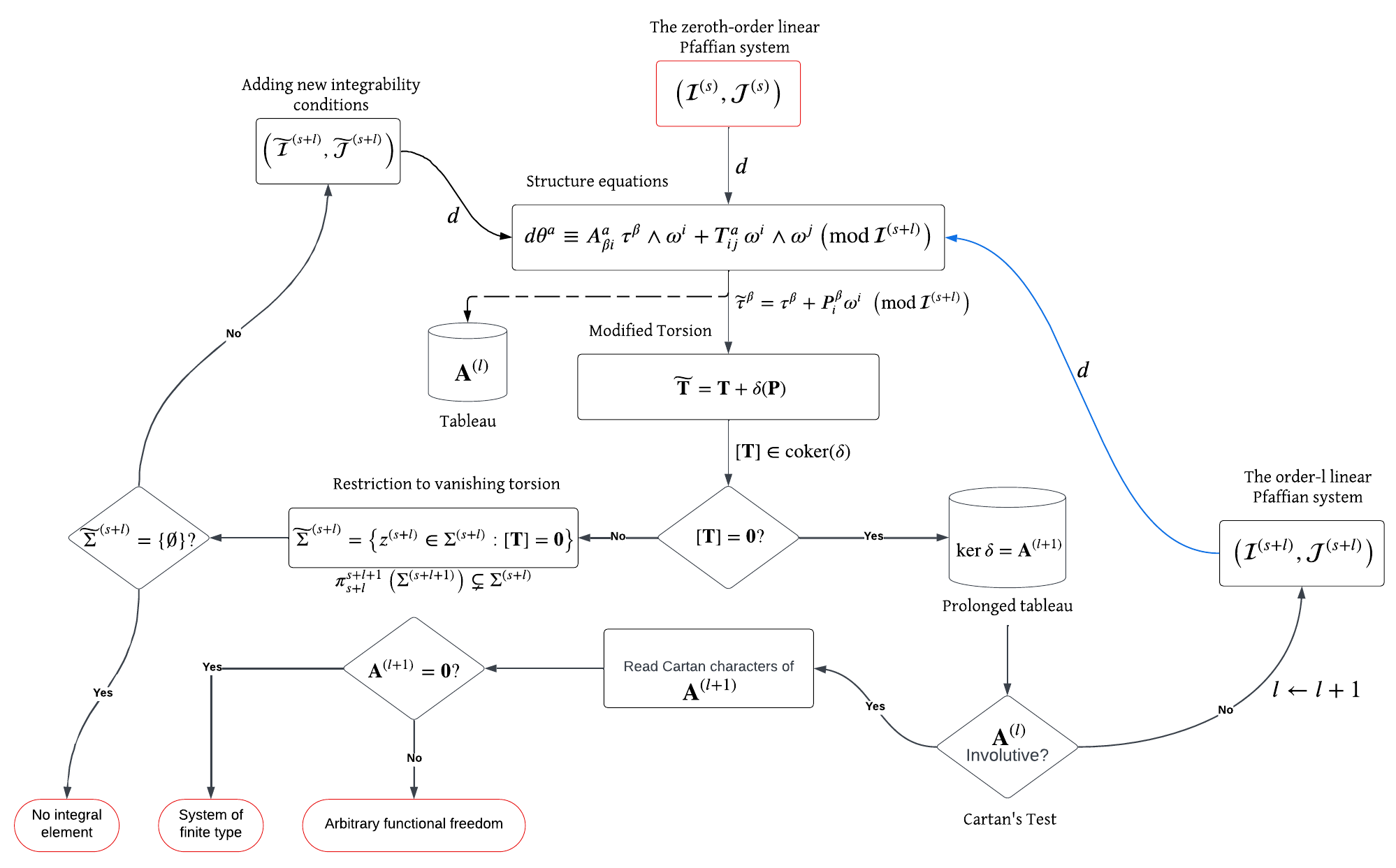}
    \caption{Flowchart of Cartan's iterative procedure for a linear Pfaffian system, including torsion compatibility, prolongation, involutivity, and the resulting differential freedom. The blue arrow indicates relabeling of the Pfaffian one-form systems for the next iteration. The prolongation index $l$ starts at $0$.}
    \label{fig:CartanPfaffianFlowchart}
\end{figure}

The interplay between tableau involutivity and torsion gives rise to an iterative completion process. If the essential torsion does not vanish identically, restricting to $\sqBrTerm{\mathbf{T}}=\mathbf{0}$ in \eqref{eqn:torsionRestrSet} introduces additional compatibility conditions, requiring the Pfaffian system and its tableau to be recomputed. Conversely, if the resulting tableau is not involutive, prolongation produces a higher-order Pfaffian system for which the torsion coefficients must again be evaluated. The complete Cartan procedure therefore alternates between these two operations until either inconsistency is detected or an involutive system is reached. Figure~\ref{fig:CartanPfaffianFlowchart} summarizes this process. The blue arrow indicates that, after prolongation, the one-forms defining the newly obtained Pfaffian pair are relabeled as the current $\paraBrTerm{\mathcal{I}^{(s+l)},\mathcal{J}^{(s+l)}}$ for the next iteration; the prolongation index $l$ begins at $0$. This flowchart also clarifies which components of Cartan's procedure would ultimately be required to replace the RSP test described in Section~\ref{sec:implAndCompConsid}.

The preceding Pfaffian formulation also admits a complementary group-theoretic interpretation through Cartan's method of equivalence. Gardner formulates an equivalence problem by choosing an adapted coframe and prescribing a matrix Lie group $G\subset GL(n,\mathbb{R})$ acting linearly on that coframe; admissible equivalences are therefore characterized, at first order, by Jacobians taking values in $G$~\cite[Lecture~1]{gardner1989method}. After lifting the problem to the corresponding $G$-space (principal bundle), the differential freedom along the group fibers is encoded by the right-invariant \textit{Maurer--Cartan} forms of $G$~\cite[Chapter~2]{olver1995equivalence}. Indeed, Gardner's lifted structure equations contain the Maurer--Cartan matrix $dS\,S^{-1}$, whose independent entries encode the Lie algebra of the structure group~\cite[Lectures~2--3]{gardner1989method}.

\section{Spline-Dictionary Redundancy and Sampling-Induced Rank Deficiency}
\label{app:splineDictRedAndSamplingIndRankDef}
The nested construction in \eqref{eqn:nestedSplineDict} provides increasing representational richness, but it also introduces an exact coefficient-space redundancy across interaction orders. In particular, the B-spline basis associated with each coordinate satisfies the \textit{partition-of-unity} property on the core domain, 
\begin{equation} 
    \label{eqn:bSplinePartitionUnity} 
    \sum_{\nu_j=1}^{K_o} B_{\nu_j}^{(r)}\paraBrTerm{\chi^j} =1 
\end{equation} 
Hence, for any $i\neq j$, 
\begin{equation} 
    \label{eqn:crossOrderSplineRedundancy} 
    B_{\nu_i}^{(r)}\paraBrTerm{\chi^i} = 
            B_{\nu_i}^{(r)}\paraBrTerm{\chi^i}
            \sum_{\nu_j=1}^{K_o} B_{\nu_j}^{(r)}\paraBrTerm{\chi^j} \
        = \sum_{\nu_j=1}^{K_o} B_{\nu_i}^{(r)}\paraBrTerm{\chi^i} B_{\nu_j}^{(r)}\paraBrTerm{\chi^j},
\end{equation} 
i.e. coefficients across all $\nu_j$'s, for a fixed $\nu_i$, are shared. Consequently, a first-order feature can be represented by a particular linear combination of second-order features, and analogous dependencies occur between successive higher interaction orders. The coefficient representation generated by $D_{(k)}$ is therefore not necessarily identifiable, so the corresponding realization map need not be injective. This redundancy does not prevent genuinely higher-order dependence from being represented: the reduction in \eqref{eqn:crossOrderSplineRedundancy} occurs only for the specific coefficient pattern that reproduces the partition-of-unity sum.

A distinct source of degeneracy arises from the finite sampling of the spline dictionary. Let
\begin{equation}
    \paraBrTerm{\Phi_i}_{l\nu} \coloneqq B_{\nu}^{(r)}\paraBrTerm{\chi_l^i},
    \; l=1,\dotsc,N,
    \; \nu=1,\dotsc,K_o.
\end{equation}
denote the \textit{evaluation matrix} of the univariate B-spline basis associated with the coordinate $\chi^i$ over sampled inputs $\curBarTerm{\chi_l^i}_{l=1}^{N}$, thereby each column corresponds to the sampled response of a particular B-spline basis function and $\Phi_i \in \mathbb{R}^{N\times K_o}$. Although the B-spline functions are linearly independent on their underlying spline space, their restrictions to a finite sample set $\mathbf{S}$ with $\vert \mathbf{S} \vert = N$ need not remain numerically independent. If the samples insufficiently cover the supports of one or more basis functions, the corresponding columns of $\Phi_i$ may vanish, become weakly activated, or become nearly proportional to other sampled columns. Consequently, the sampled evaluation matrix may exhibit exact or approximate rank deficiency, even when no additional functional dependence is present in the continuous dictionary. In the approximately rank-deficient case, this manifests numerically as
\begin{equation}
    \label{eqn:bSplineEvalRankDef}
    \sigma_{\mathrm{min}}\paraBrTerm{\Phi_i}
    \ll
    \sigma_{\mathrm{max}}\paraBrTerm{\Phi_i},
\end{equation}
whereas exact rank deficiency yields $\sigma_{\mathrm{min}}\paraBrTerm{\Phi_i}=0$.

This effect is amplified for tensor-product features in $D_{(k)}$, since weak activation of any univariate factor suppresses the corresponding multivariate feature. Thus, a mismatch between the spline grid and the effective sampled support, or more generally \textit{insufficient sample coverage of the spline supports}, can introduce additional near-null directions in coefficient space. These sampling-induced deficiencies are distinguished from the exact structural redundancy in \eqref{eqn:crossOrderSplineRedundancy}: the former depend on the particular sample set, whereas the latter are identities of the dictionary itself.

\section{Why Solvability Is Not Imposed as a Discovery Constraint}
\label{app:whyNotImposingSolvability}
A natural question arises in view of the limitations of the RSP and finite dictionary representation discussed in Section~\ref{sec:implAndCompConsid}: \textquote{Why not impose \textit{solvability} as an additional constraint during sequential recovery, so that the accepted generators are at least guaranteed to close into a solvable Lie subalgebra even when recovery of the entire intrinsic algebra cannot be certified?} The difficulty is that such a constraint changes the admissible recovery path itself and can therefore force premature termination inside a proper represented subalgebra. To make this point precise, we first recall the finite-dimensional characterization of solvability used in Definition~2.63 of~\cite{olver2000applications}.

\begin{definition}[Solvable Lie Group and Lie Algebra]
    \label{def:solvabilityOfLieGrp}
    Let $G$ be an $r$-dimensional Lie group of point transformations with Lie algebra $\mathfrak{g}_{\mathrm{pt}}$. The group $G$ is called \textit{solvable} if there exists a chain of Lie subgroups
    \begin{equation}
        \label{eqn:chainOfSolvabilityGrps}
        \curBarTerm{e} = G_0 \subset G_1 \subset \cdots \subset G_r = G,
    \end{equation}
    such that, for each $i=1,\dotsc,r$, $G_i$ is $i$-dimensional and $G_{i-1}$ is a normal subgroup of $G_i$. Equivalently, there exists a chain of Lie subalgebras
    \begin{equation}
        \label{eqn:chainOfSolvabilityAlgs}
        \curBarTerm{\mathbf{0}} = \mathfrak{s}_0 \subset \mathfrak{s}_1 \subset \cdots \subset \mathfrak{s}_r = \mathfrak{g}_{\mathrm{pt}},
    \end{equation}
    with $\dim\paraBrTerm{\mathfrak{s}_i}=i$, such that $\mathfrak{s}_{i-1}$ is an ideal of $\mathfrak{s}_i$ for every $i=1,\dotsc,r$, i.e.,
    \begin{equation}
        \label{eqn:normalSubAlgCnd}
        \sqBrTerm{\mathfrak{s}_{i-1},\mathfrak{s}_i} \subseteq \mathfrak{s}_{i-1}.
    \end{equation}
\end{definition}

Let
\begin{equation}
    \mathfrak{g}_{\mathrm{rec},i-1} \coloneqq \operatorname{span}\curBarTerm{v_1,\dotsc,v_{i-1}}
\end{equation}
denote the algebraic sector recovered after $i-1$ accepted generators in the sequential recovery procedure detailed in \textit{\nameref*{subsec:seqRecoveryAndStopping}}. Let $v_{i}$ be a candidate for the next intrinsic direction under the assumptions that $\dim\paraBrTerm{\widehat{N}_{\mathrm{nov},i}^{(k)}} > 0$ and the initialization conditions in \eqref{eqn:initCandidateAcceptance} are satisfied. If, in addition to the existing admissibility conditions, one enforces the normal-subalgebra condition required by the chain in \eqref{eqn:chainOfSolvabilityAlgs},
\begin{equation}
    \label{eqn:imposingNormSubalgChain}
    \sqBrTerm{\mathfrak{g}_{\mathrm{rec},i-1}, v_i} \subseteq \mathfrak{g}_{\mathrm{rec},i-1},
\end{equation}
then a genuinely novel represented direction need not remain feasible. Consequently, the sequential procedure may terminate prematurely, even though additional intrinsic directions remain representable by the underlying finite dictionary. The resulting algebra can therefore be a proper subalgebra of the represented intrinsic sector solely because the imposed solvability constraint excludes otherwise admissible generators.

A simple example is the \textit{special linear Lie algebra} $\mathfrak{sl}(2)$ with structure relations~\cite[Section~4.6]{hall2015lie}
\begin{equation}
    \label{eqn:strEqnSL2}
    \sqBrTerm{H,X}=2X,\; \sqBrTerm{H,Y}=-2Y,\; \sqBrTerm{X,Y}=H.
\end{equation}
Suppose that after two recovery steps,
\begin{equation}
    \mathfrak{g}_{\mathrm{rec},2} = \operatorname{span}\curBarTerm{H,X}.
\end{equation}
Although this two-dimensional space is a Lie subalgebra, it is not an ideal of $\mathfrak{sl}(2)$, since
\begin{equation}
    \label{eqn:sl2FailureMode}
    \sqBrTerm{Y,H} = 2Y \notin \mathfrak{g}_{\mathrm{rec},2}.
\end{equation}
More generally, every direction
\begin{equation}
    v_3=aH+bX+cY, \quad c\neq0,
\end{equation}
that is linearly independent of $\mathfrak{g}_{\mathrm{rec},2}$ satisfies
\begin{equation}
    \sqBrTerm{v_3,H} = -2bX+2cY \notin \mathfrak{g}_{\mathrm{rec},2}.
\end{equation}
Hence no genuinely novel third generator can satisfy \eqref{eqn:imposingNormSubalgChain}, despite the existence of the remaining represented direction required to recover the full three-dimensional algebra.

The difficulty is compounded by the fact that the sequential recovery order is not prescribed a priori. Different choices of initialization, dictionary resolution and interaction order, spline degree, optimization hyperparameters, and numerical tolerances can lead the algorithm to recover different admissible bases of the same represented intrinsic sector. Consequently, imposing the normal-subalgebra condition in \eqref{eqn:imposingNormSubalgChain} would make the set of admissible future directions depend not only on the underlying Lie algebra, but also on the particular basis and ordering encountered numerically during training. A legitimate generator could therefore be excluded solely because previously recovered generators happened to span a subalgebra that cannot be extended through the required normal chain in that order. Such a path-dependent restriction is incompatible with the objective of exhausting all numerically resolvable intrinsic directions represented by $D_{(k)}$.

Another structural issue is that solvability certifies neither finite dimensionality nor completeness of $\mathfrak{g}_{\mathrm{int}}$. Indeed, any infinite-dimensional Abelian Lie algebra is solvable, since its first derived algebra is already trivial, showing that solvability alone does not imply finite dimensionality. Conversely, a finite-dimensional intrinsic algebra need not be solvable. Consider $\mathfrak{sl}(2)$ again. Its \textit{derived series} (see Definition~3.14 in~\cite{hall2015lie}) satisfies
\begin{equation}
    \label{eqn:derivedSeriesSL2}
    \mathfrak{sl}(2) \eqqcolon \mathfrak{g}_{(0)} = \mathfrak{g}_{(1)} = \cdots =
    \mathfrak{g}_{(j)} = \mathfrak{sl}(2),
\end{equation}
for every finite $j$, and therefore never reaches the trivial algebra $\curBarTerm{\mathbf{0}}$. Hence, the finite-dimensional algebra $\mathfrak{sl}(2)$ is not solvable. Together with the premature-termination mechanism discussed above, these observations show that solvability is neither an appropriate completeness constraint for recovering all represented intrinsic directions nor a certificate of finite dimensionality. Passing the RSP test establishes only stabilization of the represented geometric rank under the prescribed dictionary refinement; it provides no reason to assume that the corresponding intrinsic algebra is solvable. Accordingly, solvability is not imposed as an admissibility or stopping condition in the present framework.

\section{Proofs}
\label{app:proofs}

\subsection{Generic Shell Sampling}
\label{subapp:genericShellSampling}
This section provides the proof for Proposition~\ref{pro:genericShellSampling}. For the first statement, let $\pi^s:J^{(s)}\paraBrTerm{\pi}\to M$ denote the canonical jet projection in \eqref{eqn:jetCanonicalProjections}. The prolonged graph $\Gamma_f^{(s)}$ is the image of the jet prolongation $j^s i_f:M\to J^{(s)}\paraBrTerm{\pi}$ in \eqref{eqn:localSecProlongation}, and
\begin{equation}
    \pi^s\circ j^s i_f=\operatorname{id}_M.
\end{equation}
Hence $j^s i_f$ is injective and its differential has full rank $\dim\paraBrTerm{M}=p$, so $\Gamma_f^{(s)}$ is an embedded $p$-dimensional submanifold. Since $\Gamma_f^{(s)}\subset\Sigma^{(s)}$ and, by assumption, $\dim\paraBrTerm{\Sigma^{(s)}}>p$, it follows that
\begin{equation}
\operatorname{codim}_{\Sigma^{(s)}}\paraBrTerm{\Gamma_f^{(s)}} = \dim\paraBrTerm{\Sigma^{(s)}}-p > 0.
\end{equation}
Therefore $\Gamma_f^{(s)}$ is a positive-codimension embedded submanifold of $\Sigma^{(s)}$ and consequently has zero measure with respect to any smooth volume measure on $\Sigma^{(s)}$.

For the second statement, let $\nu_{\Sigma^{(s)}}$ denote a \textit{smooth volume measure} on $\Sigma^{(s)}$. The first statement gives
\begin{equation}
    \nu_{\Sigma^{(s)}}\paraBrTerm{\Gamma_f^{(s)}}=0.
\end{equation}
By assumption, the sampling distribution satisfies $\mu_{\Sigma^{(s)}}\ll\nu_{\Sigma^{(s)}}$. By the definition of \textit{absolute continuity} of measures~\cite[p.~88]{folland1999real}, every $\nu_{\Sigma^{(s)}}$-null set is also $\mu_{\Sigma^{(s)}}$-null, and hence
\begin{equation}
    \mu_{\Sigma^{(s)}}\paraBrTerm{\Gamma_f^{(s)}}=0.
\end{equation}
Thus, for every sampled jet $z_l^{(s)}$ drawn from $\mu_{\Sigma^{(s)}}$,
\begin{equation}
    \mathbb{P}\paraBrTerm{z_l^{(s)}\in\Gamma_f^{(s)}} = 0.
\end{equation}
Since
\begin{equation}
    \curBarTerm{\mathbf{S}^{\mathrm{on}} \subset \Gamma_f^{(s)}} 
    \subseteq
    \curBarTerm{z_l^{(s)} \in \Gamma_f^{(s)}}
\end{equation}
for any $l\in\curBarTerm{1,\dotsc,N_{\mathrm{on}}}$, monotonicity of probability yields
\begin{equation}
    \mathbb{P}\paraBrTerm{\mathbf{S}^{\mathrm{on}}\subset\Gamma_f^{(s)}} = 0,
\end{equation}
which proves \eqref{eqn:ensuringNonSltnSmpls}. Notice that no independence assumption among the finite collection of samples is required for this conclusion.

\qedWhtThmEnv

\subsection{Additive Solution Subalgebra and Ideal}
\label{subapp:additiveSolutionSubalgebraAndIdeal}
This section provides the proof of Proposition~\ref{pro:addSltnIdeal}. We first establish the general statement for a linear DE system
\begin{equation}
    \Delta^{(s)} \coloneqq \mathrm{L}^{(s)}\sqBrTerm{\mathbf{u}} - \mathbf{g} = \mathbf{0}.
\end{equation}
Let $v_{\boldsymbol{\varphi}}\in\mathcal{A}_{\mathrm{sol}}$, so that $\boldsymbol{\varphi}$ belongs to the kernel of the associated \textit{homogeneous operator},
\begin{equation}
    \mathrm{L}^{(s)}\sqBrTerm{\sqBrTerm{j^s i_{\boldsymbol{\varphi}}}\paraBrTerm{\mathbf{x}}} = \mathbf{0}.
\end{equation}
The local one-parameter transformation generated by $v_{\boldsymbol{\varphi}}$ acts only in the dependent-variable directions and is given by
\begin{equation}
    \mathbf{u} \longmapsto \mathbf{u} + \varepsilon\boldsymbol{\varphi}\paraBrTerm{\mathbf{x}}.
\end{equation}
By linearity of $\mathrm{L}^{(s)}$,
\begin{equation}
    \begin{split}
        \mathrm{L}^{(s)}\sqBrTerm{\mathbf{u} + \varepsilon\boldsymbol{\varphi}} - \mathbf{g} &=
        \mathrm{L}^{(s)}\sqBrTerm{\mathbf{u}} - \mathbf{g} + \varepsilon\mathrm{L}^{(s)} \sqBrTerm{\boldsymbol{\varphi}} \\
        &= \mathrm{L}^{(s)}\sqBrTerm{\mathbf{u}} - \mathbf{g}.
    \end{split}
\end{equation}
Hence the transformation preserves the solution set of the original, possibly inhomogeneous, linear DE system, and therefore every $v_{\boldsymbol{\varphi}}\in\mathcal{A}_{\mathrm{sol}}$ is an admitted Lie point-symmetry generator. Since $\ker\mathrm{L}^{(s)}$ is a vector space, $\mathcal{A}_{\mathrm{sol}}$ is closed under linear combinations.

It remains to verify closure under the Lie bracket. For any $v_{\boldsymbol{\varphi}_1},v_{\boldsymbol{\varphi}_2}\in\mathcal{A}_{\mathrm{sol}}$,
the coefficient functions $\boldsymbol{\varphi}_1\paraBrTerm{\mathbf{x}}$ and $\boldsymbol{\varphi}_2\paraBrTerm{\mathbf{x}}$ depend only on the independent variables. Since both vector fields are purely vertical,
\begin{equation}
    v_{\boldsymbol{\varphi}_1}\paraBrTerm{\varphi_2^\alpha} =
    v_{\boldsymbol{\varphi}_2}\paraBrTerm{\varphi_1^\alpha} =
    0,
    \quad
    \alpha=1,\dotsc,q.
\end{equation}
Consequently,
\begin{equation}
    \sqBrTerm{v_{\boldsymbol{\varphi}_1}, v_{\boldsymbol{\varphi}_2}} = \mathbf{0}.
\end{equation}
Thus $\mathcal{A}_{\mathrm{sol}}\leq\mathfrak{g}$ and is an Abelian Lie subalgebra of the admitted Lie point-symmetry algebra $\mathfrak{g}_{\mathrm{pt}}$.

We now assume that the admitted Lie point-symmetry generators have the projectable fiber-affine form
\begin{equation}
    v = \sum_{i=1}^{p}\xi^i\paraBrTerm{\mathbf{x}}\partial_{x^i} +
        \sum_{\alpha=1}^{q}
        \paraBrTerm{
            A^\alpha_\beta\paraBrTerm{\mathbf{x}}u^\beta
            +
            b^\alpha\paraBrTerm{\mathbf{x}}
        }\partial_{u^\alpha}, 
\end{equation}
where summation over $\beta$ is dropped for notational simplicity. Let $v_{\boldsymbol{\varphi}}\in\mathcal{A}_{\mathrm{sol}}$ be arbitrary. Since the coefficients $\xi^i$ are independent of $\mathbf{u}$, the horizontal component of the Lie bracket vanishes,
\begin{equation}
    v_{\boldsymbol{\varphi}}\paraBrTerm{\xi^i} = 0,
    \quad
    i=1,\dotsc,p.
\end{equation}
For the vertical component, direct evaluation gives
\begin{equation}
    \begin{split}
        \sqBrTerm{v,v_{\boldsymbol{\varphi}}} &=
        \sum_{\alpha=1}^{q}\widetilde{\varphi}^{\,\alpha}\paraBrTerm{\mathbf{x}}\partial_{u^\alpha},
        \\
        \widetilde{\varphi}^{\,\alpha} &\coloneqq
        \sum_{i=1}^{p} \xi^i\partial_{x^i}\varphi^\alpha -
        \sum_{\beta=1}^{q}A^\alpha_\beta\varphi^\beta.
    \end{split}
\end{equation}
Thus the bracket is again purely vertical, and its coefficients depend only on the independent variables.

Because both $v$ and $v_{\boldsymbol{\varphi}}$ are admitted Lie point-symmetry generators, closure of the admitted symmetry algebra under the Lie bracket implies that $\sqBrTerm{v,v_{\boldsymbol{\varphi}}}$ is also admitted. Its local flow is an additive transformation
\begin{equation}
    \mathbf{u} \longmapsto \mathbf{u} + \varepsilon\widetilde{\boldsymbol{\varphi}}\paraBrTerm{\mathbf{x}}.
\end{equation}
Applying the same linearity argument used above to this admitted transformation gives
\begin{equation}
    \left. \mathrm{pr}^{(s)}\paraBrTerm{v_{\widetilde{\varphi}}}\sqBrTerm{\Delta^{(s)}} \right\vert_{\Sigma^{(s)}} = 
    \mathrm{L}^{(s)}\sqBrTerm{\widetilde{\boldsymbol{\varphi}}} = \mathbf{0}.
\end{equation}
Hence
\begin{equation}
    \sqBrTerm{v, v_{\boldsymbol{\varphi}}} =
    v_{\widetilde{\boldsymbol{\varphi}}} \in \mathcal{A}_{\mathrm{sol}}.
\end{equation}
Since $v\in\mathfrak{g}$ and $v_{\boldsymbol{\varphi}}\in\mathcal{A}_{\mathrm{sol}}$ were arbitrary,
\begin{equation}
    \sqBrTerm{\mathfrak{g}, \mathcal{A}_{\mathrm{sol}}} \subseteq \mathcal{A}_{\mathrm{sol}}.
\end{equation}
Therefore, whenever the admitted symmetry algebra has the PFA form, $\mathcal{A}_{\mathrm{sol}}$ is an Abelian ideal of $\mathfrak{g}$.

\qedWhtThmEnv

\subsection{Sampled Triviality and Geometric Quotient}
\label{subapp:sampledTrivAndGeomQuotient}
This subsection provides the proof of Proposition~\ref{pro:sampledTrivQuotient}. Let
$V\coloneqq\widehat{N}_{\mathrm{nadd}}^{(k)}$ and
\begin{equation}
    T \coloneqq \left. \widehat{\mathcal{T}}_{\mathrm{triv}}^{(k)} \right\vert_{\mathbf{S}_{\mathrm{anc}}^{\mathrm{on}},V}
\end{equation}
for notational simplicity. Since $T$ is linear, $\ker T$ is a vector subspace of $V$. By the definition of the sampled trivial sector in \eqref{eqn:trivVecSubsp},
\begin{equation}
    \ker T = \widehat{K}_{\mathrm{triv}}^{(k)}.
\end{equation}
This proves the first statement.

For the second statement, the \textit{First Isomorphism Theorem}~\cite[Section~3.3]{dummit2003abstract} applied to the linear map $T$ gives
\begin{equation}
    V / \ker T \cong \operatorname{im}\paraBrTerm{T}.
\end{equation}
Using the definition of the geometric quotient in \eqref{eqn:geomNonTrivSubsp}, we therefore obtain
\begin{equation}
    \widehat{N}_{\mathrm{geom}}^{(k)} = 
    V / \widehat{K}_{\mathrm{triv}}^{(k)} =
    V / \ker T \cong \operatorname{im}\paraBrTerm{T}.
\end{equation}
Thus the quotient identifies coefficient directions that differ only by an element of the sampled trivial sector, while its image under the induced isomorphism records the corresponding sampled prolonged-characteristic information.

Finally,
\begin{equation}
    \widehat{\mathcal{V}}_{\mathrm{ntr}}^{(k)} = V \setminus \ker T.
\end{equation}
Since $\mathbf{0}\in\ker T$, it follows immediately that $\mathbf{0}\notin\widehat{\mathcal{V}}_{\mathrm{ntr}}^{(k)}$.
Hence $\widehat{\mathcal{V}}_{\mathrm{ntr}}^{(k)}$ is not a vector subspace of $V$. Moreover, whenever this set is nonempty, any $\boldsymbol{\Theta}\in\widehat{\mathcal{V}}_{\mathrm{ntr}}^{(k)}$ also satisfies $-\boldsymbol{\Theta}\in\widehat{\mathcal{V}}_{\mathrm{ntr}}^{(k)}$, whereas
\begin{equation}
    \boldsymbol{\Theta} + \paraBrTerm{-\boldsymbol{\Theta}} =
    \mathbf{0} \in \ker T.
\end{equation}
Thus the sampled nontrivial feasible set is not closed under vector addition, completing the third statement.

\qedWhtThmEnv

\subsection{First-Order ODEs and Residual Functional Freedom}
\label{subapp:firstOrdODEResidFuncFreedom}
This appendix proves Proposition~\ref{pro:firstOrdODEResidFuncFreedom} and explains why the characteristic-based removal of solution-tangent symmetries does not eliminate the arbitrary functional freedom inherent to unrestricted first-order ODE point symmetries.

Associated with the first-order system~\eqref{eqn:firstOrdODESolved} is the nonvanishing vector field
\begin{equation}
    \label{eqn:firstOrdODEDirField}
    X = \partial_x + \sum_{\alpha=1}^{q} F^\alpha\paraBrTerm{x,\mathbf{u}}\partial_{u^\alpha}
\end{equation}
on the local total space $E$. Its integral curves are precisely the lifted solution curves of the ODE system. By the \textit{local flow-box theorem}~\cite[Theorem~1.2.2]{ivey2003cartan}, around every regular point there exist point coordinates $\paraBrTerm{s,r^1,\dotsc,r^q}$ such that
\begin{equation}
    \label{eqn:firstOrdODEFlowBox}
    X = \partial_s.
\end{equation}
The coordinates $r^\alpha$ are therefore constant along the integral curves of $X$ and locally label the solution leaves. In these coordinates, the ODE system assumes the rectified form~\eqref{eqn:firstOrdODERectified}.

Consider a general local point vector field
\begin{equation}
    \label{eqn:firstOrdODEGenGeneral}
    v = A\paraBrTerm{s,\mathbf{r}} \partial_s + 
        \sum_{\alpha=1}^{q} B^\alpha\paraBrTerm{s,\mathbf{r}}\partial_{r^\alpha}.
\end{equation}
Its first prolongation along the derivative coordinate $r_s^\alpha$ has coefficient (see \eqref{eqn:prolongationOfGens})
\begin{equation}
    \label{eqn:firstOrdODEFirstProlCoeff}
    \varphi_s^\alpha = D_s B^\alpha - r_s^\alpha D_s A.
\end{equation}
Restricting to the equation manifold $r_s^\alpha=0$
gives
\begin{equation}
    \label{eqn:firstOrdODEDetRect}
    \left. \varphi_s^\alpha \right\vert_{\Sigma^{(1)}} = \frac{\partial B^\alpha}{\partial s}.
\end{equation}
Therefore, the infinitesimal invariance condition is equivalent to
\begin{equation}
    \label{eqn:firstOrdODEBIndepS}
    \frac{\partial B^\alpha}{\partial s} = 0,
    \quad
    \alpha=1,\dotsc,q,
\end{equation}
while no corresponding restriction is imposed on $A\paraBrTerm{s,\mathbf{r}}$. Hence every admitted point generator is locally of the form
\begin{equation}
    \label{eqn:firstOrdODEGenSolved}
    v = A\paraBrTerm{s,\mathbf{r}} \partial_s +
        \sum_{\alpha=1}^{q} B^\alpha\paraBrTerm{\mathbf{r}}\partial_{r^\alpha}.
\end{equation}

For the rectified system, the characteristic components are
\begin{equation}
    \label{eqn:firstOrdODECharProof}
    Q_v^\alpha = B^\alpha \paraBrTerm{\mathbf{r}} - r_s^\alpha A\paraBrTerm{s,\mathbf{r}}.
\end{equation}
Thus, on $\Sigma^{(1)}$,
\begin{equation}
    \label{eqn:firstOrdODECharOnShellProof}
    \left. Q_v^\alpha \right\vert_{\Sigma^{(1)}} = B^\alpha\paraBrTerm{\mathbf{r}}.
\end{equation}
It follows that the arbitrary vector-field family $A\paraBrTerm{s,\mathbf{r}}\partial_s$ has vanishing characteristic on the equation manifold and hence, is tangent to the solution leaves. This is precisely the sector removed by the characteristic-based nontriviality criterion $\mathcal{L}_\mathrm{tail}$ in \eqref{eqn:gatedNonTrivCost}.

In contrast, whenever at least one $B^\alpha\paraBrTerm{\mathbf{r}}$ is nonzero, the characteristic is nonvanishing and the corresponding generator is nontrivial. Since the functions $B^\alpha\paraBrTerm{\mathbf{r}}$ are otherwise arbitrary, the resulting nontrivial quotient contains the infinite-dimensional Lie algebra of local vector fields on the $q$-dimensional space of solution leaves. Therefore, removing the characteristic-zero tangent sector does not reduce the unrestricted local Lie point-symmetry structure of a regular first-order ODE system to a finite-dimensional algebra.

\qedWhtThmEnv

\subsection{Projectable Fiber-Affine Form (PFA)}
\label{subapp:projectableFiberAffineForm}
The proof of Proposition~\ref{pro:projFibAffSym}, which states that the specified classes of linear DEs admit only Lie point-symmetry generators of the PFA form in~\eqref{eqn:projFibAffSym}, is provided in this subsection. We begin from the general point-symmetry representation in~\eqref{eqn:algEltRepOnE}. It is sufficient to establish
\begin{equation}
    \label{eqn:pfaTargetProjectability}
    \partial_{u^\alpha}\xi^i=0,
    \quad
    i=1,\dotsc,p,\; \alpha=1,\dotsc,q,
\end{equation}
together with
\begin{equation}
    \label{eqn:pfaTargetFiberAffinity}
    \partial_{u^\beta}\partial_{u^\gamma}\varphi^\alpha=0,
    \quad
    \alpha,\beta,\gamma=1,\dotsc,q.
\end{equation}
Indeed, \eqref{eqn:pfaTargetProjectability} implies $\xi^i=\xi^i\paraBrTerm{\mathbf{x}}$, while \eqref{eqn:pfaTargetFiberAffinity} implies that each fiber coefficient is affine in the dependent variables,
\begin{equation}
    \varphi^\alpha\paraBrTerm{\mathbf{x},\mathbf{u}} =
    \sum_{\beta=1}^q A^\alpha_{\beta}\paraBrTerm{\mathbf{x}}u^\beta + b^\alpha\paraBrTerm{\mathbf{x}},
\end{equation}
which yields precisely the PFA form in~\eqref{eqn:projFibAffSym}.

\textit{Linear ODEs with $s\geq3$:}
For the ODE case, let $x$ denote the single independent variable and write
$u_r^\alpha\coloneqq \nicefrac{d^r u^\alpha}{d x^r}$, with $u_0^\alpha\coloneqq u^\alpha$. A general order-$s$ linear system may be written locally as
\begin{equation}
    \label{eqn:linODEGeneralPFAProof}
    \Delta^\nu = a^\nu_{\alpha_s}\paraBrTerm{x} u_s^{\alpha_s} +
    \sum_{r=0}^{s-1} a^\nu_{\alpha_r}\paraBrTerm{x} u_r^{\alpha_r} - f^\nu\paraBrTerm{x} = 0,
\end{equation}
where summation over the repeated dependent-variable indices $\alpha_r$ is understood. By the assumed nondegeneracy of the highest-order part in Section~\ref{sec:preliminaries}, the system can be solved locally for the highest derivatives and equivalently written as
\begin{equation}
    \label{eqn:linODENormalPFAProof}
    u_s^\alpha = F^\alpha\paraBrTerm{x,\mathbf{u}_{(s-1)}},
\end{equation}
where each $F^\alpha$ is affine in the lower-order jet coordinates~\cite[The Cauchy-Kovalevskaya Theorem]{olver2000applications}.

For the general generator in~\eqref{eqn:algEltRepOnE}, the one-dimensional specialization of~\eqref{eqn:prolongationOfGens} satisfies the recursion
\begin{equation}
    \label{eqn:odeProlRecPFAProof}
    \varphi_{r+1}^\alpha = D_x\varphi_r^\alpha - u_{r+1}^\alpha D_x\xi,
    \quad
    0\leq r<s,
\end{equation}
with $\varphi_0^\alpha\coloneqq\varphi^\alpha$. Iterating~\eqref{eqn:odeProlRecPFAProof} gives
\begin{equation}
    \label{eqn:odeProlClosedPFAProof}
    \varphi_s^\alpha = D_x^s\varphi^\alpha -
    \sum_{r=1}^{s} \binom{s}{r} u_{s-r+1}^\alpha D_x^r\xi.
\end{equation}
The infinitesimal symmetry condition in \eqref{eqn:symCondOverInfiniActOnEqnMan} for the normal form~\eqref{eqn:linODENormalPFAProof} is therefore
\begin{equation}
    \label{eqn:odeDetCondPFAProof}
    \left. 
        \curBarTerm{\varphi_s^\alpha - \mathrm{pr}^{(s-1)}\paraBrTerm{v}\sqBrTerm{F^\alpha}}
    \right\vert_{u_s^\alpha=F^\alpha}
    = 0.
\end{equation}
Because the lower jet coordinates $\mathbf{u}_{(s-1)}$ remain functionally independent after eliminating $\mathbf{u}_s$, this identity may be split with respect to their independent monomials.

We first establish projectability. At the lowest order covered by the proposition, $s=3$, the recursion~\eqref{eqn:odeProlRecPFAProof} produces in $\varphi_3^\alpha$ the quadratic term
\begin{equation}
    \label{eqn:odeXiUTermS3}
    -3 u_2^\alpha u_2^\beta \partial_{u^\beta}\xi.
\end{equation}
No term with the same quadratic dependence on $\mathbf{u}_2$ is generated by $\mathrm{pr}^{(2)}\paraBrTerm{v}\sqBrTerm{F^\alpha}$ or by the on-shell substitution for $\mathbf{u}_3$, since $F^\alpha$ is affine in the lower jet coordinates. Splitting~\eqref{eqn:odeDetCondPFAProof} with respect to the products $u_2^\alpha u_2^\beta$ therefore yields
\begin{equation}
    \partial_{u^\beta}\xi=0,
    \qquad
    \beta=1,\dotsc,q.
\end{equation}

For $s>3$, the same recursive structure produces the terms
\begin{equation}
    \label{eqn:odeXiUTermHigherS}
    -\binom{s}{2} u_{s-1}^\alpha u_2^\beta\partial_{u^\beta}\xi 
    -s u_2^\alpha u_{s-1}^\beta \partial_{u^\beta}\xi
\end{equation}
inside $\varphi_s^\alpha$. Again, neither the lower-order prolonged contribution in~\eqref{eqn:odeDetCondPFAProof} nor elimination of the linearly occurring highest derivatives generates quadratic monomials with this derivative-order structure. For $\beta\neq\alpha$, the coefficients of
$u_{s-1}^\alpha u_2^\beta$ and $u_{2}^\alpha u_{s-1}^\beta$ are $-\binom{s}{2}\partial_{u^\beta}\xi$ and $-s\partial_{u^\beta}\xi$, respectively, whereas for $\beta=\alpha$ the two displayed contributions combine with the nonzero coefficient $-\sqBrTerm{\binom{s}{2}+s}\partial_{u^\alpha}\xi$. Hence,
\begin{equation}
    \label{eqn:odeProjectabilityPFAProof}
    \partial_{u^\beta}\xi=0
    \quad
    \Longrightarrow
    \quad
    \xi=\xi\paraBrTerm{x}
\end{equation}
for every $s\geq3$.

With projectability established, all derivatives of $\xi$ with respect to the dependent variables vanish. The remaining nonlinear dependence on $\mathbf{u}$ that is relevant to the highest prolongation therefore comes from $D_x^s\varphi^\alpha$. In particular, it contains
\begin{equation}
    \label{eqn:odePhiUUTermPFAProof}
    s\, u_1^\beta u_{s-1}^\gamma \partial_{u^\beta}\partial_{u^\gamma}\varphi^\alpha.
\end{equation}
For $s=3$, this is precisely the term $3u_1^\beta u_2^\gamma\partial_{u^\beta}\partial_{u^\gamma}\varphi^\alpha$. After~\eqref{eqn:odeProjectabilityPFAProof} has been imposed, no other contribution to~\eqref{eqn:odeDetCondPFAProof} has the same $u_1^\beta u_{s-1}^\gamma$ dependence. Splitting with respect to these independent jet monomials therefore gives
\begin{equation}
    \label{eqn:odeFiberAffinityPFAProof}
    \partial_{u^\beta} \partial_{u^\gamma} \varphi^\alpha = 0,
    \quad
    \alpha,\beta,\gamma=1,\dotsc,q.
\end{equation}
Consequently,
\begin{equation}
    \xi=\xi\paraBrTerm{x},
    \quad
    \varphi^\alpha = A^\alpha_{\;\beta}\paraBrTerm{x}u^\beta + b^\alpha\paraBrTerm{x},
\end{equation}
which proves the PFA characterization for linear ODE systems of genuine order $s\geq3$.

The restriction $s\geq3$ is essential: the unrestricted PFA conclusion does not hold in general for linear ODEs of first or second order. Rather than reproducing the corresponding lower-order determining systems, we illustrate these two exceptions by explicit counterexamples below.

For $s=1$, consider the scalar linear equation
\begin{equation}
    \Delta^{(1)} \coloneqq u_1 = 0
\end{equation}
and the point vector field
\begin{equation}
    v = u\partial_x + u^2\partial_u.
\end{equation}
Its first prolonged coefficient is
\begin{equation}
    \varphi_1 = D_x\paraBrTerm{u^2} - u_1D_x\paraBrTerm{u} = 2uu_1-u_1^2,
\end{equation}
and therefore
\begin{equation}
    \left. \mathrm{pr}^{(1)}\paraBrTerm{v}\sqBrTerm{\Delta^{(1)}} \right\vert_{\Sigma^{(1)}} =
    \left. \varphi_1 \right\vert_{u_1=0} =
    0.
\end{equation}
Hence $v$ is an admitted point-symmetry generator, although
\begin{equation}
    \partial_u\xi=1\neq0,
    \qquad
    \partial_u^2\varphi=2\neq0,
\end{equation}
so it is neither projectable nor fiber-affine.

For $s=2$, consider similarly
\begin{equation}
    \Delta^{(2)} \coloneqq u_2 = 0
\end{equation}
and
\begin{equation}
    v = xu\partial_x + u^2\partial_u.
\end{equation}
The first two prolonged coefficients are
\begin{equation}
    \varphi_1 = uu_1-xu_1^2
\end{equation}
and
\begin{equation}
    \varphi_2 = D_x\varphi_1-u_2D_x\paraBrTerm{xu} = -3xu_1u_2.
\end{equation}
Consequently,
\begin{equation}
    \left. \mathrm{pr}^{(2)}\paraBrTerm{v}\sqBrTerm{\Delta^{(2)}} \right\vert_{\Sigma^{(2)}} =
    \left. \varphi_2 \right\vert_{u_2=0} = 0,
\end{equation}
so $v$ is again an admitted point-symmetry generator. However,
\begin{equation}
    \partial_u\xi=x,
    \qquad
    \partial_u^2\varphi=2,
\end{equation}
and hence the generator is not of PFA form. These examples show explicitly why the unrestricted ODE statement in Proposition~\ref{pro:projFibAffSym} begins at genuine order $s=3$.

\textit{Linear PDEs with $s\geq2$:}
Let $\mathbf{x}=\paraBrTerm{x^1,\dotsc,x^p}$, with $p\geq2$. A linear PDE system of genuine order $s$ may be written locally as
\begin{equation}
    \label{eqn:linPDEGeneralPFAProof}
    \begin{split}
        \Delta^\nu &= \sum_{\alpha=1}^{q} \sum_{\vert I\vert=s} a_{\alpha,I}^{\nu}\paraBrTerm{\mathbf{x}} u_I^\alpha + \\
        &\quad
        \sum_{\alpha=1}^{q} \sum_{\vert I\vert<s} a_{\alpha,I}^{\nu}\paraBrTerm{\mathbf{x}} u_I^\alpha - 
        f^\nu\paraBrTerm{\mathbf{x}} = 0.
    \end{split}
\end{equation}
As in the ODE case, the assumed nondegeneracy of the highest-order part permits the appropriate dependent order-$s$ derivatives to be eliminated locally, while the remaining jet coordinates provide functionally independent coordinates on $\Sigma^{(s)}$.

We first consider the lowest order covered by the proposition, $s=2$. From the prolongation recursion in~\eqref{eqn:prolongationOfGens}, the first-order prolonged coefficients are
\begin{equation}
    \label{eqn:firstPDEProlCoeffPFAProof}
    \varphi_i^\alpha = 
            \partial_{x^i}\varphi^\alpha + 
            u_i^\beta \partial_{u^\beta}\varphi^\alpha -
            u_j^\alpha \partial_{x^i}\xi^j 
        -u_j^\alpha u_i^\beta \partial_{u^\beta}\xi^j .
\end{equation}
Applying the recursion once more gives
\begin{equation}
    \label{eqn:secondPDEProlStructPFAProof}
    \begin{split}
        \varphi_{ij}^\alpha &= \cdots +
        u_i^\beta u_j^\gamma \partial_{u^\beta} \partial_{u^\gamma} \varphi^\alpha \\
        &\quad
        -u_{k,j}^\alpha u_i^\beta\partial_{u^\beta}\xi^k
        -u_{i,k}^\alpha u_j^\beta\partial_{u^\beta}\xi^k
        -u_k^\alpha u_{i,j}^\beta\partial_{u^\beta}\xi^k 
        +\cdots,
    \end{split}
\end{equation}
where the omitted terms are not required for the coefficient splittings below.

Consider first the terms containing one first-order and one second-order jet coordinate. Since~\eqref{eqn:linPDEGeneralPFAProof} is linear, the lower-order prolonged contributions do not generate terms of this type via \eqref{eqn:firstPDEProlCoeffPFAProof}. Restriction to $\Sigma^{(2)}$ eliminates only the dependent linear combinations of the second-order derivatives as in the previous case of an ODE in the normal form. Consequently, after this elimination, the coefficients of the remaining independent first-order--second-order products in~\eqref{eqn:secondPDEProlStructPFAProof} are linear combinations of the principal coefficients $a_{\alpha,I}^{\nu}$ multiplying
$\partial_{u^\beta}\xi^k$. By the assumed nondegeneracy of the highest-order part, these coefficients can vanish identically only if
\begin{equation}
    \label{eqn:pdeProjectabilityPFAProof}
    \partial_{u^\beta}\xi^k = 0,
    \quad
    k=1,\dotsc,p,
    \;
    \beta=1,\dotsc,q.
\end{equation}
Thus the base coefficients are projectable,
\begin{equation}
    \xi^k = \xi^k\paraBrTerm{\mathbf{x}}.
\end{equation}

Once~\eqref{eqn:pdeProjectabilityPFAProof} is imposed, all terms involving derivatives of $\xi^k$ with respect to the dependent variables disappear. The quadratic first-order contribution
\begin{equation}
    u_i^\beta u_j^\gamma \partial_{u^\beta}\partial_{u^\gamma}\varphi^\alpha
\end{equation}
therefore remains among the structurally highest nonlinear terms in~\eqref{eqn:secondPDEProlStructPFAProof}. Neither the lower-order prolonged terms nor the on-shell elimination of second-order derivatives can generate compensating quadratic products of the first-order jets, because the PDE system itself is linear. Splitting the determining equations with respect to the independent products
$u_i^\beta u_j^\gamma$, and again using nondegeneracy of the principal part, yields
\begin{equation}
    \label{eqn:pdeFiberAffinityPFAProof}
    \partial_{u^\beta}\partial_{u^\gamma}\varphi^\alpha = 0,
    \quad
    \alpha,\beta,\gamma=1,\dotsc,q.
\end{equation}
Hence every admitted generator of a second-order linear PDE system satisfying the hypotheses of Proposition~\ref{pro:projFibAffSym} is projectable and fiber-affine.

The same structure persists at every order $s>2$. Indeed, repeated use of the prolongation recursion gives, for $\vert I\vert=s$,
\begin{equation}
    \label{eqn:generalPDEProlCoeffPFAProof}
    \varphi_I^\alpha = D_I\varphi^\alpha
    -\sum_{k=1}^{p}\sum_{\mathbf{0}<J\leq I}\binom{I}{J}\paraBrTerm{D_J\xi^k} u_{I-J,k}^\alpha .
\end{equation}
Here $J\leq I$ denotes componentwise multi-index ordering, and
\begin{equation}
    \binom{I}{J} \coloneqq \frac{I!}{J!\paraBrTerm{I-J}!} =
    \prod_{i=1}^{p}\binom{I_i}{J_i},
\end{equation}
is the corresponding multi-index binomial coefficient arising from the repeated \textit{Leibniz rule} for $D_{x^i}$ and $p > 1$. The terms with $\vert J\vert=1$ contain products of an order-$s$ derivative with a first-order derivative multiplied by $\partial_{u^\beta}\xi^k$. Since the order-$s$ shell constraints remain linear in the highest-order jets, elimination of the dependent highest-order derivatives leaves independent products of this type. Splitting their coefficients therefore gives~\eqref{eqn:pdeProjectabilityPFAProof} for every $s>2$.

After projectability has been established, $D_I\varphi^\alpha$ contains, with nonzero \textit{combinatorial coefficients}, products of the form
\begin{equation}
    u_{I-e_i}^\beta u_i^\gamma\partial_{u^\beta}\partial_{u^\gamma}\varphi^\alpha,
    \quad
    \vert I\vert=s,
\end{equation}
where $e_i$ denotes the unit multi-index in the $x^i$ direction. These products involve an order-$(s-1)$ jet and a first-order jet and are unaffected by the linear elimination of the order-$s$ on-shell variables. No lower-order term in the linear determining system can cancel these nonlinear jet products. Splitting with respect to them therefore yields~\eqref{eqn:pdeFiberAffinityPFAProof}. Consequently,
\begin{equation}
    \xi^i = \xi^i\paraBrTerm{\mathbf{x}},
    \quad
    \varphi^\alpha = 
        A^\alpha_{\;\beta}\paraBrTerm{\mathbf{x}}u^\beta +
        b^\alpha\paraBrTerm{\mathbf{x}},
\end{equation}
which proves the PFA characterization for linear PDE systems of genuine order $s\geq2$.

The corresponding first-order linear PDE case is structurally different and does not admit the unrestricted conclusion above. The PFA subclass at $s=1$, together with its classical and operator-theoretic admissibility conditions, is treated separately in \ref{subapp:pfaSymOfFirstOrderLinPDEs}.

\qedWhtThmEnv

\subsection{PFA Symmetries of a First-Order Linear PDE}
\label{subapp:pfaSymOfFirstOrderLinPDEs}
This subsection provides the proofs of Propositions~\ref{pro:firstOrderLinPFA} and~\ref{pro:firstOrderLinPFAOpTh}. For the scalar first-order linear PDE in Proposition~\ref{pro:firstOrderLinPFA},
\begin{equation}
    \Delta^{(1)}[u] = a^i\paraBrTerm{\mathbf{x}}u_i + c\paraBrTerm{\mathbf{x}}u - f\paraBrTerm{\mathbf{x}} = 0,
\end{equation}
consider the prescribed PFA generator
\begin{equation}
    v = \xi^i\paraBrTerm{\mathbf{x}}\partial_{x^i} +
        \sqBrTerm{A\paraBrTerm{\mathbf{x}}u + b\paraBrTerm{\mathbf{x}}} \partial_u.
\end{equation}
The first prolonged coefficient obtained from~\eqref{eqn:prolongationOfGens} is
\begin{equation}
    \label{eqn:firstOrdPFAFirstProlProof}
    \varphi_i = \partial_{x^i}A\,u + A u_i + \partial_{x^i}b - u_j\partial_{x^i}\xi^j.
\end{equation}

For the present calculation, it is convenient to use the multiplier form of the infinitesimal symmetry condition in~\eqref{eqn:symCondOverInfiniAct}. Since the PDE is scalar, there exists locally a smooth multiplier $\lambda$ such that
\begin{equation}
    \label{eqn:firstOrdPFAMultProof}
    \mathrm{pr}^{(1)}\paraBrTerm{v}\sqBrTerm{\Delta^{(1)}} = \lambda\,\Delta^{(1)}.
\end{equation}
Substituting~\eqref{eqn:firstOrdPFAFirstProlProof} into the left-hand side gives
\begin{equation}
    \begin{split}
        \mathrm{pr}^{(1)}\paraBrTerm{v}\sqBrTerm{\Delta^{(1)}} &=
            \paraBrTerm{\xi^j\partial_{x^j}a^i - a^j\partial_{x^j}\xi^i + A a^i } u_i
        \\
        &\quad
        + \paraBrTerm{\boldsymbol{a}\cdot\nabla A + \boldsymbol{\xi}\cdot\nabla c + Ac} u
        \\
        &\quad
        +
        L[b] - \boldsymbol{\xi}\cdot\nabla f.
    \end{split}
\end{equation}
The left-hand side and $\Delta^{(1)}$ are affine in the jet coordinates $u_i$ and $u$. Hence the identity~\eqref{eqn:firstOrdPFAMultProof}, together with regularity of the genuine first-order PDE, forces the multiplier to be independent of these jet coordinates; we may therefore write $\lambda=\lambda\paraBrTerm{\mathbf{x}}$.

Splitting~\eqref{eqn:firstOrdPFAMultProof} first with respect to the independent coordinates $u_i$ yields
\begin{equation}
    \label{eqn:firstOrdPFAUiSplitProof}
    \sqBrTerm{\boldsymbol{\xi}, \boldsymbol{a}}^i + A a^i = \lambda a^i.
\end{equation}
Introducing $\mu \coloneqq \lambda-A$ we obtain
\begin{equation}
    \sqBrTerm{\boldsymbol{\xi}, \boldsymbol{a}} = \mu\boldsymbol{a}.
\end{equation}
Geometrically, this condition states that the \textit{rank-one characteristic distribution} generated by $\boldsymbol{a}$ is preserved by the flow of $\boldsymbol{\xi} = \xi^i (\mathbf{x})\partial_{x^i}$ up to a local rescaling.

Splitting next with respect to $u$ gives
\begin{equation}
    \boldsymbol{a}\cdot\nabla A + \boldsymbol{\xi}\cdot\nabla c + Ac = \lambda c,
\end{equation}
and hence
\begin{equation}
    \boldsymbol{a}\cdot\nabla A + \boldsymbol{\xi}\cdot\nabla c = \mu c.
\end{equation}
Finally, the constant term gives
\begin{equation}
    L[b] - \boldsymbol{\xi}\cdot\nabla f = -\lambda f,
\end{equation}
or, using $\lambda=A+\mu$,
\begin{equation}
    \boldsymbol{\xi}\cdot\nabla f - L[b] - Af = \mu f.
\end{equation}
These are precisely the three conditions in~\eqref{eqn:firstOrdPFACond}.

Conversely, suppose that the conditions in~\eqref{eqn:firstOrdPFACond} hold for some smooth $\mu\paraBrTerm{\mathbf{x}}$. Setting
$\lambda\coloneqq A+\mu$ in the preceding expansion gives
\begin{equation}
    \mathrm{pr}^{(1)}\paraBrTerm{v}\sqBrTerm{\Delta^{(1)}} = \paraBrTerm{A+\mu} \Delta^{(1)}.
\end{equation}
Hence the multiplier condition~\eqref{eqn:symCondOverInfiniAct} is satisfied, proving Proposition~\ref{pro:firstOrderLinPFA}.

We now prove Proposition~\ref{pro:firstOrderLinPFAOpTh}. Let
\begin{equation}
    L = \boldsymbol{a}\cdot\nabla+c,
    \quad
    R = A-\boldsymbol{\xi}\cdot\nabla.
\end{equation}
For an arbitrary smooth test function $h$, direct expansion of the commutator gives
\begin{equation}
    \sqBrTerm{L,R}h = \sqBrTerm{\boldsymbol{\xi}, \boldsymbol{a}}\cdot\nabla h
    + \paraBrTerm{\boldsymbol{a}\cdot\nabla A + \boldsymbol{\xi}\cdot\nabla c} h.
\end{equation}
Therefore, $\sqBrTerm{L,R} = \mu L$ holds if and only if
\begin{equation}
    \sqBrTerm{\boldsymbol{\xi}, \boldsymbol{a}} = \mu\boldsymbol{a},
    \quad
    \boldsymbol{a}\cdot\nabla A + \boldsymbol{\xi}\cdot\nabla c = \mu c,
\end{equation}
which are exactly the first two conditions in~\eqref{eqn:firstOrdPFACond}. Moreover,
\begin{equation}
        L[b] + \paraBrTerm{R+\mu} f = L[b] + Af - \boldsymbol{\xi}\cdot\nabla f + \mu f.
\end{equation}
Thus
\begin{equation}
    L[b] + \paraBrTerm{R+\mu}f = 0
\end{equation}
is equivalent to the third condition in~\eqref{eqn:firstOrdPFACond}. Hence the two operator identities in~\eqref{eqn:firstOrdPFACondOpTh} are equivalent to the complete PFA symmetry criterion of Proposition~\ref{pro:firstOrderLinPFA}, proving Proposition~\ref{pro:firstOrderLinPFAOpTh}.

\qedWhtThmEnv

\clearpage

\section{Flowcharts for Algorithms}
\label{app:flowChartsForAlgorithms}

\begin{figure}[h!]
    \centering
    \includegraphics[width=\textwidth]{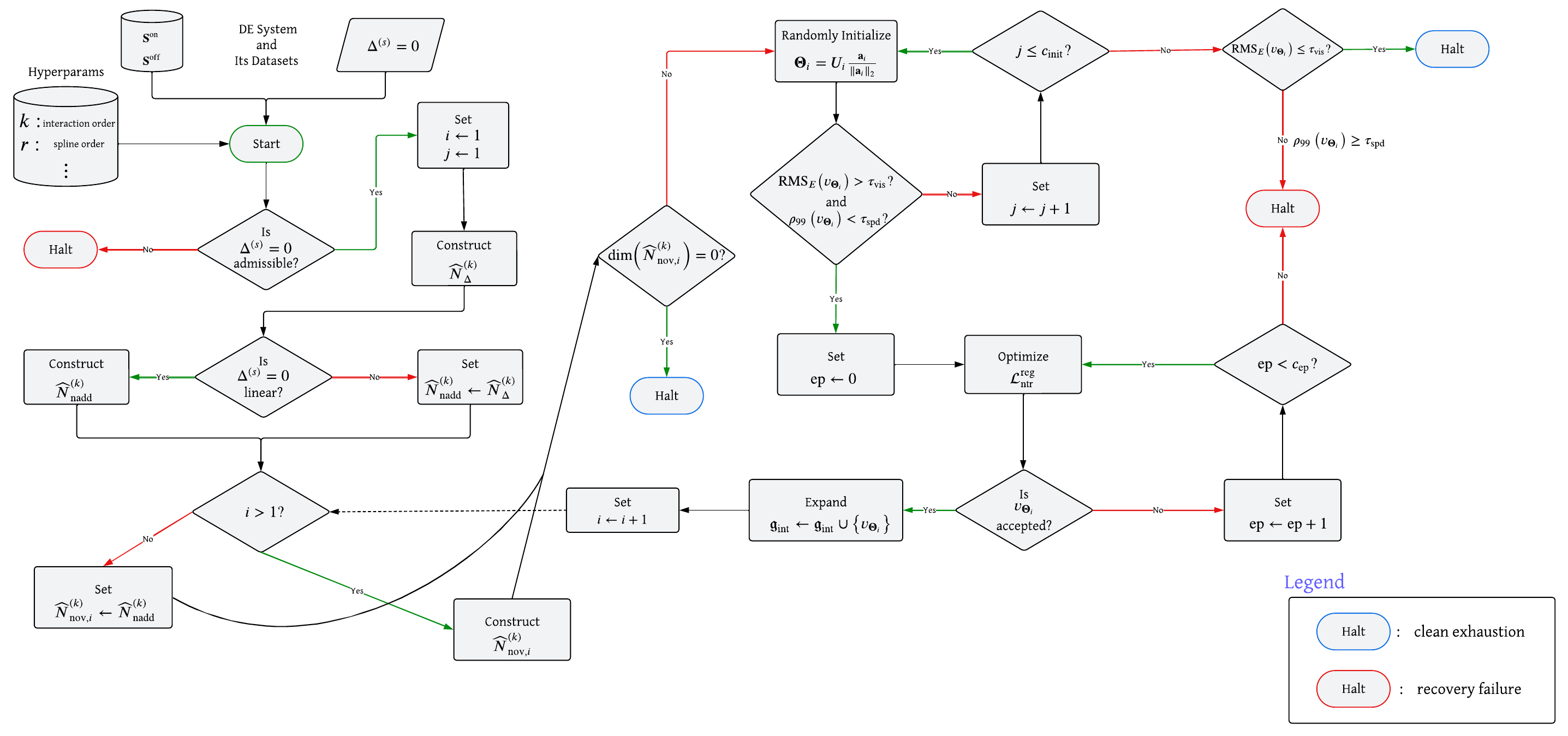}
    \caption{
        Flowchart of the sequential symmetry-discovery procedure. Starting from the DE system, its sampled datasets, and the prescribed hyperparameters, the algorithm constructs the represented determining and nonadditive admissible spaces, applies the sequential novelty restriction, initializes a candidate generator within the remaining search subspace, and optimizes the regularized nontriviality objective. Accepted generators are appended successively to the recovered intrinsic algebra and used to construct the novelty restriction for the next discovery step. Blue-bordered halt states indicate \textit{clean exhaustion} of the numerically resolvable represented symmetry directions, whereas red-bordered halt states indicate \textit{recovery failure} despite the presence of a represented candidate direction. The initial halt associated with a DE system outside the admissible scope of the framework is treated separately in the applicability flowchart of Section~\ref{sec:discApplAndLimit}.
    }
    \label{fig:sequentialRecoveryFlowchart}
\end{figure}

\begin{figure}[h!]
    \centering
    \includegraphics[width=.9\textwidth]{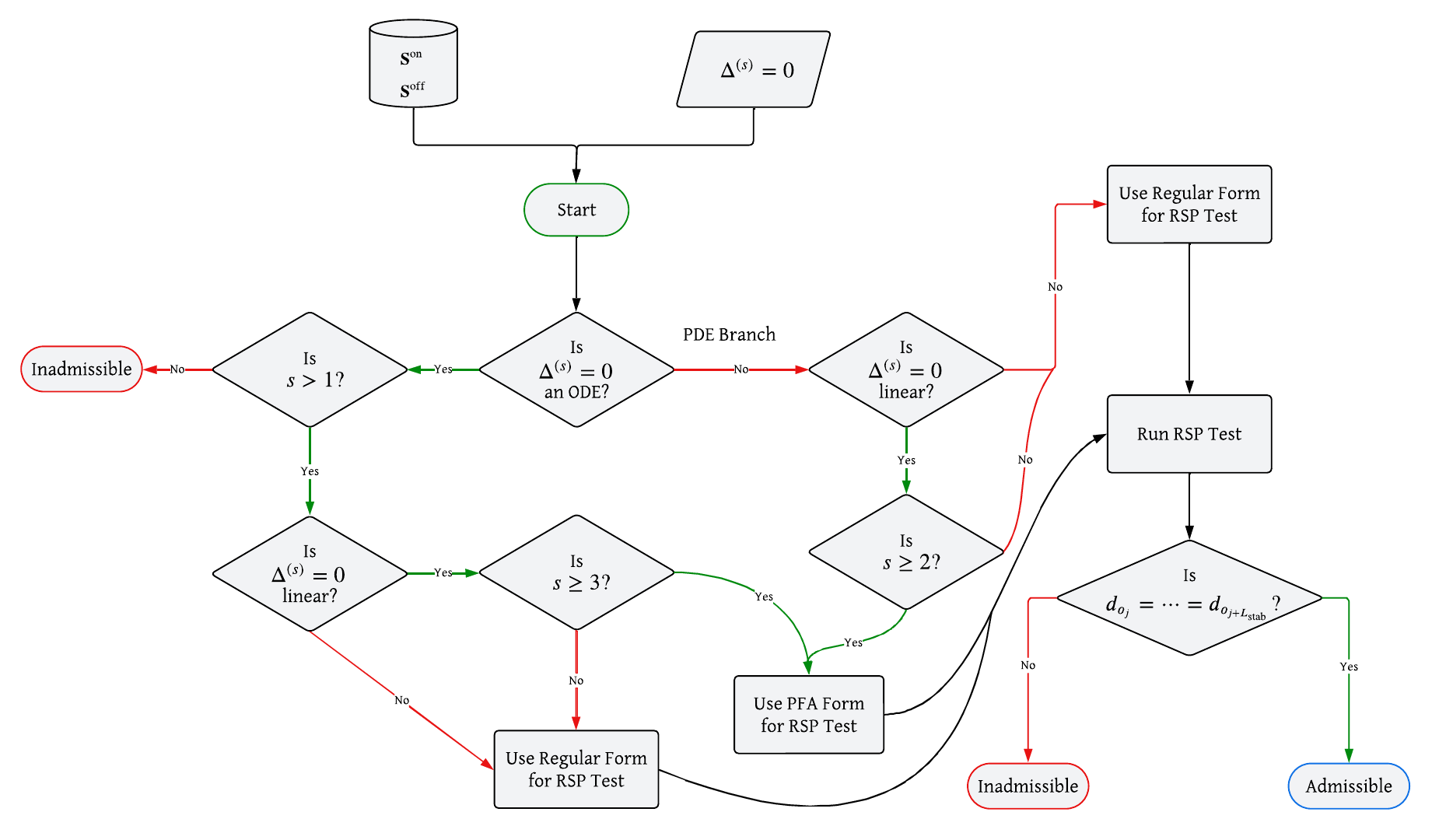}
    \caption{
        Applicability and admissible-generator realization of the proposed framework. Starting from the DE system $\Delta^{(s)}=\mathbf{0}$ and the associated on- and off-shell datasets, the flowchart first distinguishes ODEs from PDEs, then branches according to differential order and linearity. First-order ODEs are excluded from the present scope because arbitrary nontrivial functional freedom remains after removal of the characteristic-zero solution-tangent sector. For regular, determined ODE systems with $s>1$, the local point-symmetry algebra is finite-dimensional; linear ODEs with $s\geq3$ use the projectable fiber-affine (PFA) realization, whereas the remaining admissible ODE classes use the regular generator realization. For PDEs, linear systems with $s\geq2$ use the PFA realization, while nonlinear PDEs and first-order linear PDEs use the regular realization in the present implementation. All retained branches are subjected to the RSP test. A stable represented geometric-rank plateau over the prescribed refinement window permits entry into sequential recovery, whereas failure to stabilize places the DE outside the completeness-certified scope of the current framework. Here, ``Admissible'' refers to admissibility within the stabilized finite represented and sampled setting; it does not by itself establish exact formal closure or exclude unresolved arbitrary functional freedom in the underlying determining system.
    }
    \label{fig:deAdmissibility_FlowChart}
\end{figure}

\clearpage

\section{Supplementary Numerical Visualizations}
\label{app:suppNumVisualizations}
\begin{figure}[h!]
    \centering

    \begin{minipage}[t]{0.48\textwidth}
        \centering
        \includegraphics[
            width=\linewidth,
            keepaspectratio
        ]{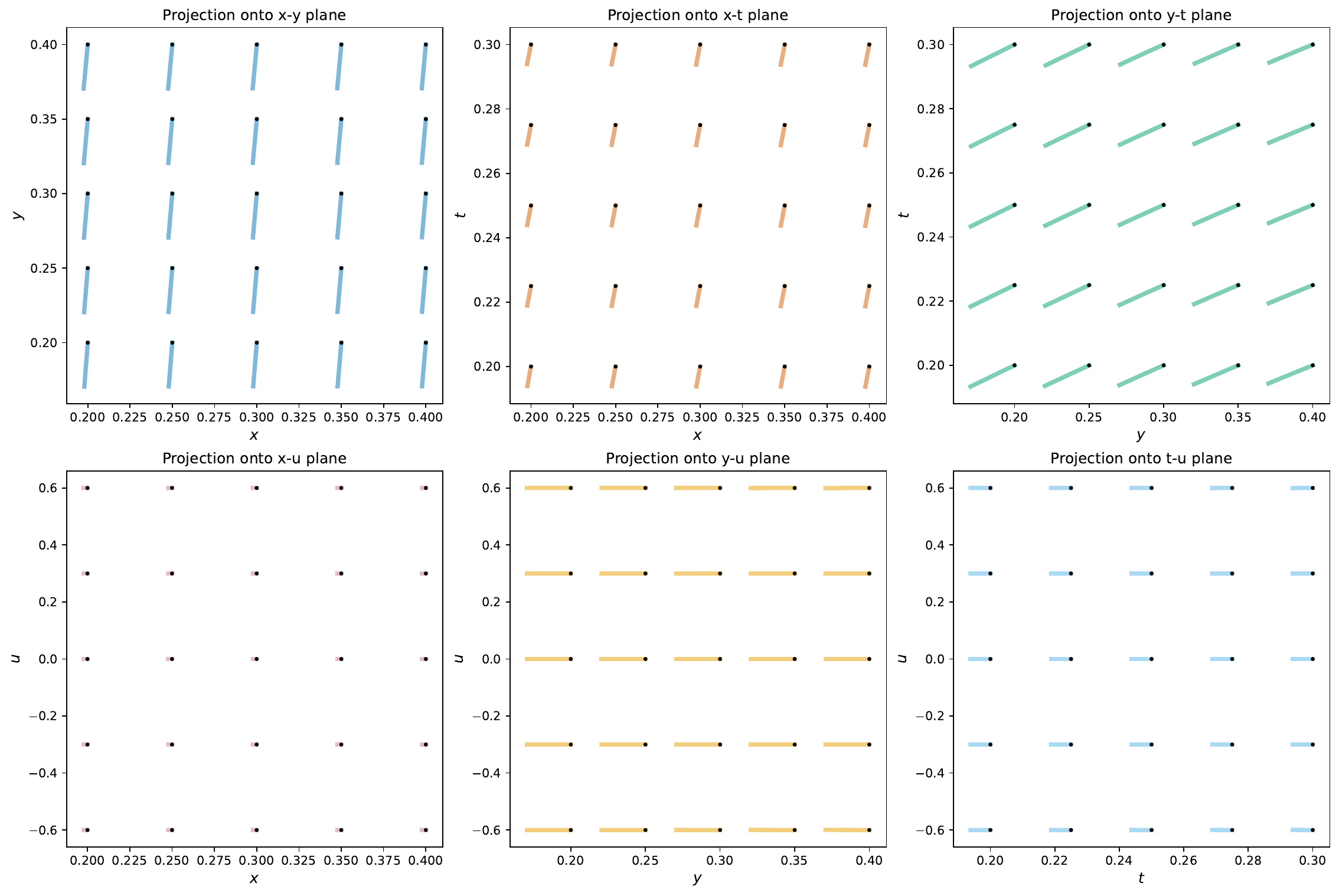}

        \vspace{-0.5em}
        \small
        (a) $v_4$, $\varepsilon_{\max}=0.15$
    \end{minipage}
    \hfill
    \begin{minipage}[t]{0.48\textwidth}
        \centering
        \includegraphics[
            width=\linewidth,
            keepaspectratio
        ]{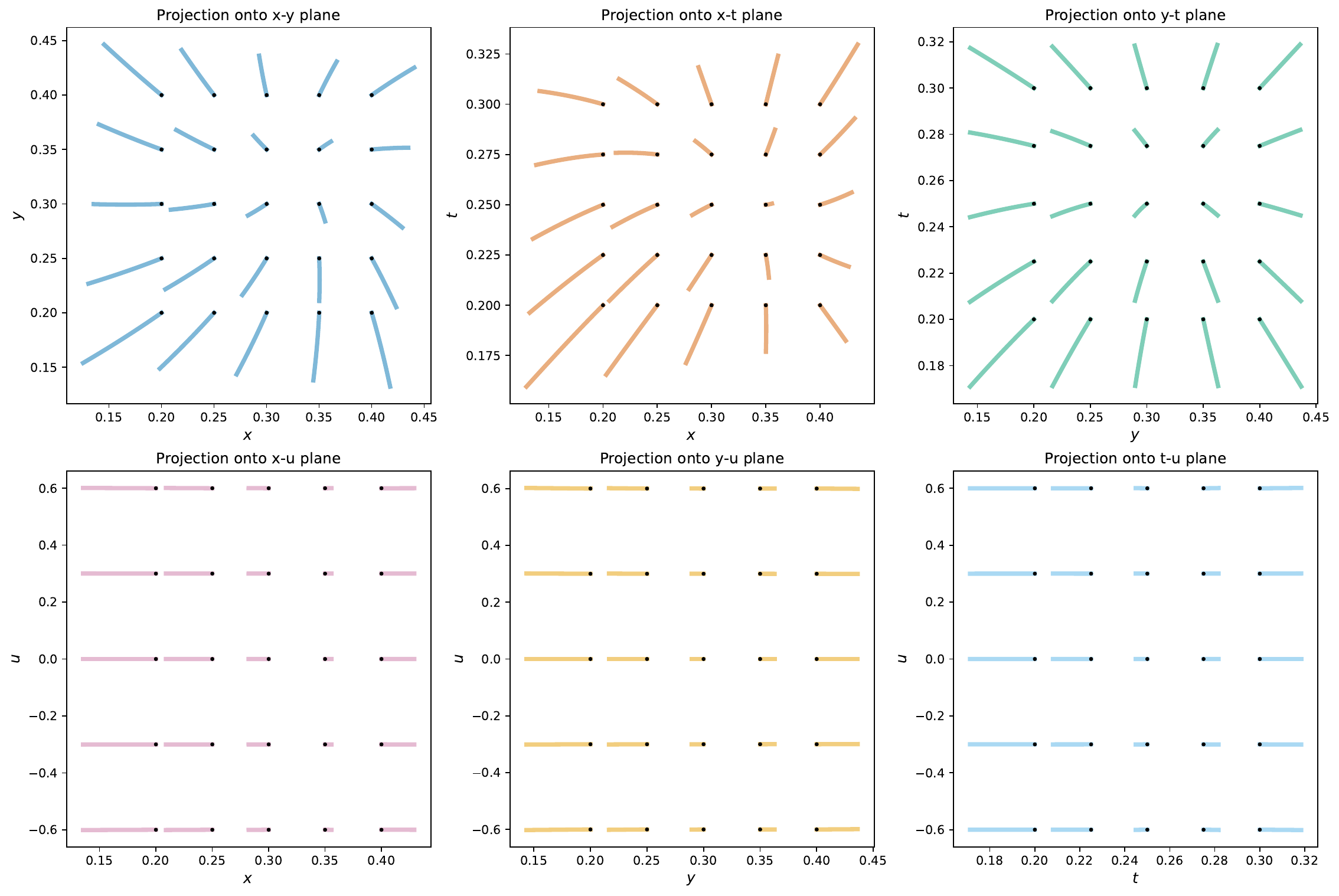}

        \vspace{-0.5em}
        \small
        (b) $v_7$, $\varepsilon_{\max}=0.05$
    \end{minipage}

    \vspace{0.5em}

    \begin{minipage}[t]{0.48\textwidth}
        \centering
        \includegraphics[
            width=\linewidth,
            keepaspectratio
        ]{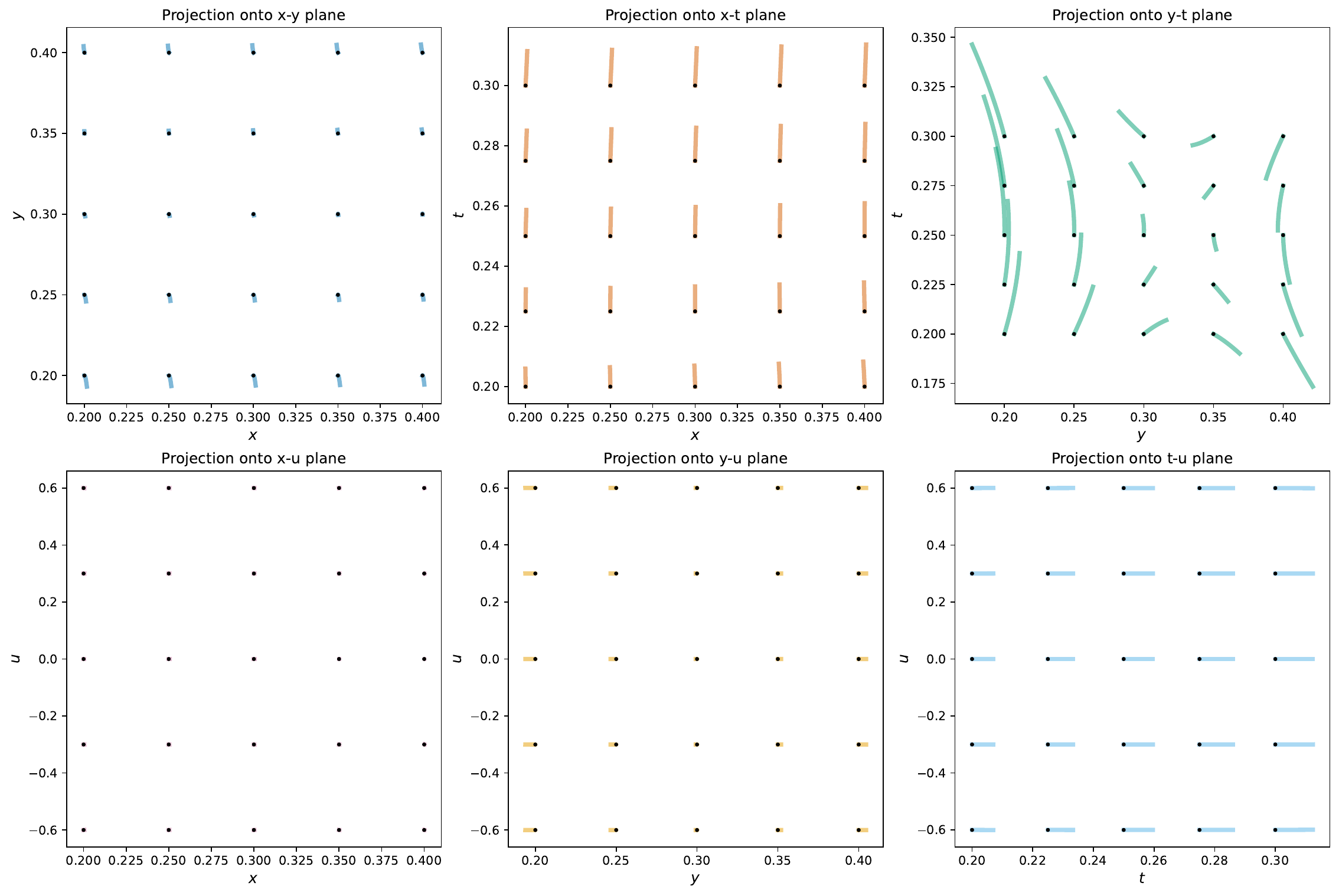}

        \vspace{-0.5em}
        \small
        (c) $v_{10}$, $\varepsilon_{\max}=0.03$
    \end{minipage}

    \caption{Representative integral curves of three recovered generators selected from the low-, intermediate-, and high-speed-tail regimes reported in Table~\ref{tab:waveEqnSeqDiscovery}. The trajectories are shown through their projections onto the $x$--$y$, $x$--$t$, $y$--$t$, $x$--$u$, $y$--$u$, and $t$--$u$ coordinate planes, with the black markers indicating the initial points. The selected generators are (a) $v_4$, with $\rho_{99}=1.002$ and $\varepsilon_{\max}=0.15$; (b) $v_7$, with $\rho_{99}=1.654$ and $\varepsilon_{\max}=0.05$; and (c) $v_{10}$, with $\rho_{99}=1.993$ and $\varepsilon_{\max}=0.03$. The integration horizons are chosen separately so that the displayed trajectories remain within, or sufficiently close to, the region in which the corresponding locally learned vector fields remain reliable; the behavior outside the training support is examined separately in \textit{\nameref*{subsec:approxOfKnownSymGenAndLocalityOfRep}}.}
    \label{fig:waveEqnRecoveredGenIntegralCurves}
\end{figure}

\begin{figure}[h!]
    \centering
    \includegraphics[width=.7\textwidth]{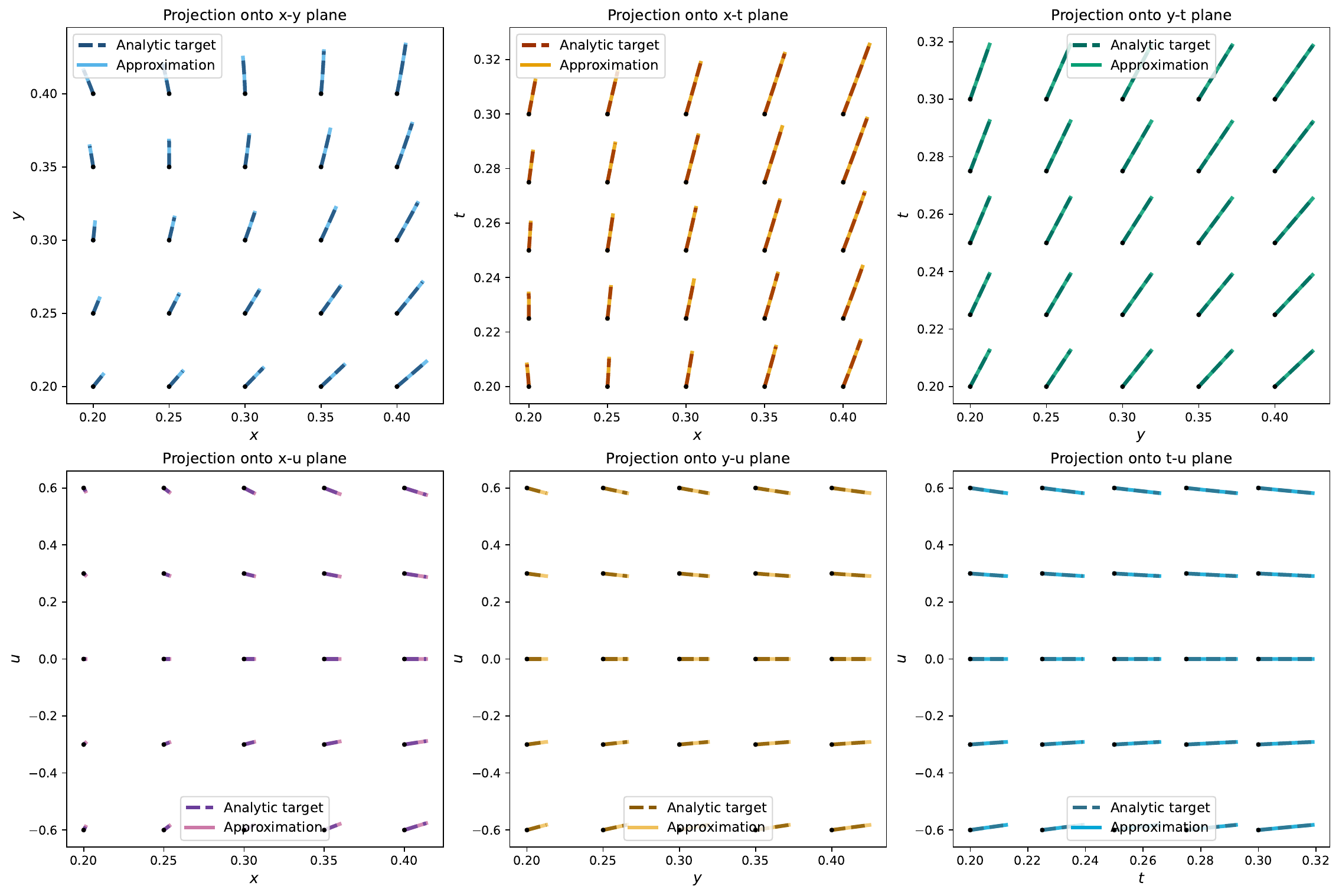}
    \caption{Integral-curve comparison between the analytical inversion-in-$x$ generator $v_{\mathrm{inv},x}$ and its constant-span reconstruction $\widetilde{v}_{\mathrm{inv},x}$, integrated from the same initial points up to $\varepsilon_{\max}=0.1$ using $100$ steps.}
    \label{fig:waveEqn2D_custSymGenSpanFit_InversionX_integCurves2D}
\end{figure}

\begin{figure}[h!]
    \centering
    \includegraphics[width=.7\textwidth]{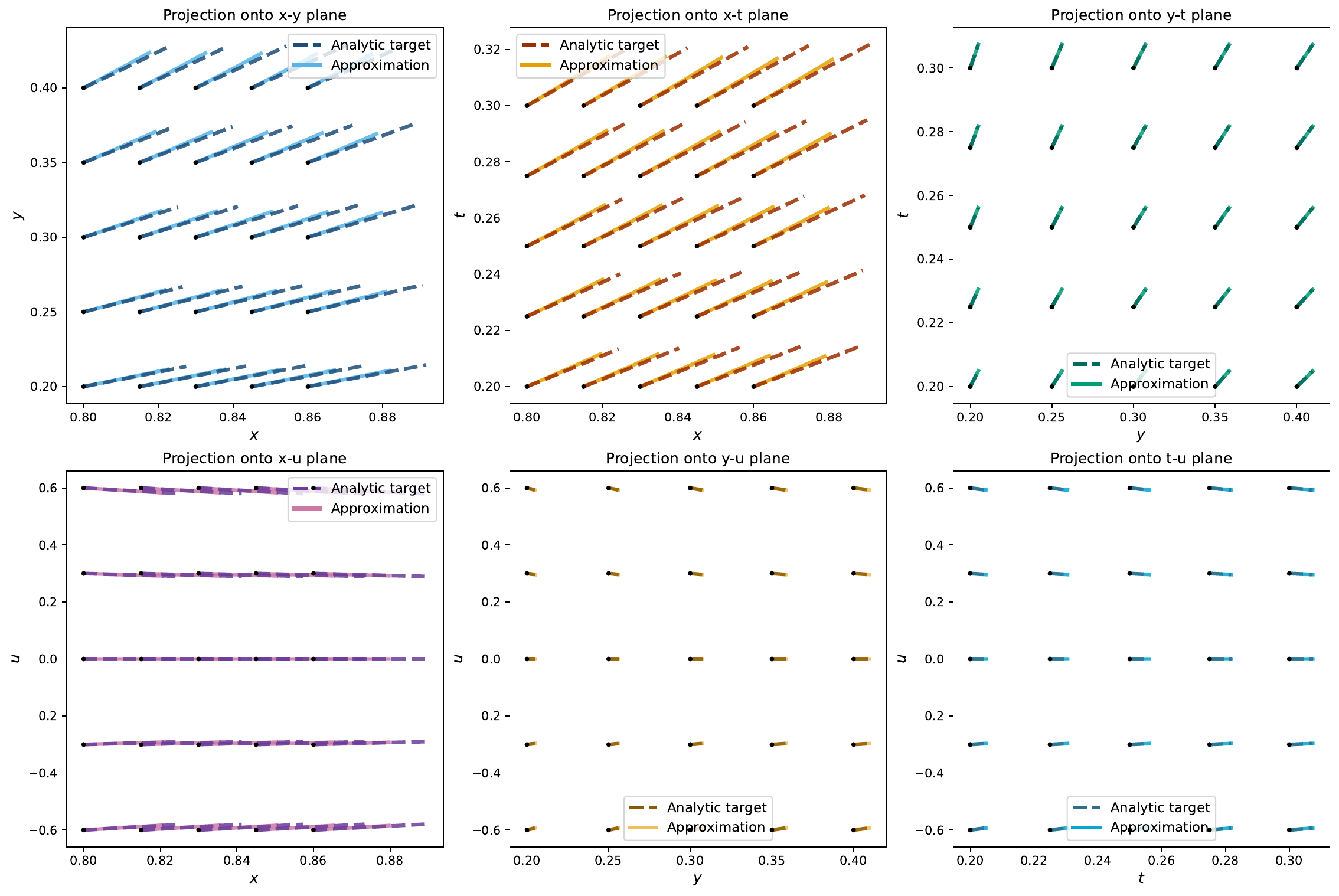}
    \caption{Integral-curve comparison between the analytical inversion-in-$x$ generator $v_{\mathrm{inv},x}$ and its constant-span reconstruction $\widetilde{v}_{\mathrm{inv},x}$ after shifting the initial $x$ interval to $\sqBrTerm{0.8,0.86}$, with the flows integrated up to $\varepsilon_{\max}=0.04$.}
    \label{fig:waveEqn2D_custSymGenSpanFit_InversionX_integCurves2D_shiftedX}
\end{figure}

\clearpage

\section{Table of Symbols}
\label{app:tableOfSymbols}

The table lists the principal manuscript-wide meaning of recurring notation. When a symbol is locally overloaded, only its principal role is recorded; section-specific uses are defined at their point of introduction.

\begingroup
\small
\setlength{\tabcolsep}{5pt}
\renewcommand{\arraystretch}{1.12}
\begin{longtable}{L{0.23\textwidth}L{0.54\textwidth}L{0.16\textwidth}}
    \toprule
    \textbf{Symbol} & \textbf{Principal meaning} & \textbf{Context} \\
    \midrule
    \endfirsthead

    \toprule
    \textbf{Symbol} & \textbf{Principal meaning} & \textbf{Context} \\
    \midrule
    \endhead

    \midrule
    \multicolumn{3}{r}{\small\textit{Continued on next page}} \\
    \endfoot

    \bottomrule
    \endlastfoot

    \multicolumn{3}{l}{\textbf{Jet-space and differential-equation geometry}} \\
    \addlinespace[0.35em]
    $\mathbf{X}$
        & Base space of the independent variables.
        & Jet geometry \\
    $E,\;\pi:E\to\mathbf{X}$
        & Total space and fiber-bundle projection over the base space.
        & Jet geometry \\
    $\mathbf{x},\;\mathbf{u}$
        & Independent- and dependent-variable coordinate vectors, respectively.
        & Jet geometry \\
    $p,\;q,\;m,\;s$
        & Numbers of independent variables, dependent variables, and constituent differential equations, and the differential order of the DE system, respectively.
        & DE system \\
    $J^{(s)}\paraBrTerm{\pi}$
        & Order-$s$ jet bundle of local sections of $\pi$.
        & Jet geometry \\
    $z^{(s)},\;\mathbf{u}_{(s)}$
        & Generic order-$s$ jet and the collection of dependent-variable coordinates and derivatives through order $s$.
        & Jet geometry \\
    $\pi_0^s,\;\pi_r^s,\;\pi^s$
        & Canonical projections from $J^{(s)}\paraBrTerm{\pi}$ onto $E$, $J^{(r)}\paraBrTerm{\pi}$, and $\mathbf{X}$, respectively.
        & Jet geometry \\
    $\Delta^{(s)},\;\Sigma^{(s)}$
        & Vector-valued differential function defining the DE system and its regular zero locus (equation manifold).
        & DE system \\
    $i_f,\;j^s i_f$
        & Local solution section and its order-$s$ jet prolongation.
        & Solution geometry \\
    $\Gamma_f,\;\Gamma_f^{(s)}$
        & Solution graph in $E$ and its order-$s$ prolonged graph in jet space.
        & Solution geometry \\

    \addlinespace[0.6em]
    \multicolumn{3}{l}{\textbf{Lie symmetries and intrinsic symmetry structure}} \\
    \addlinespace[0.35em]
    $G,\;\mathfrak{g}$
        & Lie group of local point transformations and its Lie algebra of infinitesimal generators.
        & Lie symmetry \\
    $\mathfrak{X}\paraBrTerm{E}$
        & Space of smooth vector fields on the total space $E$.
        & Lie symmetry \\
    $v$
        & Lie point-symmetry generator represented as a vector field on $E$.
        & Lie symmetry \\
    $\boldsymbol{\xi},\;\boldsymbol{\varphi}$
        & Base (horizontal) and fiber (vertical) coefficient vectors of a point generator.
        & Generator representation \\
    $\mathrm{pr}^{(s)}\paraBrTerm{v}$
        & Order-$s$ prolongation of the point generator $v$ to $J^{(s)}\paraBrTerm{\pi}$.
        & Prolongation \\
    $D_{x^i},\;D_J$
        & Total derivative with respect to $x^i$ and its multi-index iteration.
        & Prolongation \\
    $\mathbf{Q}_v,\;\mathbf{E}_v^{(s-1)}$
        & Characteristic vector of $v$ and its prolonged characteristic information through order $s-1$.
        & Nontriviality \\
    $\mathfrak{g}_{\mathrm{int}},\;\mathfrak{g}_{\mathrm{rec}}$
        & Target intrinsic symmetry algebra and the span of the generators accepted by the numerical recovery procedure, respectively.
        & Intrinsic symmetry \\
    $\mathcal{A}_{\mathrm{sol}}$
        & Additive solution subalgebra; under the stated projectable fiber-affine conditions, an Abelian ideal of the Lie symmetry algebra.
        & Linear DEs \\
    $c_{ij}^{k}$
        & Constant structure coefficients in the recovered-basis Lie-bracket representation.
        & Lie algebra \\
    $\operatorname{Fl}_{v}^{z}\paraBrTerm{\epsilon}$
        & Local flow generated by $v$ through $z$, with flow parameter $\epsilon$.
        & Finite transformations \\

    \addlinespace[0.6em]
    \multicolumn{3}{l}{\textbf{Sampled operators and represented admissible spaces}} \\
    \addlinespace[0.35em]
    $\mathbf{S}^{\mathrm{on}},\;\mathbf{S}^{\mathrm{off}}$
        & Finite on-shell jet dataset and auxiliary off-shell jet dataset.
        & Sampling \\
    $\mathbf{S}^{\mathrm{on}}_{\mathrm{anc}},\;\mathbf{S}^{\mathrm{off}}_{\mathrm{anc}}$
        & Fixed on-shell and off-shell anchor datasets used to construct sampled operators and global novelty restrictions.
        & Sampling \\
    $\mathcal{D}_{\Delta},\;\widehat{\mathcal{D}}_{\Delta},\;\widehat{\mathcal{D}}_{\Delta}^{(k)}$
        & Pointwise determining operator, its stacked sampled realization, and its represented sampled realization induced by $D_{(k)}$.
        & Sampled operators \\
    $\Psi^{(k)},\;\boldsymbol{\Theta},\;P$
        & Fixed-dictionary linear realization map, its coefficient vector, and the dimension of the global coefficient space.
        & Representation \\
    $\widehat{M}_{\Delta}^{(k)}$
        & Fixed matrix representation (Jacobian) of the represented sampled determining residual with respect to $\boldsymbol{\Theta}$.
        & Sampled operators \\
    $\widehat{N}_{\Delta}^{(k)}$
        & Represented sampled determining nullspace $\ker\paraBrTerm{\widehat{M}_{\Delta}^{(k)}}$.
        & Admissibility \\
    $\mathcal{T}_{\mathrm{nadd}},\;\widehat{\mathcal{T}}_{\mathrm{nadd}}^{(k)}$
        & Nonadditive-sector test and its sampled represented realization.
        & Additive-sector exclusion \\
    $\widehat{\mathcal{A}}_{\mathrm{sol}}^{(k)}$
        & Sampled represented additive solution sector within $\widehat{N}_{\Delta}^{(k)}$.
        & Additive-sector exclusion \\
    $\widehat{N}_{\mathrm{nadd}}^{(k)}$
        & Additive-free represented admissible coefficient subspace.
        & Admissibility \\
    $\widehat{\mathcal{T}}_{\mathrm{triv}}^{(k)}$
        & Sampled represented prolonged-characteristic operator used to identify sampled-trivial directions.
        & Nontriviality \\
    $\widehat{K}_{\mathrm{triv}}^{(k)},\;\widehat{N}_{\mathrm{geom}}^{(k)}$
        & Sampled-trivial kernel and the corresponding sampled geometric quotient modulo sampled-trivial directions.
        & Nontriviality \\
    $\widehat{\operatorname{B}}_{E,i}^{(k)},\;\widehat{N}_{\mathrm{nov},i}^{(k)}$
        & Sampled represented novelty operator and the novelty-restricted admissible search subspace at discovery step $i$.
        & Sequential discovery \\
    $\widehat{N}_{\mathrm{adm},i}^{(k)}$
        & Represented admissible coefficient space used at refinement level $i$ in the RSP test.
        & RSP \\

    \addlinespace[0.6em]
    \multicolumn{3}{l}{\textbf{Dictionary, stabilization, and numerical diagnostics}} \\
    \addlinespace[0.35em]
    $\Omega_{\mathbf{X}}$
        & Base-coordinate sampling domain; superscripted variants denote local patches in the proposed patchwise extension.
        & Sampling / locality \\
    $\boldsymbol{\chi}$
        & Input-coordinate vector supplied to the fixed generator dictionary.
        & Fixed dictionary \\
    $B_{\nu_i}^{(r)}\paraBrTerm{\chi^i}$
        & Degree-$r$ univariate B-spline basis function associated with dictionary coordinate $\chi^i$.
        & Fixed dictionary \\
    $D_{(k)}$
        & Fixed spline dictionary containing tensor-product interactions through maximum order $k$.
        & Fixed dictionary \\
    $k,\;r,\;o,\;K_o$
        & Maximum interaction order, B-spline degree, core-grid resolution, and number of univariate B-spline functions per input coordinate, with $K_o=o+r$.
        & Fixed dictionary \\
    $n_{(k)}^{D}$
        & Number of scalar features listed in the dictionary $D_{(k)}$.
        & Fixed dictionary \\
    $d_{o_i}$
        & Represented geometric rank at grid-refinement level $o_i$.
        & RSP \\
    $\mathcal{R}_{\mathrm{ntr},l}\paraBrTerm{v},\;\mathcal{R}_{\mathrm{tail}}^{\rho}\paraBrTerm{v}$
        & Relative prolonged-characteristic response at sample $l$ and its empirical lower-tail $\rho$-quantile.
        & Nontriviality \\
    $\mathcal{L}_{\mathrm{ntr}}^{\mathrm{reg}}$
        & Regularized nontriviality objective used during sequential recovery.
        & Optimization \\
    $\rho_{99}\paraBrTerm{v}$
        & Speed-tail ratio defined by the empirical $99$th-to-$50$th percentile ratio of sampled pointwise speeds.
        & Numerical diagnostics \\
    $\operatorname{RMS}_{E}\paraBrTerm{v}$
        & Off-shell ambient-response magnitude used to assess visibility of a represented vector field.
        & Sequential recovery \\
    $\tau_{\mathrm{rank}},\;\tau_{\mathrm{vis}},\;\tau_{\mathrm{spd}},\;\tau_{\mathrm{sym}},\;\tau_{\mathrm{ntr}}$
        & Principal numerical thresholds for rank determination, ambient visibility, speed-tail concentration, symmetry acceptance, and nontriviality, respectively.
        & Numerical implementation \\

    \addlinespace[0.6em]
    \multicolumn{3}{l}{\textbf{Exterior differential systems and formal integrability}} \\
    \addlinespace[0.35em]
    $\mathcal{C}^{(s)},\;\mathcal{I}_{\mathrm{diff}}^{(s)}$
        & Order-$s$ contact differential ideal on jet space and its pullback to the equation manifold, respectively.
        & Formal integrability \\
    $\Omega$
        & Independence $p$-form $dx^1\wedge\cdots\wedge dx^p$.
        & Formal integrability \\
    $\mathcal{E}_{z^{(s)}}$
        & $p$-dimensional integral element of the EDS at $z^{(s)}\in\Sigma^{(s)}$.
        & Formal integrability \\
    $\paraBrTerm{\mathcal{I}^{(s)},\mathcal{J}^{(s)}}$
        & Linear Pfaffian system formed from the restricted contact one-forms and the independence one-forms.
        & Formal integrability \\
    $\mathbf{A},\;\mathbf{A}^{(1)}$
        & Tableau of the linear Pfaffian system and its first prolongation.
        & Formal integrability \\
    $\sqBrTerm{\mathbf{T}}$
        & Essential-torsion equivalence class in the cokernel of the alternation map.
        & Formal integrability \\
\end{longtable}
\endgroup

\end{document}

%% file: preamble.tex
\usepackage{amsmath,amssymb,amsfonts}
\usepackage{amsthm}
\usepackage{graphicx}
\usepackage{textcomp}

\usepackage{bm}

\def\BibTeX{{\rm B\kern-.05em{\sc i\kern-.025em b}\kern-.08em
    T\kern-.1667em\lower.7ex\hbox{E}\kern-.125emX}}

\usepackage{array}
\usepackage{float}
\usepackage{dsfont}%
\usepackage{bbold}%
\usepackage[scr=boondoxo,scrscaled=1.05]{mathalfa}%
\usepackage{mathrsfs}%
\usepackage[inline]{enumitem}
\usepackage{multirow}
\usepackage{mathtools}%
\usepackage{nicefrac}
\usepackage{csquotes}
\usepackage{hyperref}%
\usepackage{booktabs}%
\usepackage{tabularx}%
\usepackage{longtable}
\usepackage{xcolor}
\hypersetup{%
  colorlinks=true,%
  linkcolor={red!50!black},
  citecolor={blue!65!black},
  urlcolor={blue!80!black},
  bookmarksnumbered=true,%
  bookmarksopen=true
}
\usepackage[boxruled]{algorithm2e}%
\usepackage{algcompatible}
\usepackage{etoolbox}

\newcommand{\sqBrTerm}[1]{
    \ensuremath{\left[#1\right]}
}

\newcommand{\paraBrTerm}[1]{
    \ensuremath{\left(#1\right)}
}
\newcommand{\curBarTerm}[1]{
    \ensuremath{\left\{#1\right\}}
}

\newcommand{\qedWhtThmEnv}{$\null\hfill\qedsymbol$}
\newcommand{\RNum}[1]{\uppercase\expandafter{\romannumeral #1\relax}}
\newcommand{\rNum}[1]{\lowercase\expandafter{\romannumeral #1\relax}}
\DeclareSymbolFont{bbold}{U}{bbold}{m}{n}
\DeclareSymbolFontAlphabet{\mathbbold}{bbold}

\AtBeginEnvironment{algorithm}{\algoequations}
\AtEndEnvironment{algorithm}{\restoreequations}
\newcounter{algosavedequation}
\newcommand{\algoequations}{%
  \setcounter{algosavedequation}{\value{equation}}%
  \setcounter{equation}{0}%
  \renewcommand{\theequation}{Alg\thealgocf.\arabic{equation}}%
}
\newcommand{\restoreequations}{%
  \setcounter{equation}{\value{algosavedequation}}%
}

\makeatletter%
\newcommand{\nosemic}{\renewcommand{\@endalgocfline}{\relax}}%
\newcommand{\dosemic}{\renewcommand{\@endalgocfline}{\algocf@endline}}%
\renewcommand{\@algocf@capt@plainruled}{above}%
\makeatother%

\newtheorem{definition}{Definition}

\newtheorem{proposition}{Proposition}
\newtheorem{remark}{Remark}
\newsavebox{\myDelta}
\newsavebox{\myOmega}
\newsavebox{\myLambda}
\savebox{\myDelta}{$\scriptstyle\boldsymbol{\Delta}$}
\savebox{\myOmega}{$\scriptstyle\boldsymbol{\Omega}$}
\savebox{\myLambda}{$\scriptstyle\boldsymbol{\Lambda}$}

\newcommand{\PreserveBackslash}[1]{\let\temp=\\#1\let\\=\temp}
\newcolumntype{C}[1]{>{\PreserveBackslash\centering}p{#1}}
\newcolumntype{R}[1]{>{\PreserveBackslash\raggedleft}p{#1}}
\newcolumntype{L}[1]{>{\PreserveBackslash\raggedright}p{#1}}

\graphicspath{{./}}